\documentclass[10pt,twocolumn,letterpaper]{article}
\usepackage{iccv}
\usepackage{times}
\usepackage{epsfig}
\usepackage{graphicx}
\usepackage{amsmath}
\usepackage{amssymb}
\usepackage{multirow}
\usepackage{booktabs}
\usepackage{subfig}
\usepackage{threeparttable}
\usepackage{caption}

\usepackage[breaklinks=true,bookmarks=false]{hyperref}

\iccvfinalcopy % *** Uncomment this line for the final submission

\def\iccvPaperID{8619} % *** Enter the ICCV Paper ID here

\ificcvfinal\fi

\begin{document}

%%%%%%%%% TITLE
\title{3DGen: Triplane Latent Diffusion for Textured Mesh Generation}

\author{
Anchit Gupta\footnotemark[1]\thanks{\,Equal contributions}\qquad
Wenhan Xiong\footnotemark[1]\qquad
Yixin Nie\footnotemark[1] \qquad
Ian Jones\footnotemark[1] \qquad
Barlas O\u{g}uz \\
Meta AI\\
{\tt\small \{anchit, xwhan, ynie, ijones, barlaso\}@meta.com}
}
\maketitle
\ificcvfinal\thispagestyle{empty}\fi
%%%%%%%%% ABSTRACT
\begin{abstract}
Latent diffusion models for image generation have crossed a quality threshold which enabled them to achieve mass adoption.  Recently, a series of works have made advancements towards replicating this success in the 3D domain, introducing techniques such as point cloud VAE, triplane representation, neural implicit surfaces and differentiable rendering based training.  We take another step along this direction, combining these developments in a two-step pipeline consisting of 1) a triplane VAE which can learn latent representations of textured meshes and 2) a conditional diffusion model which generates the triplane features. For the first time this architecture allows conditional and unconditional generation of high quality textured or untextured 3D meshes across multiple diverse categories in a few seconds on a single GPU. It outperforms previous work substantially on image-conditioned and unconditional generation on mesh quality as well as texture generation.  Furthermore, we demonstrate the scalability of our model to large datasets for increased quality and diversity. \footnote{We will release our code and trained models.}
\end{abstract}

%%%%%%%%% BODY TEXT
\section{Introduction}
Following the success and popular adoption of large scale pre-trained image generation models~\cite{dhariwal2021diffusion,dalle2,stablediffusion}, there has been an increasing effort to replicate these capabilities in the 3D domain. A series of works made advances in architectural and representation improvements for 3D generation. Point Cloud variational autoencoders (VAE) for neural implicit fields~\cite{3dilg} allow learning latent representations of 3D geometry across many object categories, ultimately enabling controllable generation of 3D meshes.  High resolution tetrahedral grids~\cite{dmtet} and triplane diffusion models~\cite{NFD,rodin} are suitable for learning high quality mesh representations.  Rendering based losses and GAN training were proposed in \cite{get3d} to enable generating textures along with mesh generation.  

\begin{figure*}[t]
    \centering
    \includegraphics[width=0.95\linewidth]{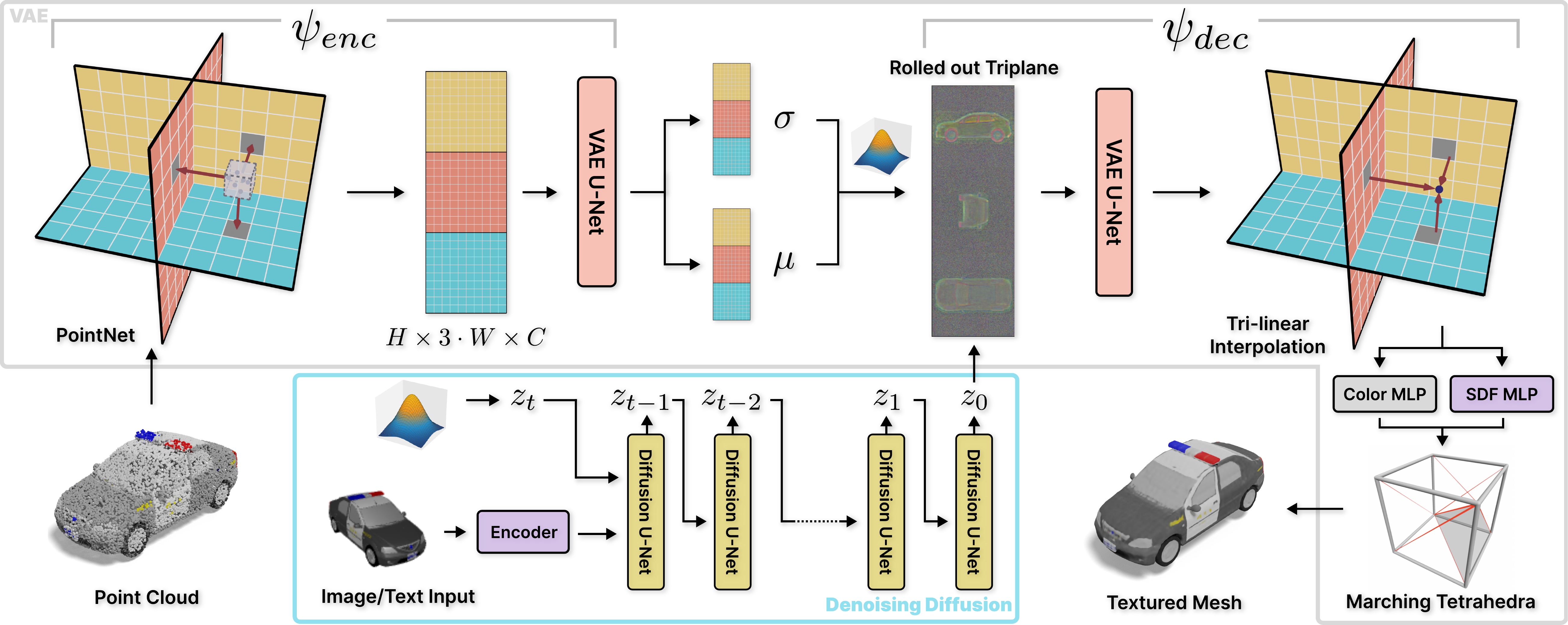}\vspace{-5pt}
    \caption{Our two stage model pipeline combines a triplane VAE ($\psi_{enc}, \psi_{dec}$) along with a conditional diffusion model.}
    \label{fig:model}
\end{figure*}

Each of these techniques have clearly been useful in isolation.  However, it's unclear how they should come together to eventually enable high quality joint mesh and texture generation in a scalable and efficient way.  For instance existing VAE methods like~\cite{3dilg} have shown scalability to multiple categories, yet they learn discrete representations which are not necessarily compatible with latent diffusion methods that promise higher quality.  \cite{get3d,rodin,NFD} are single category models, which can't benefit from large scale pre-training.  \cite{NFD} does not handle color at all.  In all, it's an open question how to scalably train good latent representations for 3D, how to best represent and learn geometry and color  jointly, and how to achieve these while democratizing it with practical computational constraints.  In this work, we attempt to answer these questions.

Our proposed 3DGen architecture consists of two stages.  In the first stage, we train a Variational Autoencoder that first encodes an optionally colored point cloud sampled from an input mesh into a triplane Gaussian latent space and then learns to reconstruct the continuous textured mesh from this latent.  In the second stage, we train a diffusion model to generate the triplane features, which can be conditioned on an input image-text embedding. This allows us to perform image-conditioned, text-conditioned, and unconditional generation.  In both stages, the triplane features are decoded into a differentiable colored mesh representation which allows training with rendering based losses.  This architecture is pictured in Figure~\ref{fig:model}.  The many careful design choices and implementation details to get this novel architecture to work well are documented in section \ref{sec:method}.

In section \ref{sec:exp} we compare 3DGen to previous works on geometry-only and textured mesh generation in the unconditional as well as the image-conditioned settings.  Previous SoTA models differ for each of these categories, since no one model has previously incorporated all of these capabilities, and we observe substantial improvements in all settings.  For unconditional geometry generation, 3DGen improves FiD scores by 23\% over NFD~\cite{NFD}, the closest competitor.  For text-conditioned geometry generation, improvements range from 15-20\% based on the category, compared to the SoTA 3DILG~\cite{3dilg} model. For unconditional colored mesh generation, we compare to GET3D~\cite{get3d}, and observe FiD score improvements up to 70\%.

Finally, we scale our model to a recent dataset of almost half a million 3D objects~\cite{objaverse} (section \ref{sec:objaverse}), showing significant improvements from pre-training.  To summarize, 3DGen is a highly performant, scalable and novel architecture, which we believe takes us one step closer to a practical, high-quality 3D object generation model that can be widely adopted by practitioners.  

\section{Method}\label{sec:method}
\begin{figure*}[t]
    \centering
    \includegraphics[width=0.96\linewidth]{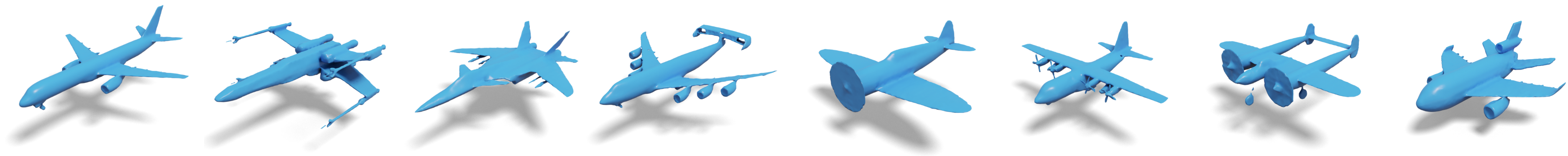}\vspace{-2.5pt}
    \includegraphics[width=0.96\linewidth]{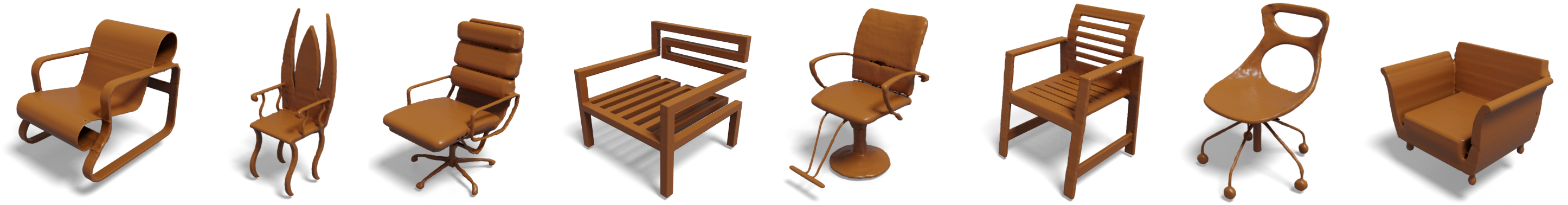}\vspace{-5pt}
    \caption{Single category unconditional mesh generation. Our model captures fine details and generates diverse outputs}
    \label{fig:uncond_examples}
\end{figure*}
\subsection{Background}
\textbf{Neural Fields}: Several different approaches have been proposed to represent 3D objects/scenes implicitly using neural fields. In our work, we use the triplane representation \cite{eg3d} which strikes an appealing balance between efficiency and quality. They represent a scene using 3 axis aligned feature planes as shown in Figure~\ref{fig:model}. Volumetric rendering is used to render images from this neural field, by combining trilinear feature interpolation with a MLP decoder.
Subsequent works such as \cite{get3d, NFD} have validated the efficacy of triplane neural fields in representing 3D objects.
%such as from ShapeNet.

Unlike EG3D\cite{eg3d} which used volumetric rendering, GET3D\cite{get3d} instead uses a tetrahedral grid along with the DMTet algorithm \cite{dmtet} to directly output a triangulated mesh. This removes the need for the slow per-image volumetric rendering operation and gives us meshes which can be directly used in downstream applications such as game engines. We adopt the mesh decoding procedure from GET3D and point readers to it for more details.

\textbf{Diffusion Models}:
Diffusion models are a class of generative models that are trained to reverse a Markovian forward process. Given a sample $z_0$ drawn from the data distribution $p(z)$, the forward process of denoising diffusion models yields a sequence of noised data $\{z_t|t\in(0,T)\}$ with $z_t = \alpha_tz_0 +\sigma_t\epsilon$, where $\epsilon$ is random normal noise drawn from $\mathcal{N}(0,1)$. The noise schedule of this forward process is defined by the fixed sequence $\alpha_t, \sigma_t$. A large enough $T$ approximately turns  $z_T$ into random noise drawn from $\mathcal{N}(0,1)$. The diffusion model is trained to reverse this process, i.e recovering $z_{t-1}$ from $z_{t}$ by predicting the added random noise $\epsilon$. Many further modifications have been proposed to improve this base process including re-parameterizations, training losses, noise schedules and ancestral sampling approaches for fast inference. We point readers to \cite{stablediffusion, imagen} for more details about image diffusion models which our model closely follows.

\subsection{Triplane VAE}\label{sec:vae}

The first component of our method is a variational autoencoder (VAE)~\cite{vae} that learns latent representations of 3D meshes (see the upper part of Figure~\ref{fig:model}). 
% As discussed earlier, we u  se a triplane representation as our latent code. 
%Triplanes offer a balance between computational efficiency when compared with full 3d grids while maintaining reconstruction quality.

The encoder $\psi_{enc}$ of the VAE models point clouds $x \in \mathbb{R}^{3 \times N}$ into a distribution of triplane latent features $q_{\psi_{enc}}(z|x)$ where $z=h_{xz},h_{xy},h_{yz}\in\mathbb{R}^{H \times W \times C}$ and each 2D plane has a resolution of $H \times W$ and $C$ channels. The architecture of $\psi_{enc}$ closely resembles that from \cite{gensdf} and consists of a PointNet followed by a UNet to further refine the triplanes. Differing from \cite{gensdf}, we add position embeddings in the PointNet and our UNet uses 3D aware convolutions and rolled out triplanes as introduced by \cite{rodin}.

To decode a mesh, the decoder $\psi_{dec}$ takes as inputs the coordinates of all vertices on a tetrahedral grid ($v_i \in V_T$) and then predict the SDF value $s_i$ and the vertice deformation $\Delta v_i$ based on each vertice's triplane features. The SDF and deformation values $(s_i, \Delta v_i )$ are used to form a triangulated mesh using the differentiable marching tetrahedra algorithm. Specifically, our decoder $\psi_{dec}$ first refines the triplane latent $z$ using a UNet to obtain $z'$ and then gather the triplane features at the 3 projected locations, which are concatenated and fed into a MLP head to predict $s_i$ and $\Delta v_i$. The MLP head consists of linear blocks with a residual connection, layer normalization and ReLU activation.

We train the VAE using a rendering based reconstruction loss. This is allowed by a differentiable renderer~\cite{nvdiffrast} and the differentiable DMTet algorithm. Compared to previous methods~\cite{occnet,3dilg,NFD} that use SDF or occupancy regression losses, using a rendering loss only requires minimal pre-processing of the meshes and preserves fine details lost by pre-processing algorithms like watertighting used to obtain SDF/occupancy values. The mesh output from the decoder is passed to a differentiable renderer to obtain a mask silhouette $m_x$ and depth map $d_x$ from randomly sampled camera angles. We use the ground truth mesh renders $m^{gt}_x, d^{gt}_x$ to supervise the model. We add a KL divergence loss to ensure the triplane feature distribution $q_{\psi_{enc}}(z|x)$ is close to a gaussian prior $p(h)=\mathcal{N}(0,1)$. Additionally we add a laplacian mesh smoothing loss to improve smoothness. We hence train the model with a combined loss $L_{vae} = ||m_x-m^{gt}_x||^2_2 + ||d_x-d^{gt}_x||_1 + \lambda L_{smooth} - \gamma D_{KL}(q_{\psi_{enc}}(z|x)|p(h))$

\subsection{Texture Prediction}
As the generated mesh can have an arbitrary topology, developing models that can generate both shapes and semantically consistent textures is non-trivial and an underexplored problem. To extend our Triplane VAE framework for texture prediction, we modify the encoder to encode colored point clouds and parameterize the surface color as a texture field ~\cite{oechsle2019texture} in the decoder. Specifically, the encoder $\psi_{enc}$ takes a point cloud with an additional $m$-channel color embedding as input $x \in \mathbb{R}^{(3+m) \times N}$. The color embedding is obtained by applying an affine layer on the normalized RGB values of the input point cloud. In the decoder we add a color prediction MLP similar to the SDF MLP discussed above, that maps the 3D location of a surface point and interpolated triplane features to a RGB value $c_x \in \mathbb{R}^3$. For training a textured VAE, in addition to the loss used to train the geometric only model, we further query the model for colors at the mesh surface and calculate a loss by comparing to ground truth surface colors $c^{gt}_{x}$ as $L_{color}=||c_x - c^{gt}_{x}||_1 + ||c_x - c^{gt}_{x}||_2^2$.

\subsection{Triplane Diffusion}
After obtaining a VAE with a well trained triplane latent space, we train a diffusion model to generate these features. Following \cite{rodin} we use rolled out triplanes $z\in \mathbb{R}^{H\times3W\times C}$, which are essentially 2D images allowing reuse of the recent work done in image diffusion models. Figure \ref{fig:model} shows a visualization of such a triplane. Our diffusion model consists of a standard UNet backbone to which we add 3D aware convolutions in the ResNet blocks. 3D aware convolutions introduced in \cite{rodin} use mean pooling to introduce cross-plane interactions, as any two of the planes share an axis such interactions improve 3D consistency and overall quality. We train this model with $1000$ denoising steps, use the $v$ prediction objective and a cosine noise schedule as in \cite{imagen}.

In case of image conditioned generation we embed the image $y$ using a pre-trained frozen image-text bi-encoder $\Upsilon$ to obtain the conditioning vector $\Upsilon(y)$. This is further passed through a layernorm and concatenated to the time-step embedding in the UNet. For computational efficiency and also to easily enable text guided applications we do not use cross attention based conditioning layers, instead relying on adaptive group normalization (AdaGN) to inject $\Upsilon(y)$ into each layer. Our conditioned model also utilizes classifier free guidance \cite{ho2021classifier} to boost sampling quality and while training we randomly zero out $20\%$ of the conditioning vectors.

\section{Experiments}\label{sec:exp}
\begin{figure*}[t]
    \centering
    \includegraphics[width=\linewidth]{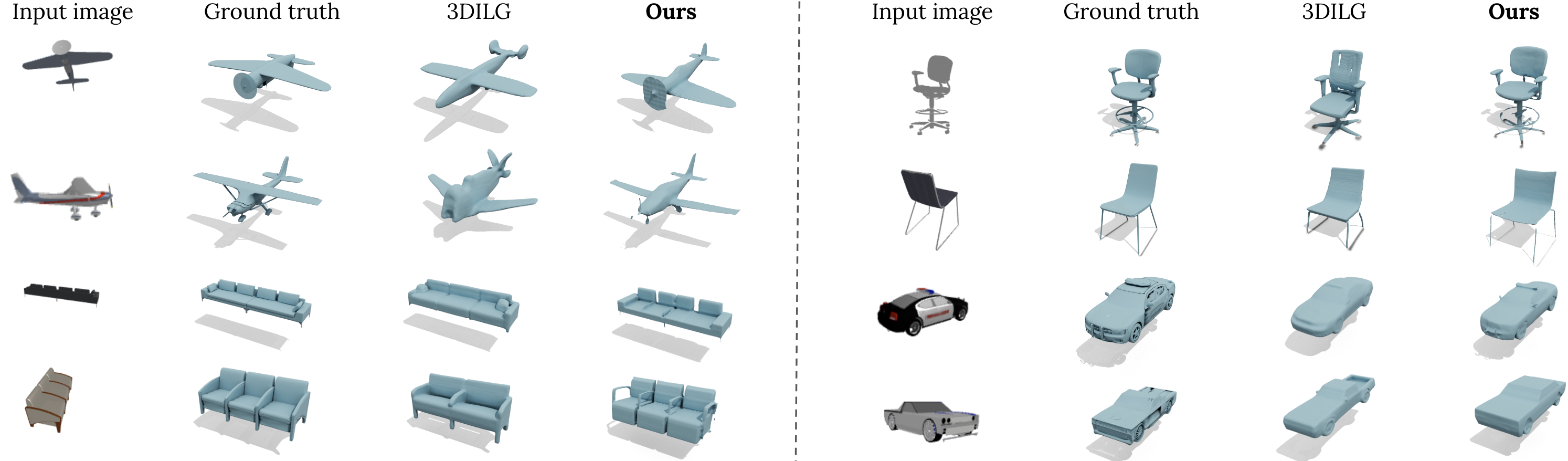}\vspace{-5pt}
    \caption{Comparison of image-conditioned generation between our model and 3DILG\cite{3dilg}}
    \label{fig:cond_examples}
\end{figure*}

\begin{table*}
\centering
    \begin{threeparttable}
    \begin{tabular}{l c cc c cc}
    \toprule
    
    && \multicolumn{2}{c}{Head Categories} && \multicolumn{2}{c}{Tail Categories} \\
    \cmidrule{3-4} \cmidrule{6-7} 
    Model && Chamfer-L1 ($\downarrow$) & Shading-FiD ($\downarrow$) &&  Chamfer-L1 ($\downarrow$) & Shading-FiD ($\downarrow$)  \\
    \midrule
    CLIP-Forge\tnote{$\dagger$} && 0.244 & 188.89 && - & -\\
    3DILG && 0.219 & 93.56 && 0.244 & 100.59 \\
    3DGen && 0.181 & 80.31 && 0.192 & 94.88\\
    3DGen + pretraining && \textbf{0.172} & \textbf{76.44} && \textbf{0.184} & \textbf{85.92} \\
    \bottomrule
    \end{tabular}
    
    \begin{tablenotes}
    \small \item[$\dagger$] For CLIP-Forge, we directly use released checkpoints which are only trained for the 13 head categories. 
    \end{tablenotes}
    
    \end{threeparttable}
\caption{Image-conditioned mesh generation. Average Chamfer-distance and shading FiD across head (13 most common categories) and tail categories.}
\label{tab:cond_gen}
\end{table*}
\subsection{Data}
\label{sec:data}
For most of our experiments we use 3D objects from the subset of ShapeNetCore~\cite{shapenet} version 2 originating from 3D Warehouse (Trimble Inc).
Using the standard data splits, this in total gives us about 45k objects. 
For our scaling experiments we use the recent Objaverse \cite{objaverse} dataset which contains around 750k assets. We filter out meshes from categories like 3D scans, people, point clouds and also ones with more than 100k vertices, leaving around 450k objects. We normalize each mesh to lie within $[-0.95,0.95]^3$ to remove scale differences by objects being too big/small. 

To train the textured mesh generation model, we obtain colored point clouds following the procedure in Point-E~\cite{nichol2022point}. We first render each 3D model with depth using Blender from uniformly sampled camera angles, then un-project the pixels back to their original 3D locations, and finally combined to give a dense set of 1M pre-computed points representing the surface and texture of the input mesh. During training, the input and reference points are sampled from this pre-computed set. Since we want the point cloud to represent the intrinsic color of the object surface, we render with a uniform sphere light to prevent uneven shadows and lighting effects. This produces flat textures that can later be lit accordingly in downstream applications.

\subsection{VAE training}
To train the VAE, we randomly sample 20k points from the surface of the meshes as inputs and the training is supervised by rendering $1024\times1024$ depth and mask images from $8$ randomly sampled camera angles per object. We use a latent dimension $C=32$ and a triplane resolution $H=W=256$. While training we randomly down-sample the triplane to a resolution in $[64,256]$ and then upsample to promote robust representations and enable flexibility in our diffusion model. 
In total this model has 20M parameters and is trained with the AdamW optimizer and linear learning rate decay. We first train until convergence with a lower tetrahedral grid resolution of $90$. Subsequently we freeze the encoder and finetune only the decoder but with a higher resolution of $128$, this is discussed more in Section \ref{sec:staged_vae}. We train this model using 8 A100 GPUs over 2 days.
To validate the efficacy of our VAE, we report the reconstruction error metrics.
% in Table \ref{tab:vae}. 
Specifically, we compute the Chamfer distance
\footnote{\href{https://pytorch3d.readthedocs.io/en/latest/_modules/pytorch3d/loss/chamfer.html}{Pytorch3D Chamfer Distance}} and intersection of union (IoU) of the rendered masks and compare to \cite{3dilg}. As shown in Table \ref{tab:vae} our VAE outperforms 3DILG and achieves even stronger reconstruction accuracy, more analysis can be found in Section \ref{sec:vae_variants}.

\subsection{Conditional generation}
We render each object in our dataset from random camera angles and use the rendered images as a conditioning inputs to the diffusion model. A large frozen image-text bi-encoder model similar to \cite{openclip} is used to embed these images. Our diffusion UNet has about 240M parameters and we train it with the AdamW optimizer with cosine learning rate decay. For efficiency we use a downsampled triplane resolution of $64$ for diffusion and scale the triplanes by a factor of $0.1$. This is trained on 16 A100 GPUs for about 4 days. During inference we use a guidance scale of $5$ and use the fast DEIS \cite{deis} sampler with $50$ sampling steps.

 We compare our model with two previous methods using different latent representations: CLIP-forge~\cite{clip_forge} which uses a single latent vector and trains a normalizing flow model to generate it; 3DILG~\cite{3dilg} which learns a sequence of discrete tokens as the latents and trains a transformer decoder as the generator. As 3DILG uses a different version of the 3D dataset, we retrain 3DILG on our data splits. For CLIP-forge and 3DILG, we generate the meshes by running marching cubes~\footnote{\url{https://pypi.org/project/PyMCubes/}} on a $128^3$ grid. Similarly, we generate meshes using the marching tetrahedra algorithm~\cite{dmtet} with the same grid resolution. Note that all the models are trained with all-category data.

As shown in Table \ref{tab:cond_gen}, our method achieves improvements in both the alignment with the groundtruth (as measured by Chamfer distance) and the overall mesh quality (as measured by the Fréchet Inception Distance on shading images~\cite{sdfstylegan} rendered from the generated and groundtruth meshes). Figure \ref{fig:cond_examples} shows qualitative comparisons between our model and 3DILG. We can see that our model can generate meshes that are more faithful to the input images and is also better at generating geometry details. 
    \begin{figure}[t]
        \includegraphics[width=\linewidth]{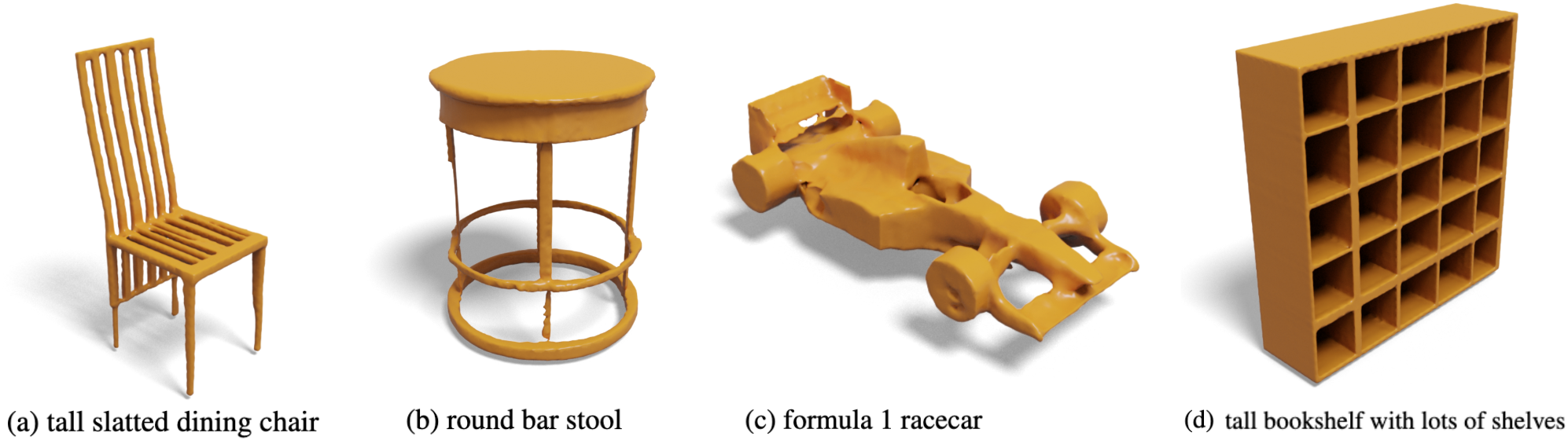}
        \caption{Text guided generation (prompts above)}
        \label{fig:textcond_examples}
    \end{figure}
% \begin{figure*}
%     \centering
%     \includegraphics[width=0.8\linewidth]{iccv2023AuthorKit/figures/uncond_textured.png}
%     \caption{Generated meshes from our single category textured mesh generation model}
%     \label{fig:uncond_textured}
% \end{figure*}

\begin{figure*}[t]
    \centering
    \includegraphics[width=0.12\linewidth,trim={380 380 380 380},clip]{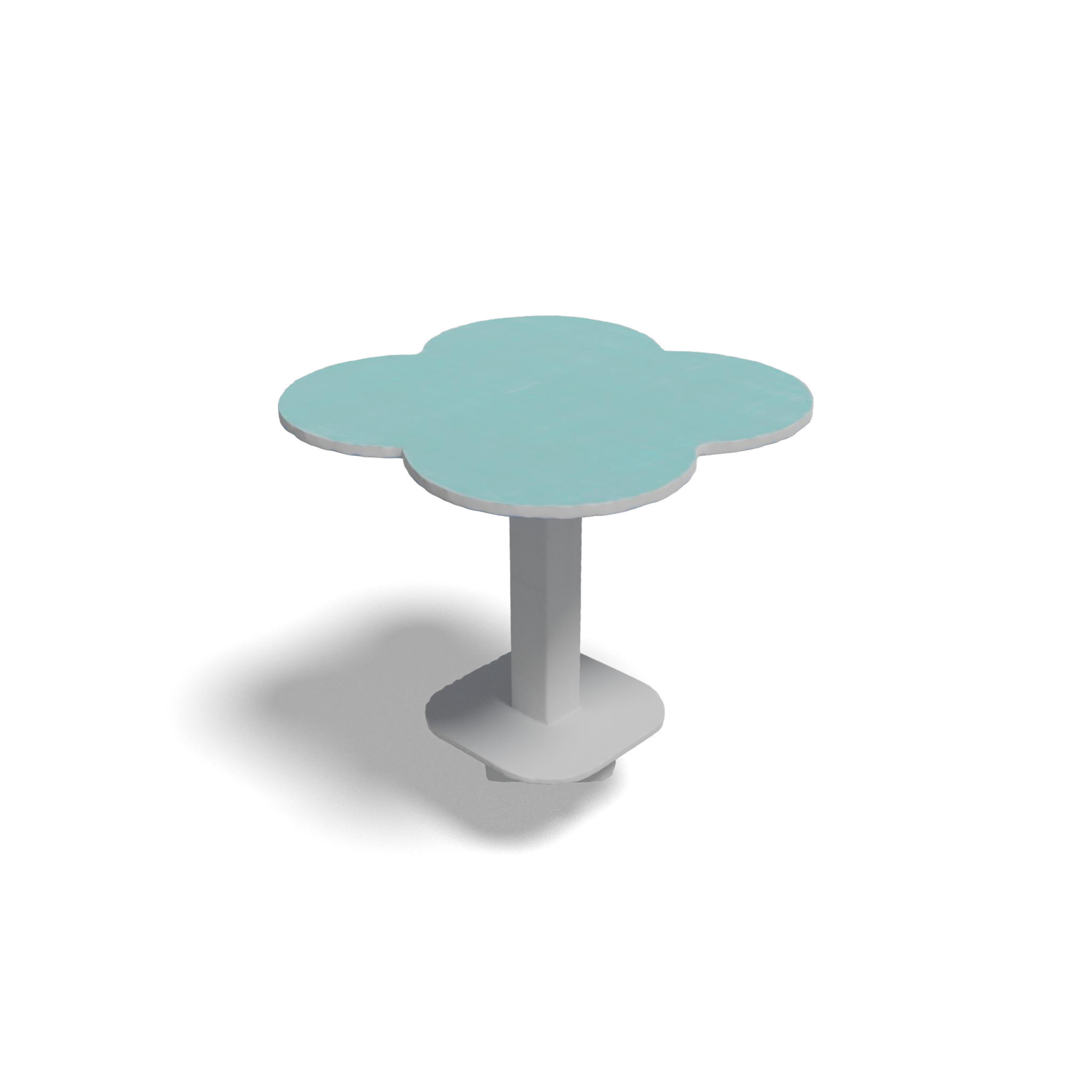}
    \includegraphics[width=0.12\linewidth,trim={380 380 380 380},clip]{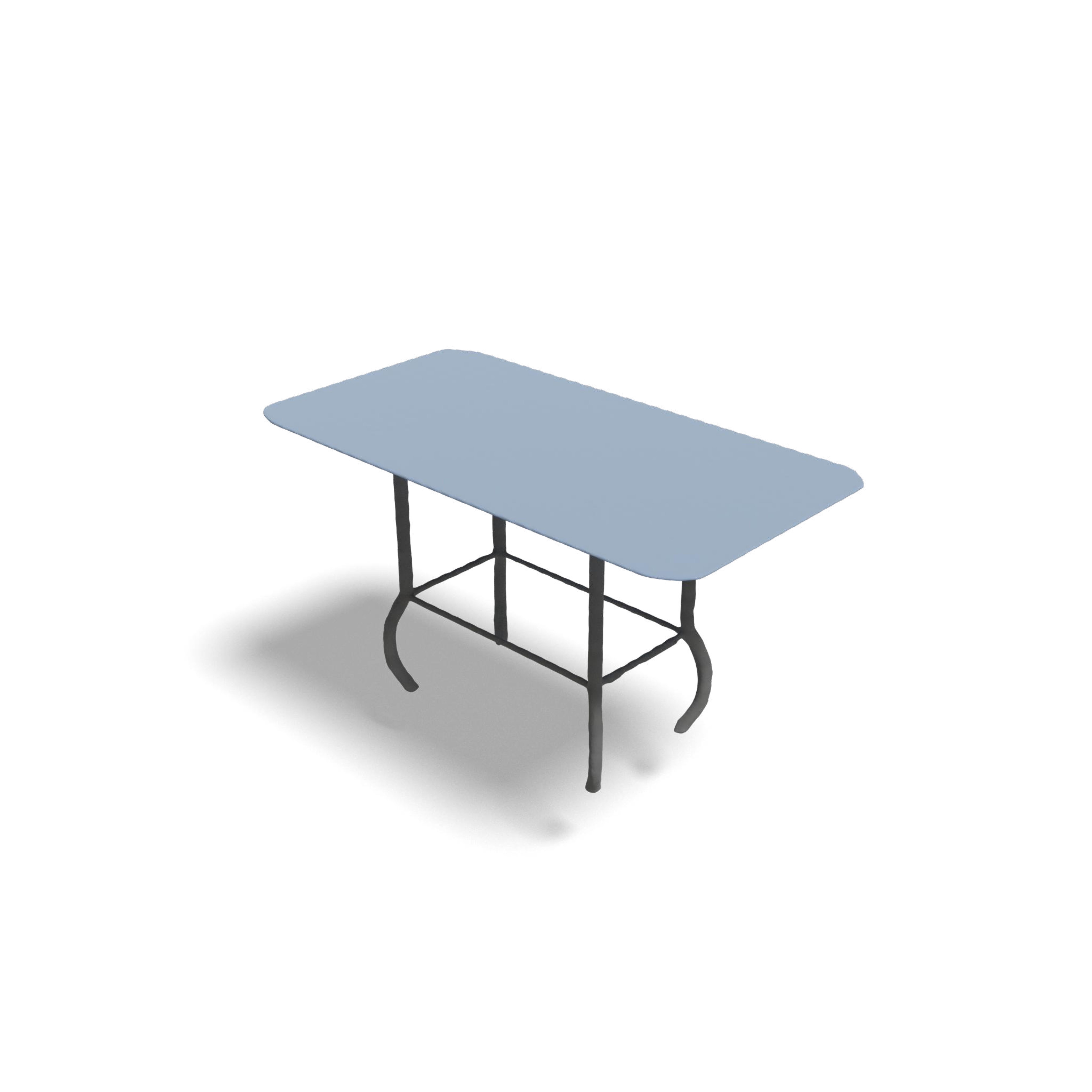}
    \includegraphics[width=0.12\linewidth,trim={380 380 380 380},clip]{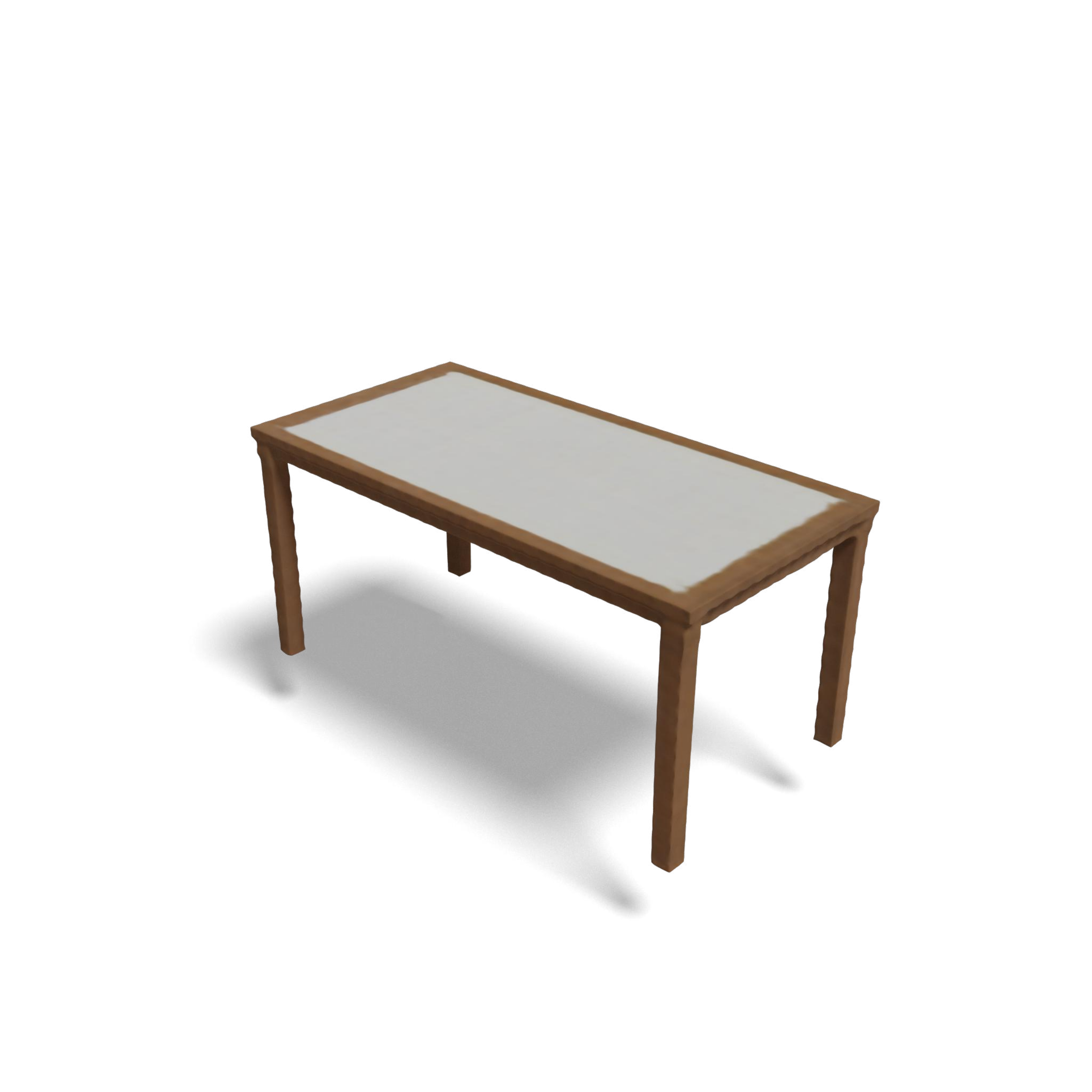}
    \includegraphics[width=0.12\linewidth,trim={380 380 380 380},clip]{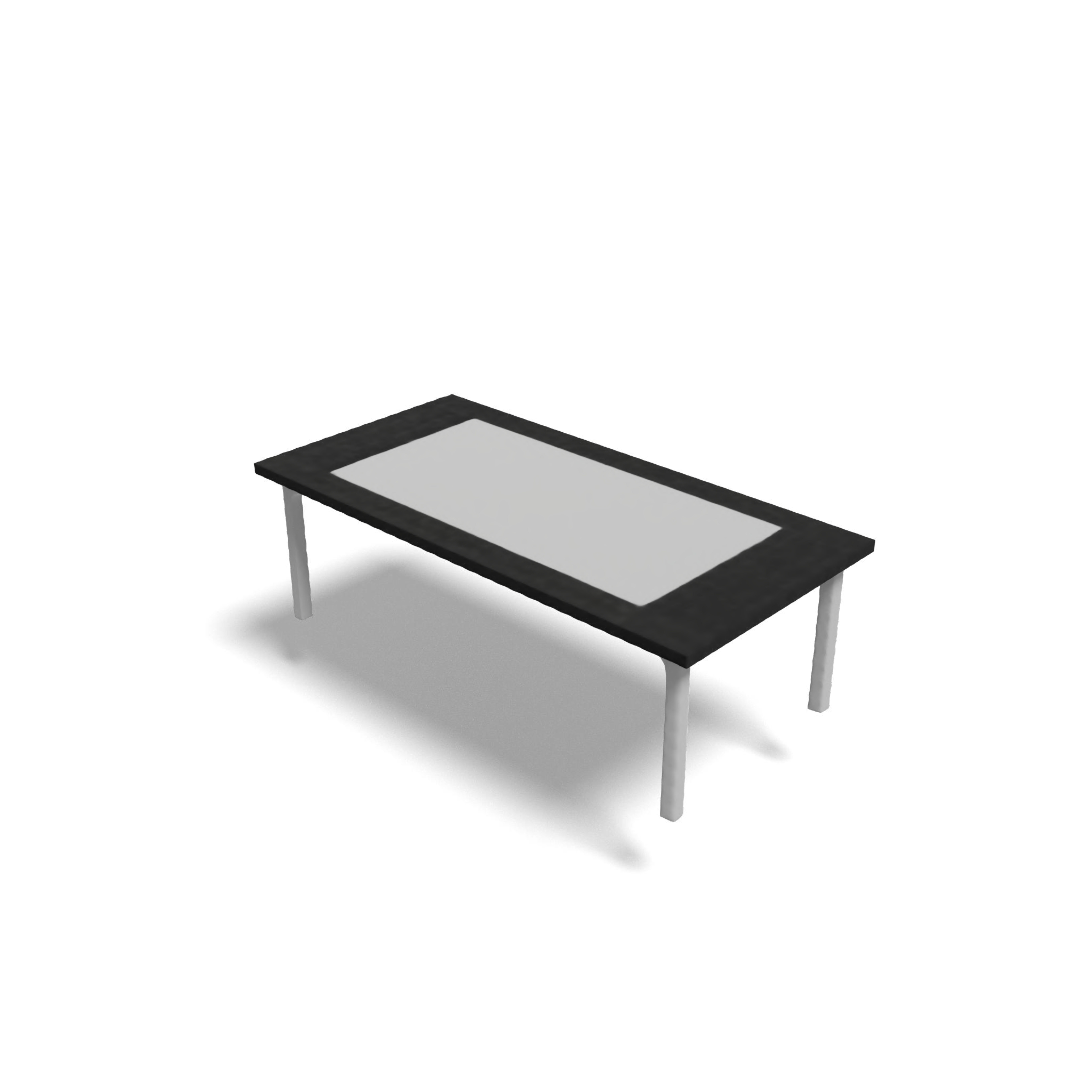}
    \includegraphics[width=0.12\linewidth,trim={380 380 380 380},clip]{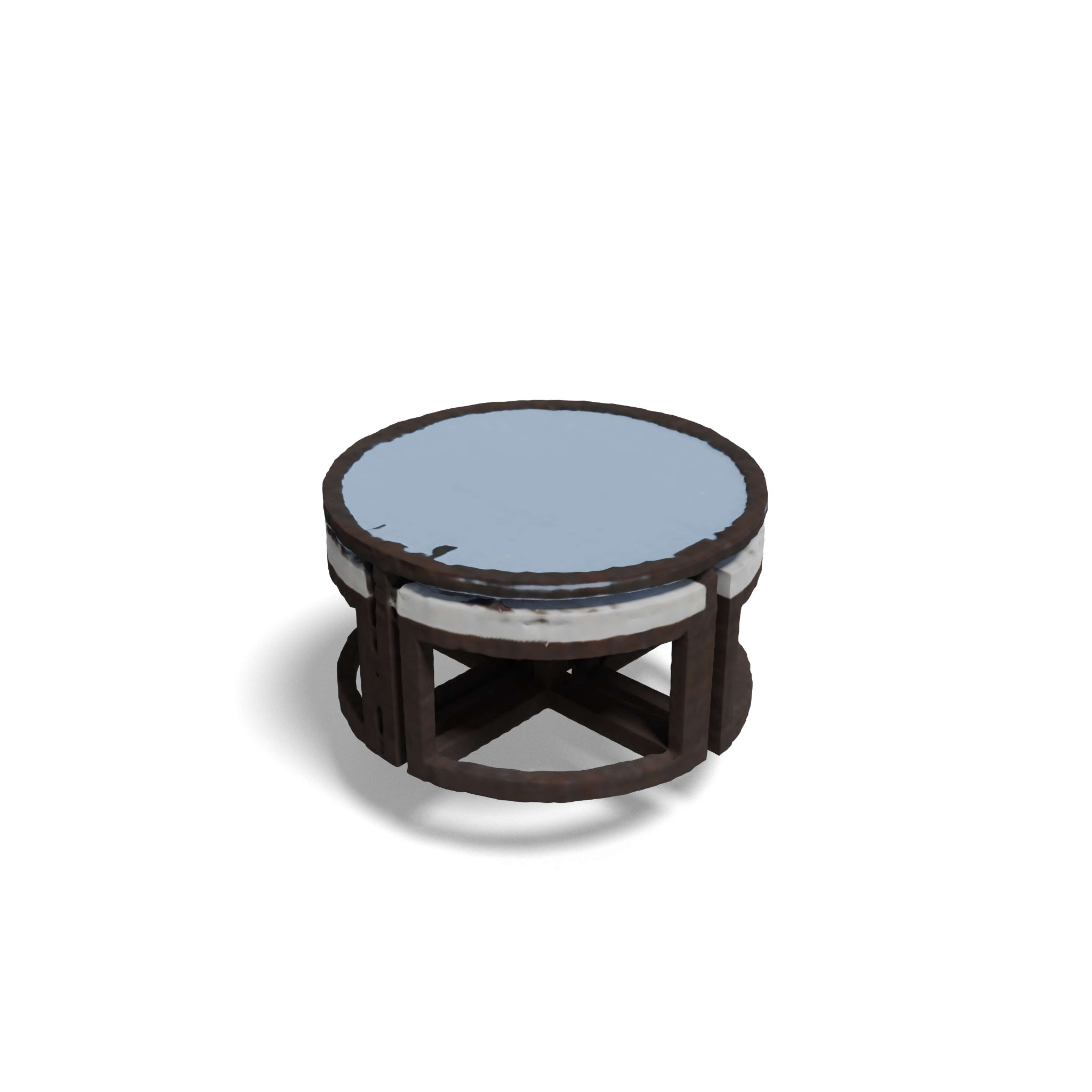}
    \includegraphics[width=0.12\linewidth,trim={380 380 380 380},clip]{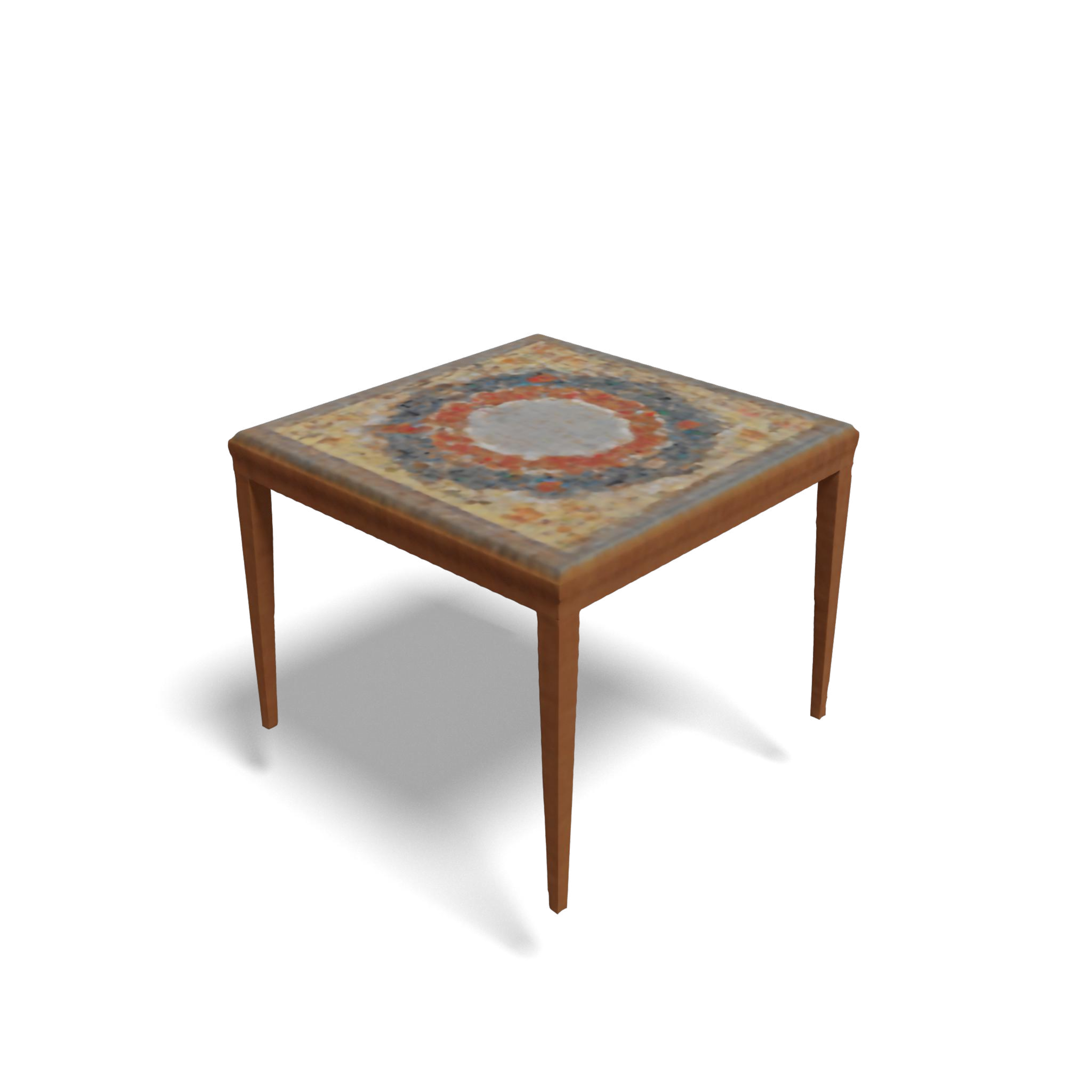}
    \includegraphics[width=0.12\linewidth,trim={380 380 380 380},clip]{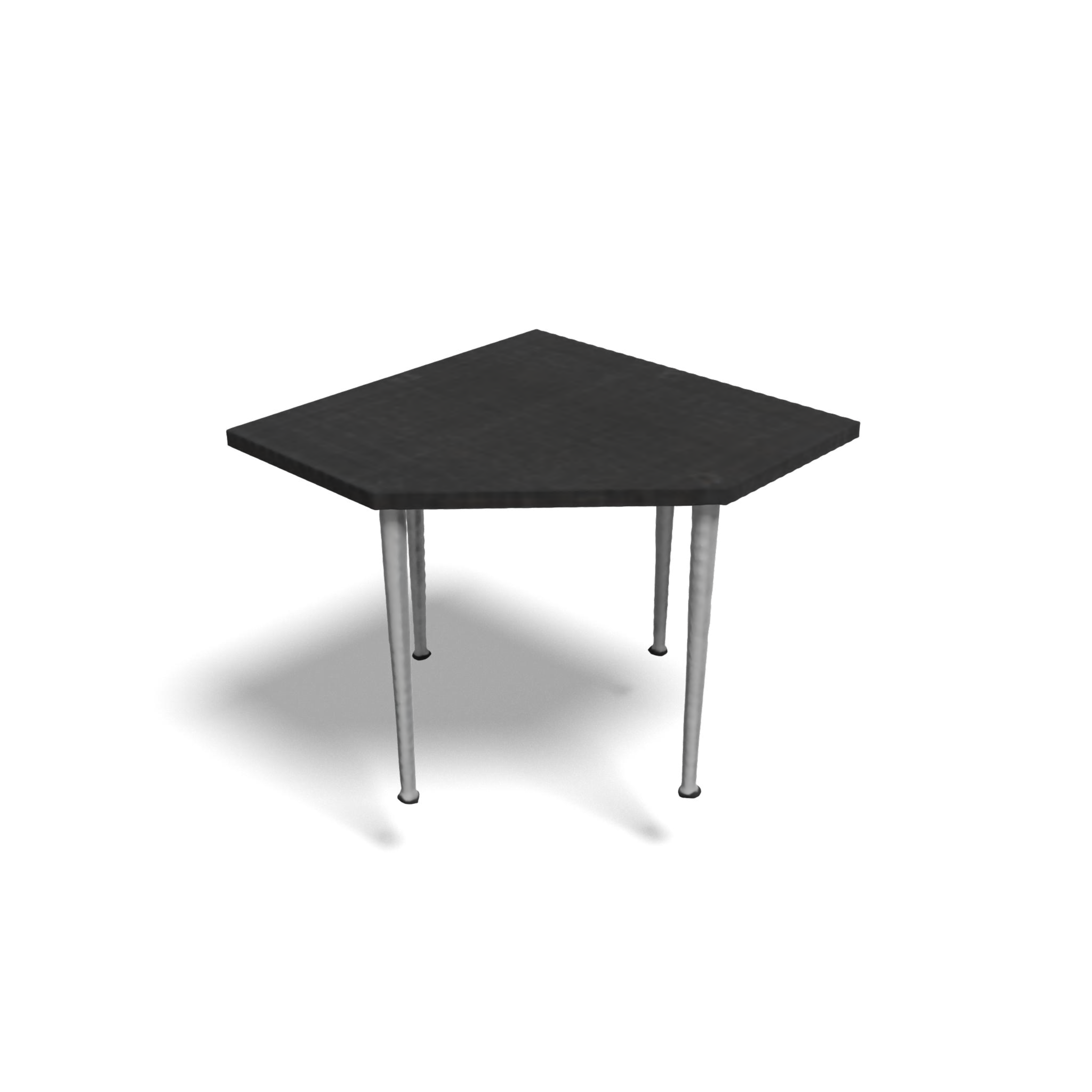}
    \includegraphics[width=0.12\linewidth,trim={380 380 380 380},clip]{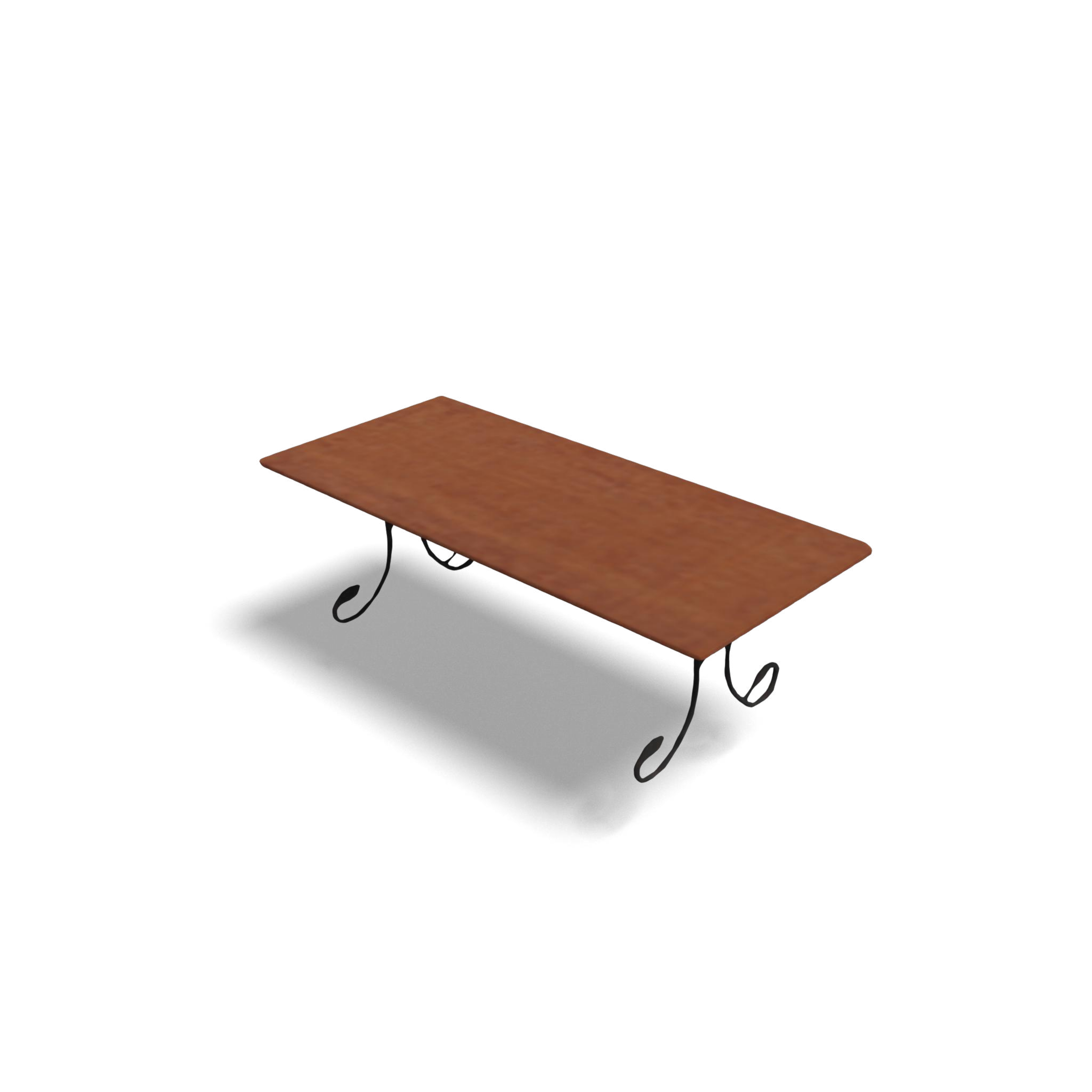}\\
    \vspace{-7pt}
    \includegraphics[width=0.12\linewidth,trim={380 380 380 380},clip]{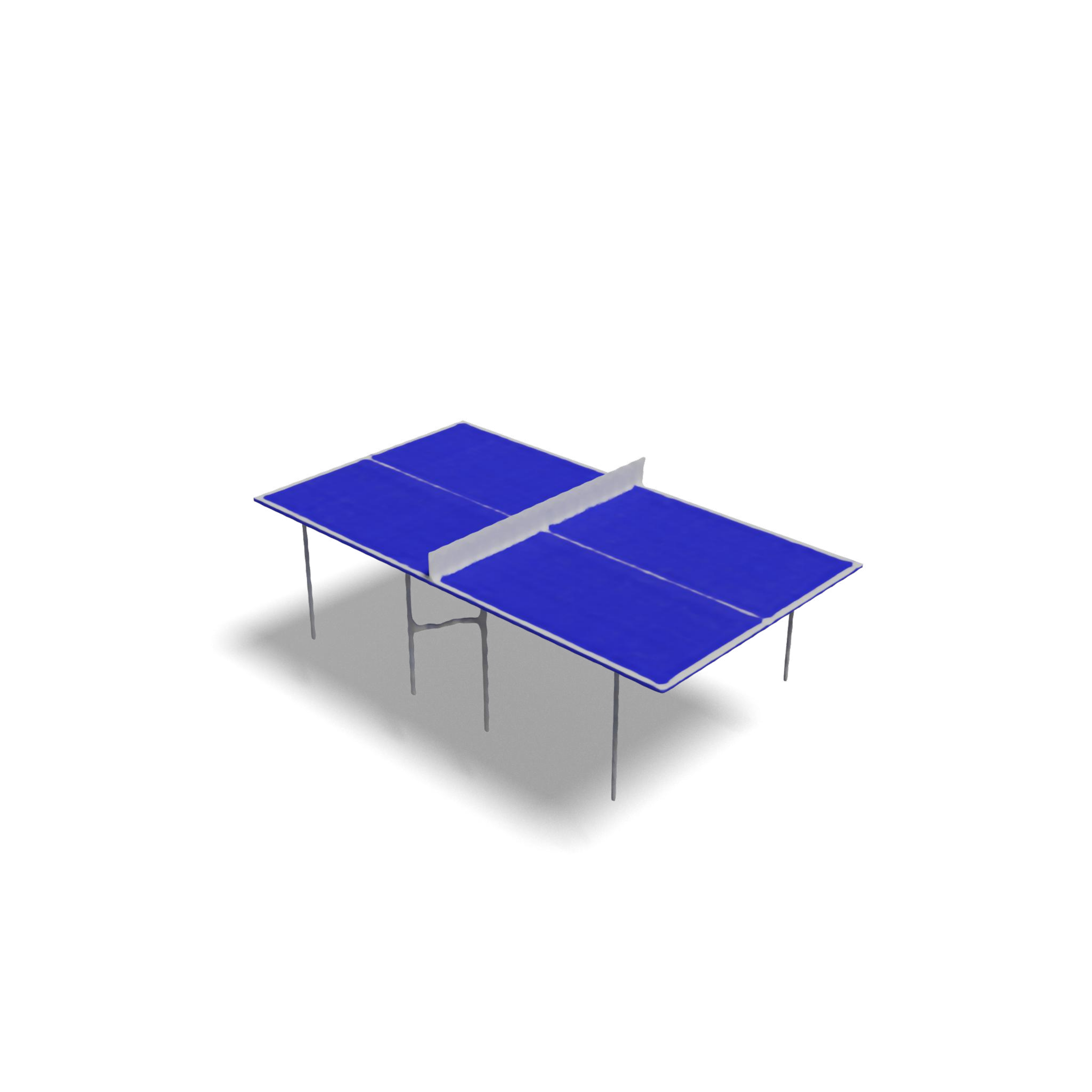}
    \includegraphics[width=0.12\linewidth,trim={380 380 380 380},clip]{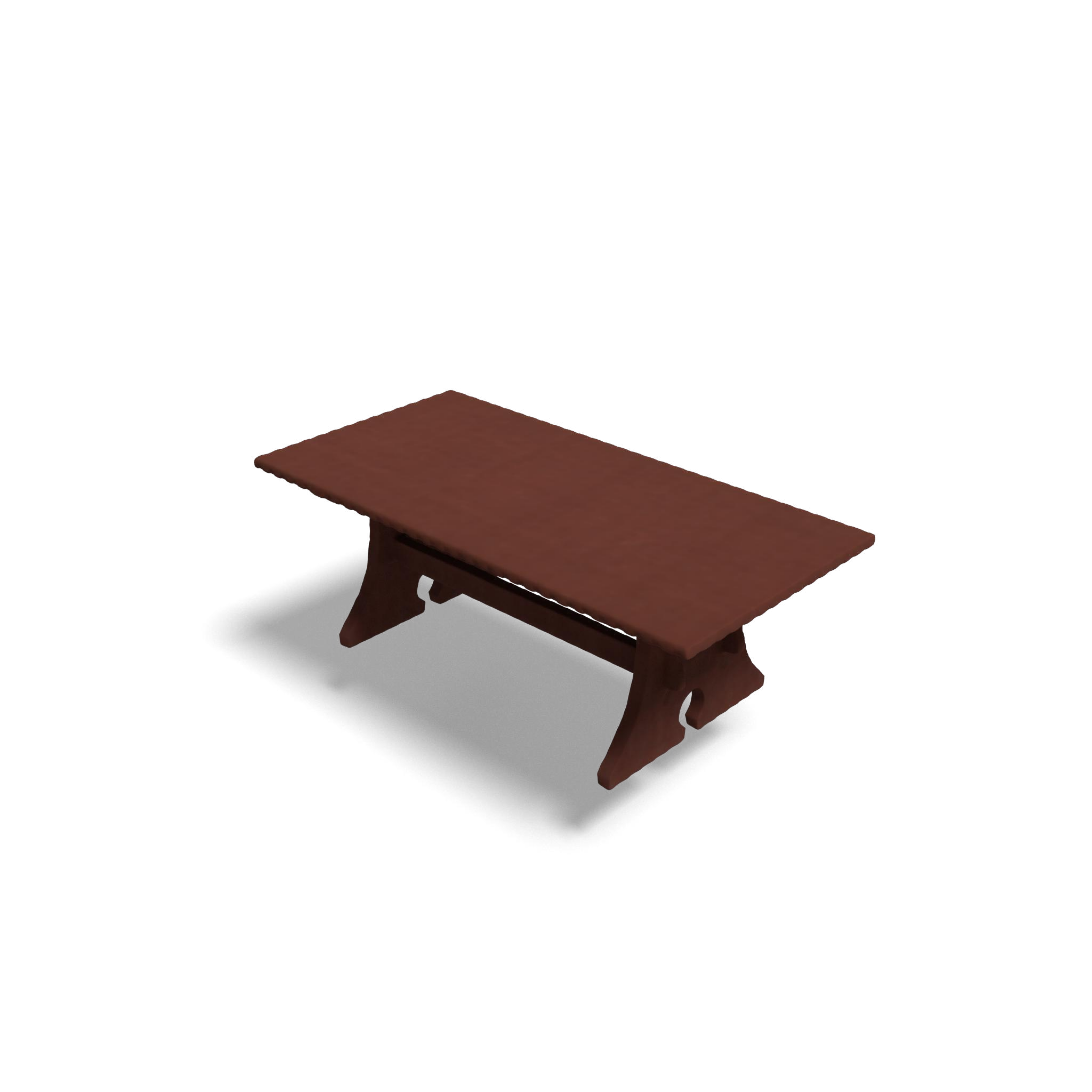}
    \includegraphics[width=0.12\linewidth,trim={280 280 280 280},clip]{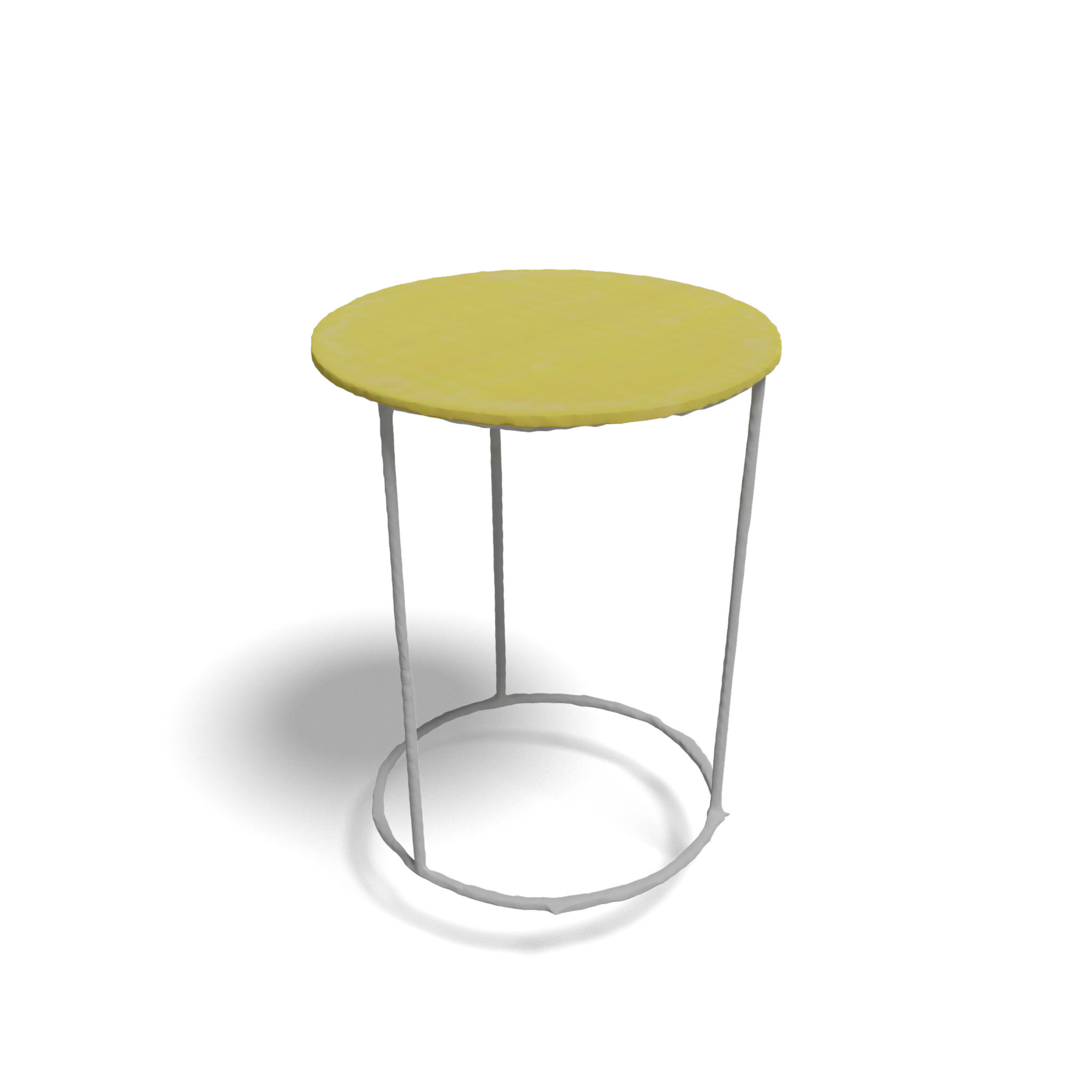}
    \includegraphics[width=0.12\linewidth,trim={380 380 380 380},clip]{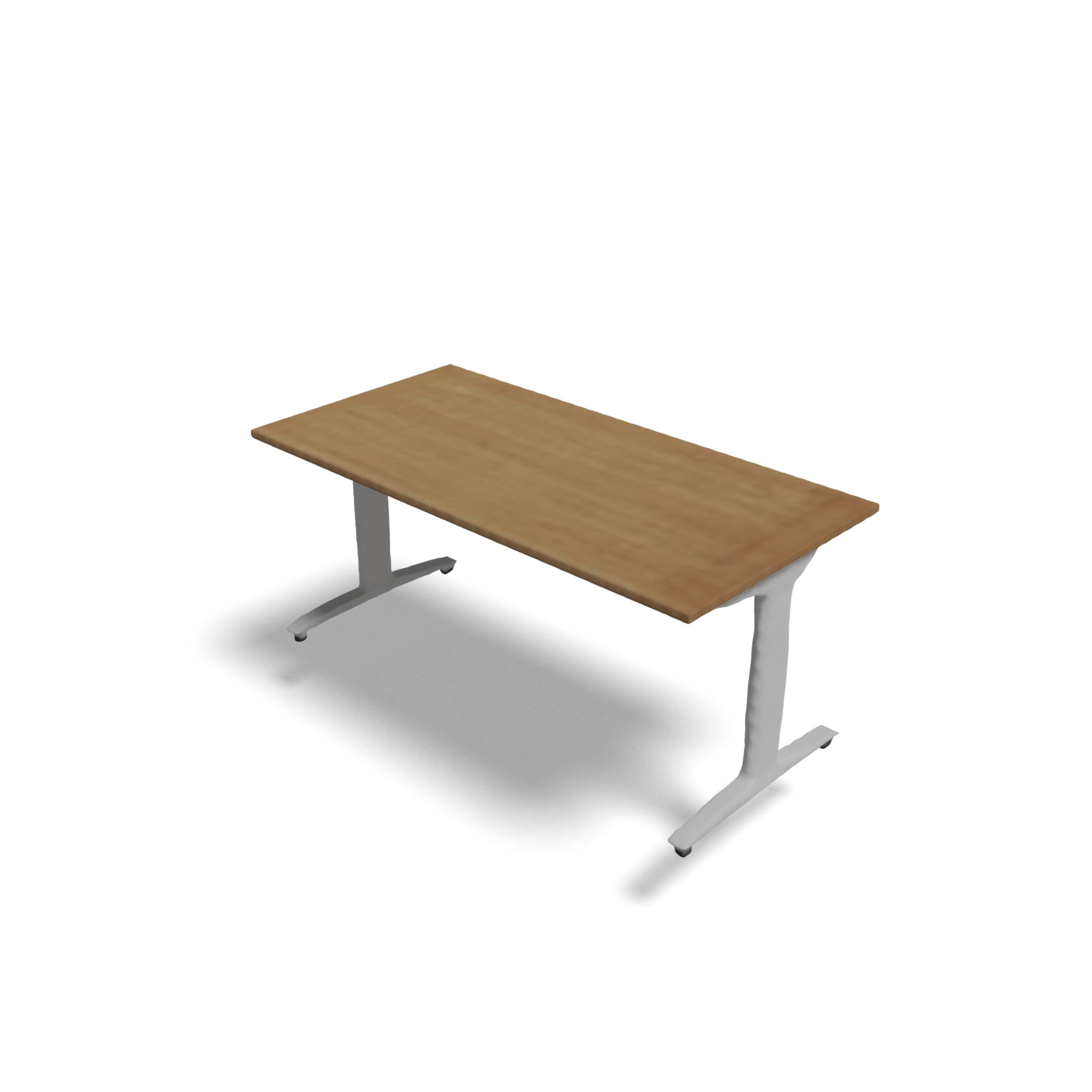}
    \includegraphics[width=0.12\linewidth,trim={380 380 380 380},clip]{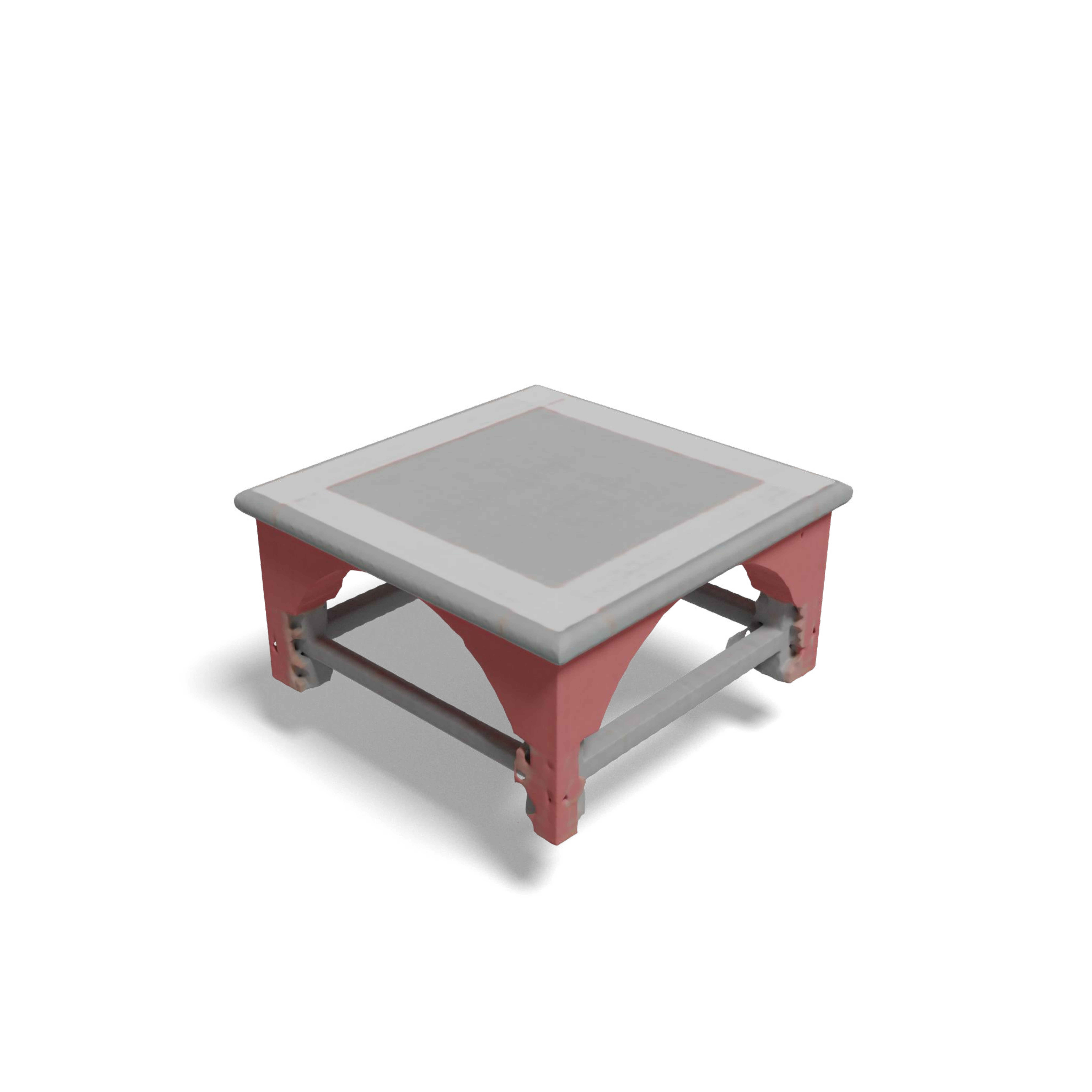}
    \includegraphics[width=0.12\linewidth,trim={380 380 380 380},clip]{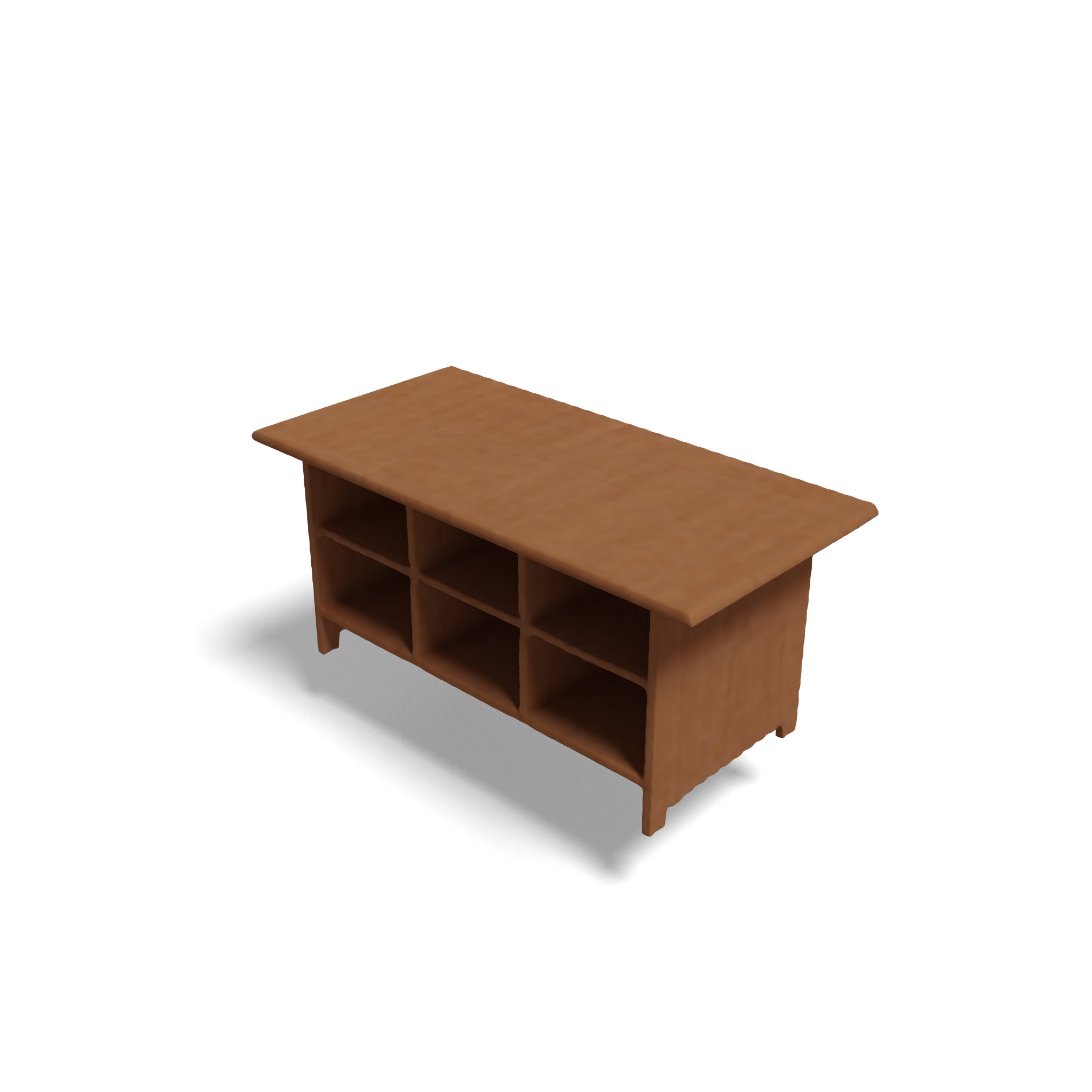}
    \includegraphics[width=0.12\linewidth,trim={380 380 380 380},clip]{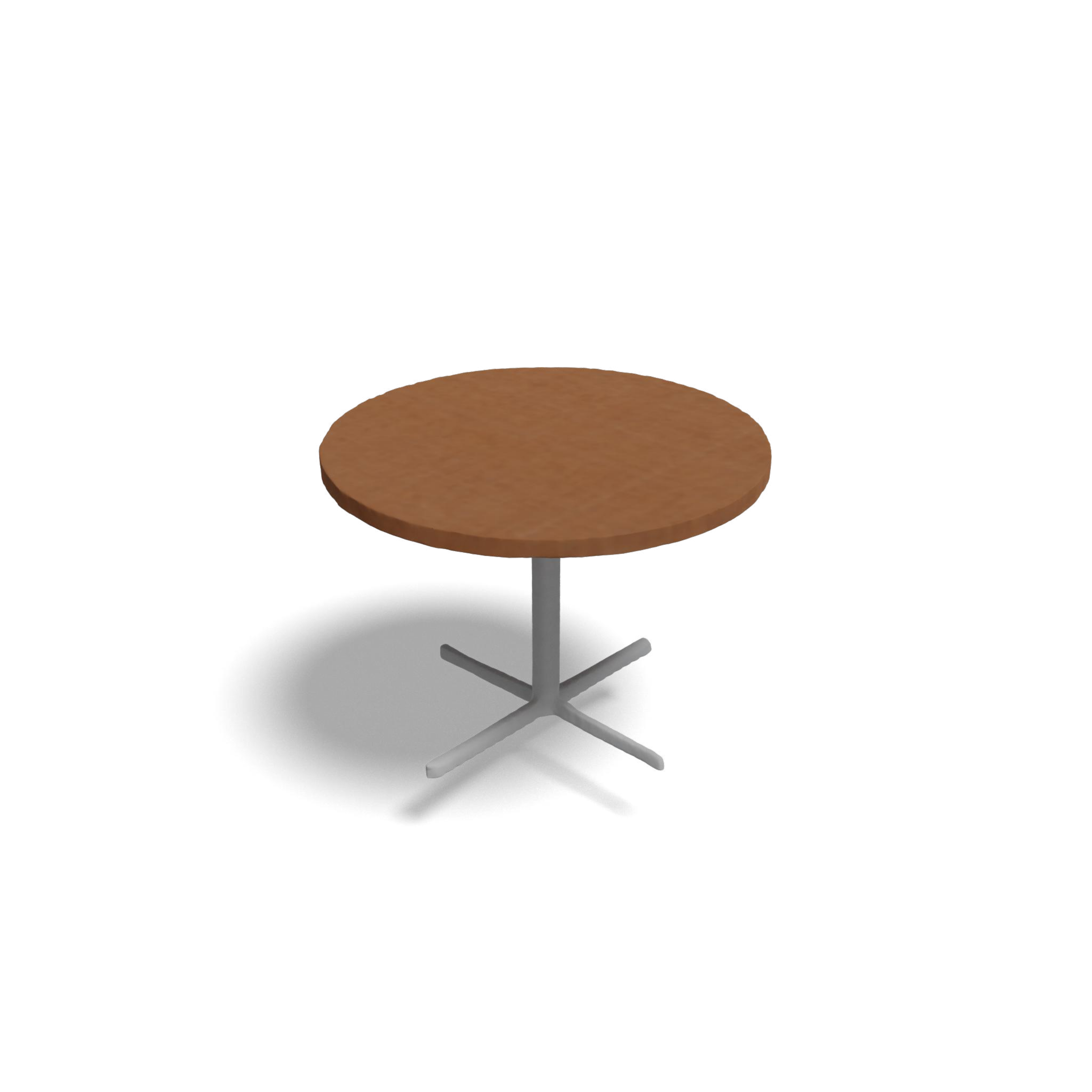}
    \includegraphics[width=0.12\linewidth,trim={380 380 380 380},clip]{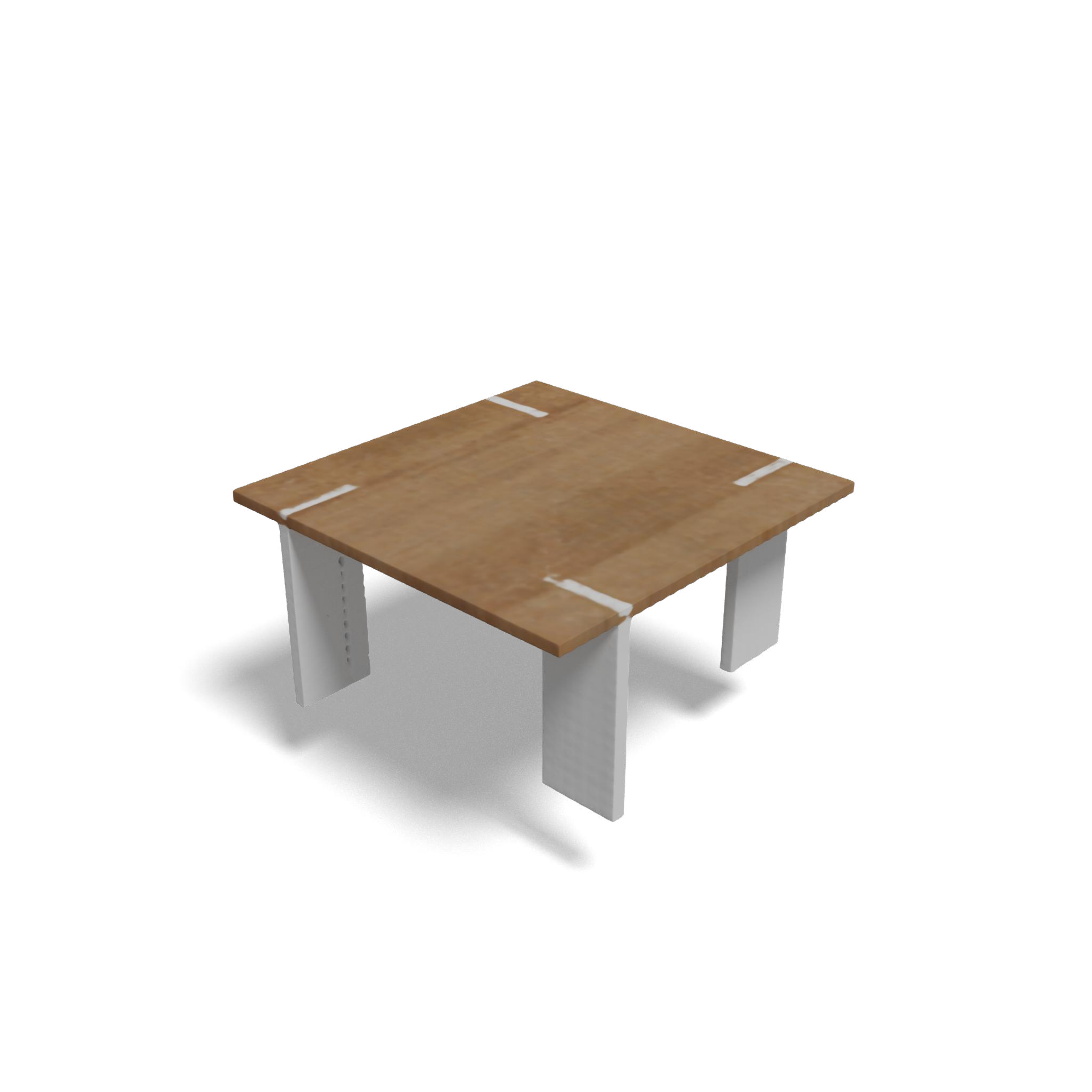}\\
    \vspace{-5pt}
    \includegraphics[width=0.12\linewidth,trim={380 380 380 380},clip]{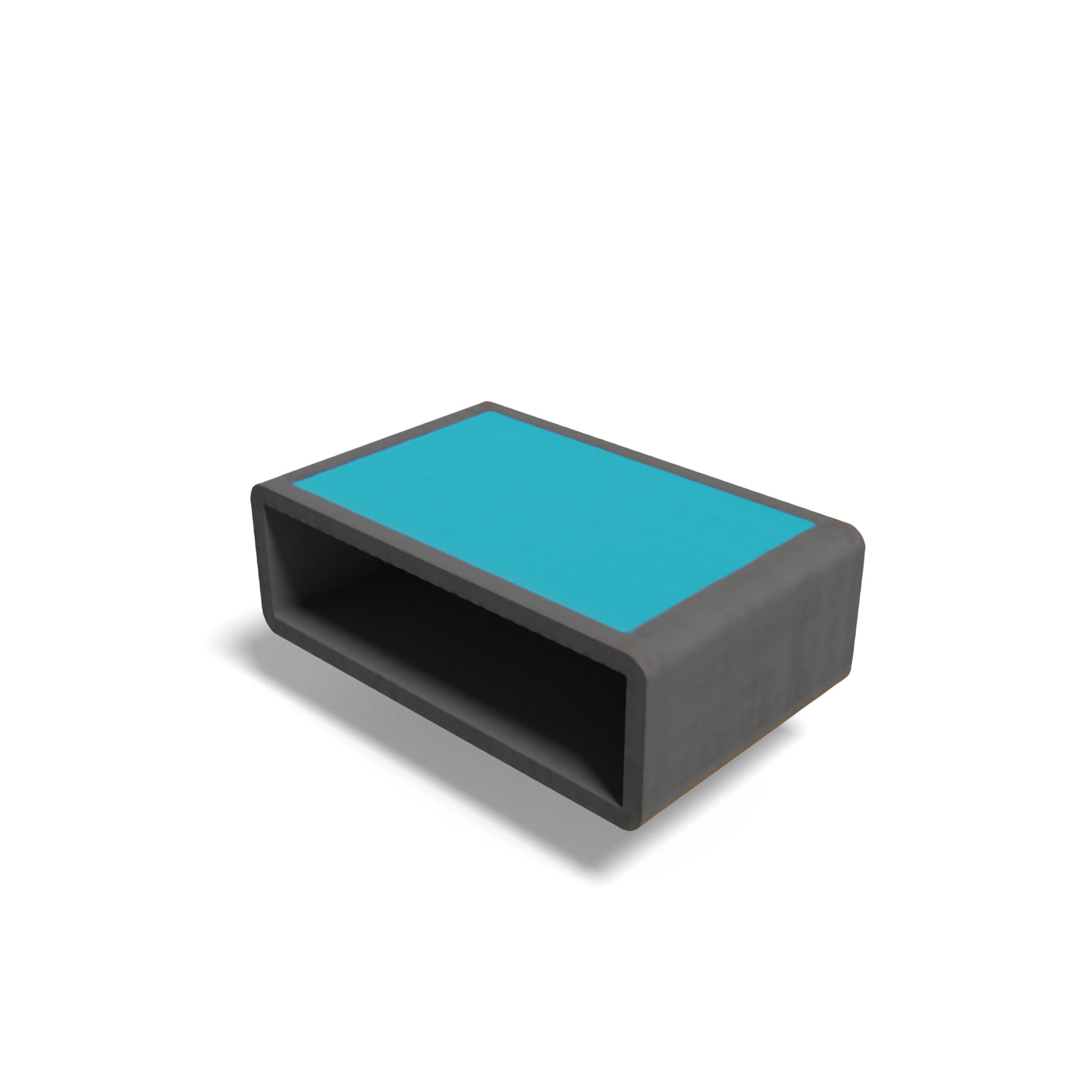}
    \includegraphics[width=0.12\linewidth,trim={380 380 380 380},clip]{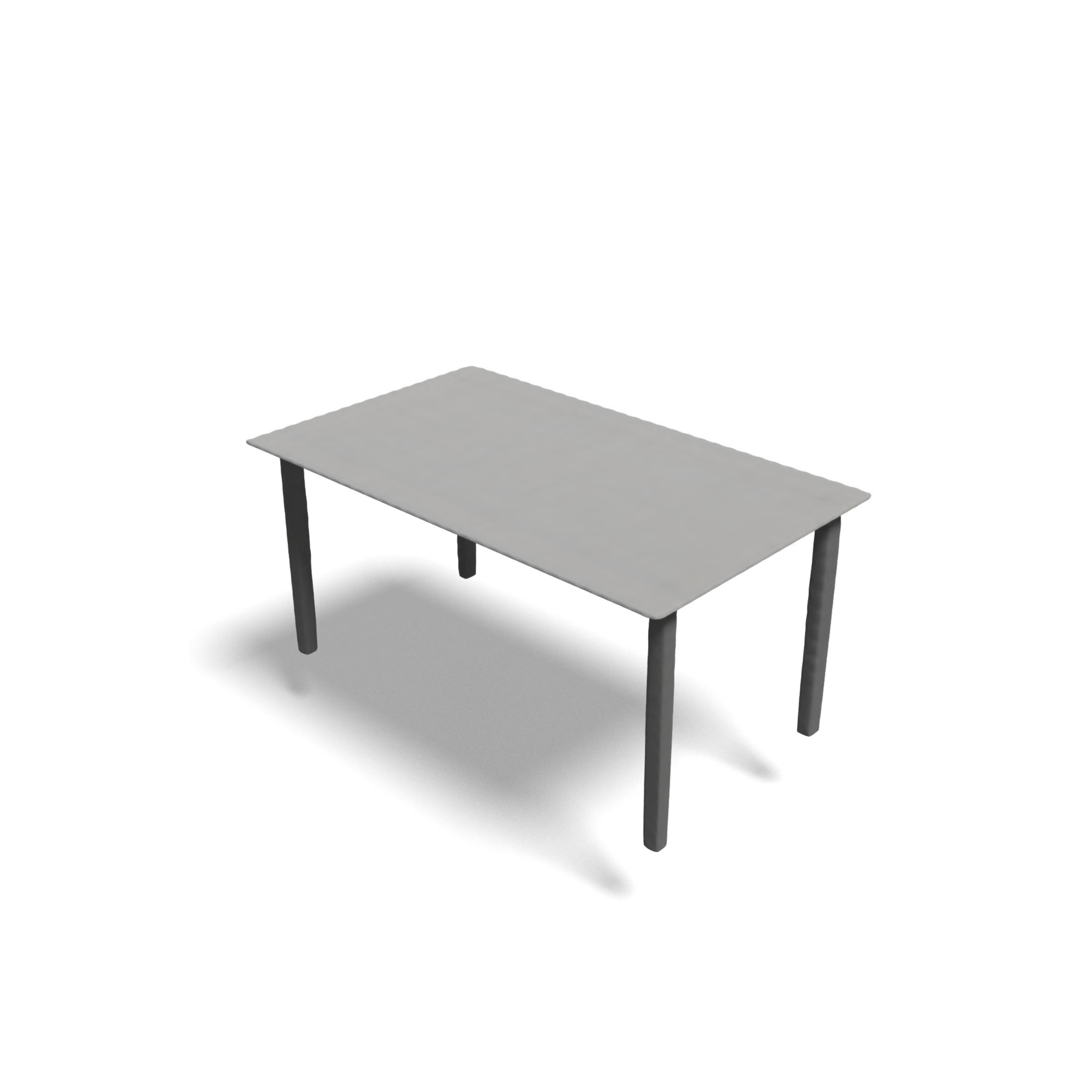}
    \includegraphics[width=0.12\linewidth,trim={205 205 205 205},clip]{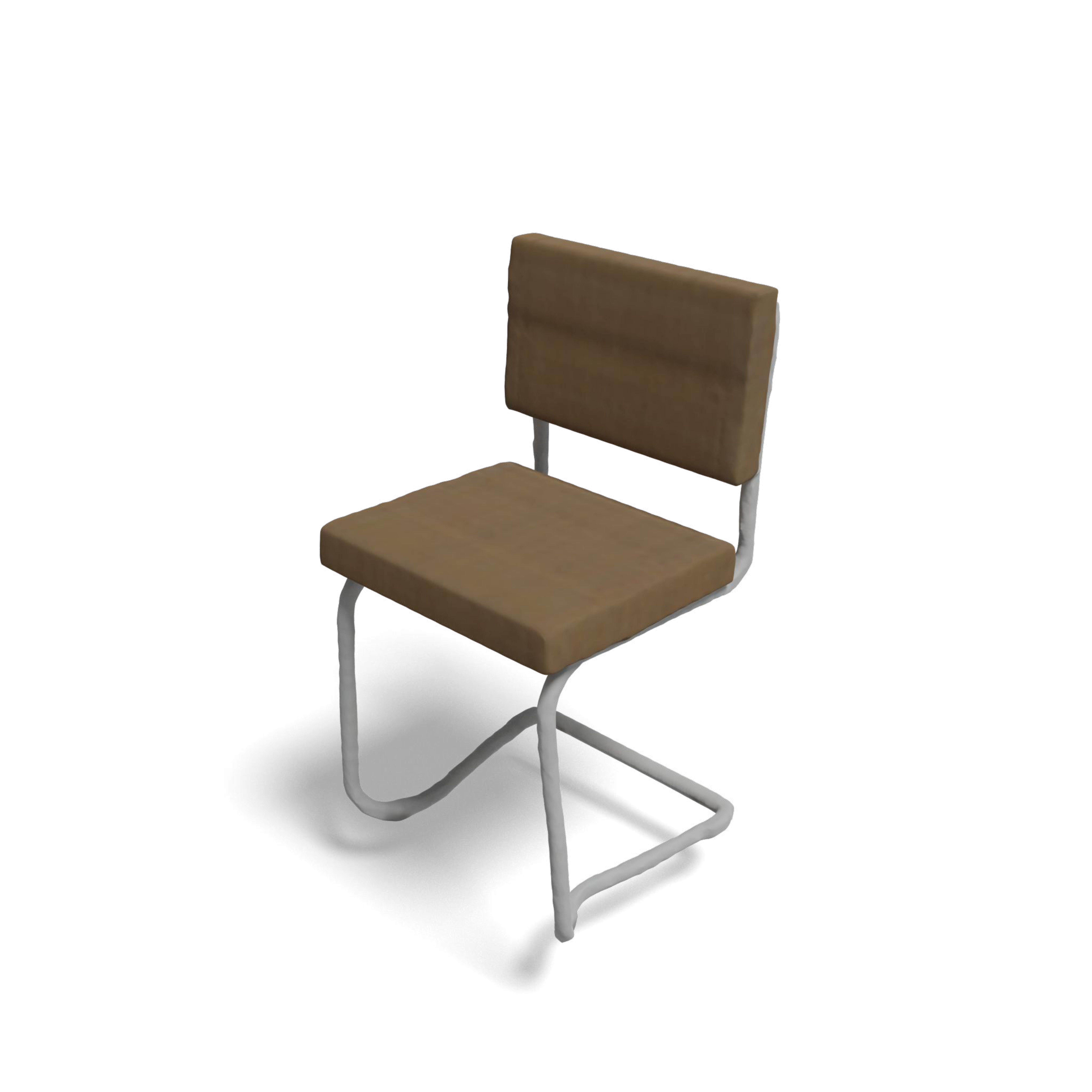}
    \includegraphics[width=0.12\linewidth,trim={205 205 205 205},clip]{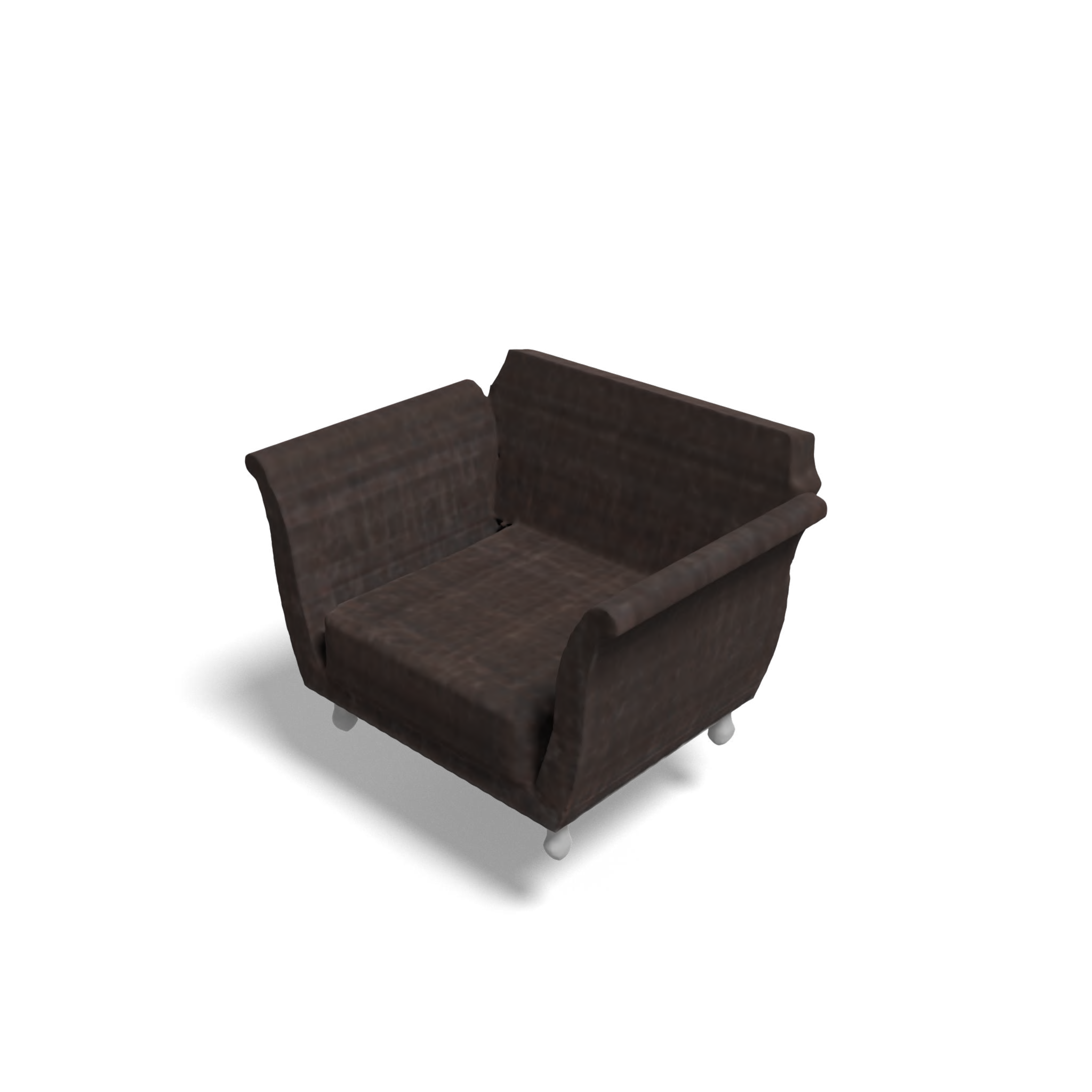}
    \includegraphics[width=0.12\linewidth,trim={205 205 205 205},clip]{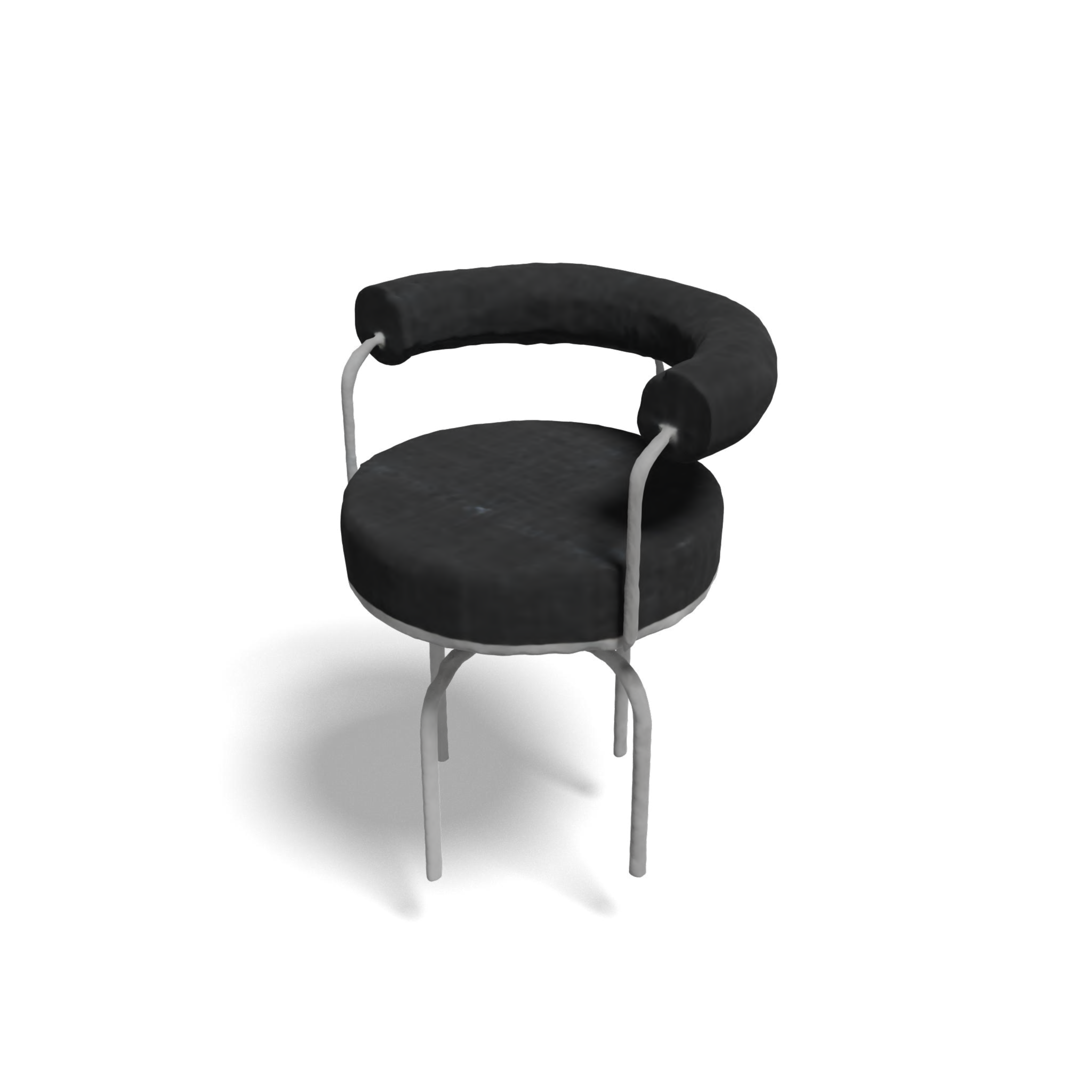}
    \includegraphics[width=0.12\linewidth,trim={205 205 205 205},clip]{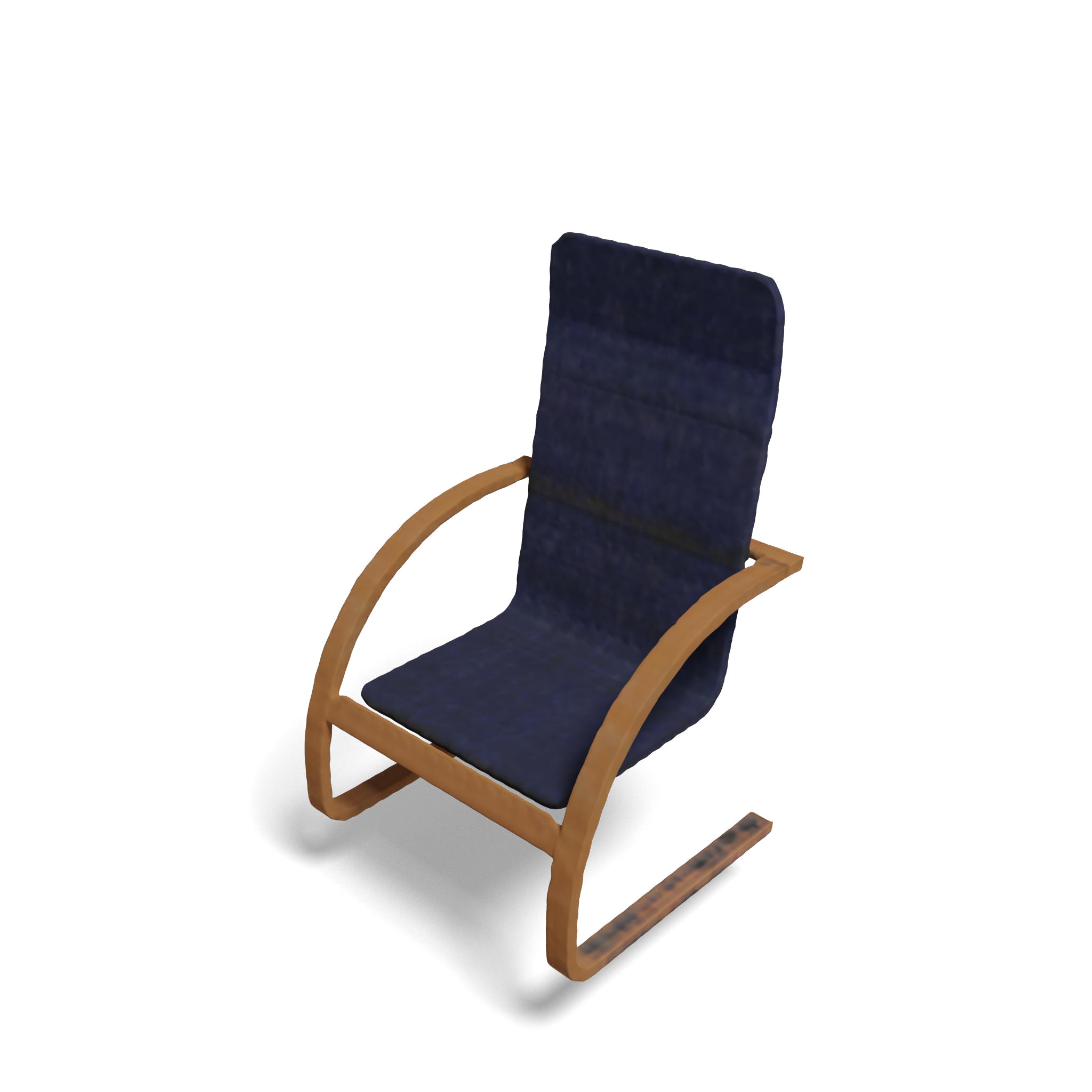}
    \includegraphics[width=0.12\linewidth,trim={205 205 205 205},clip]{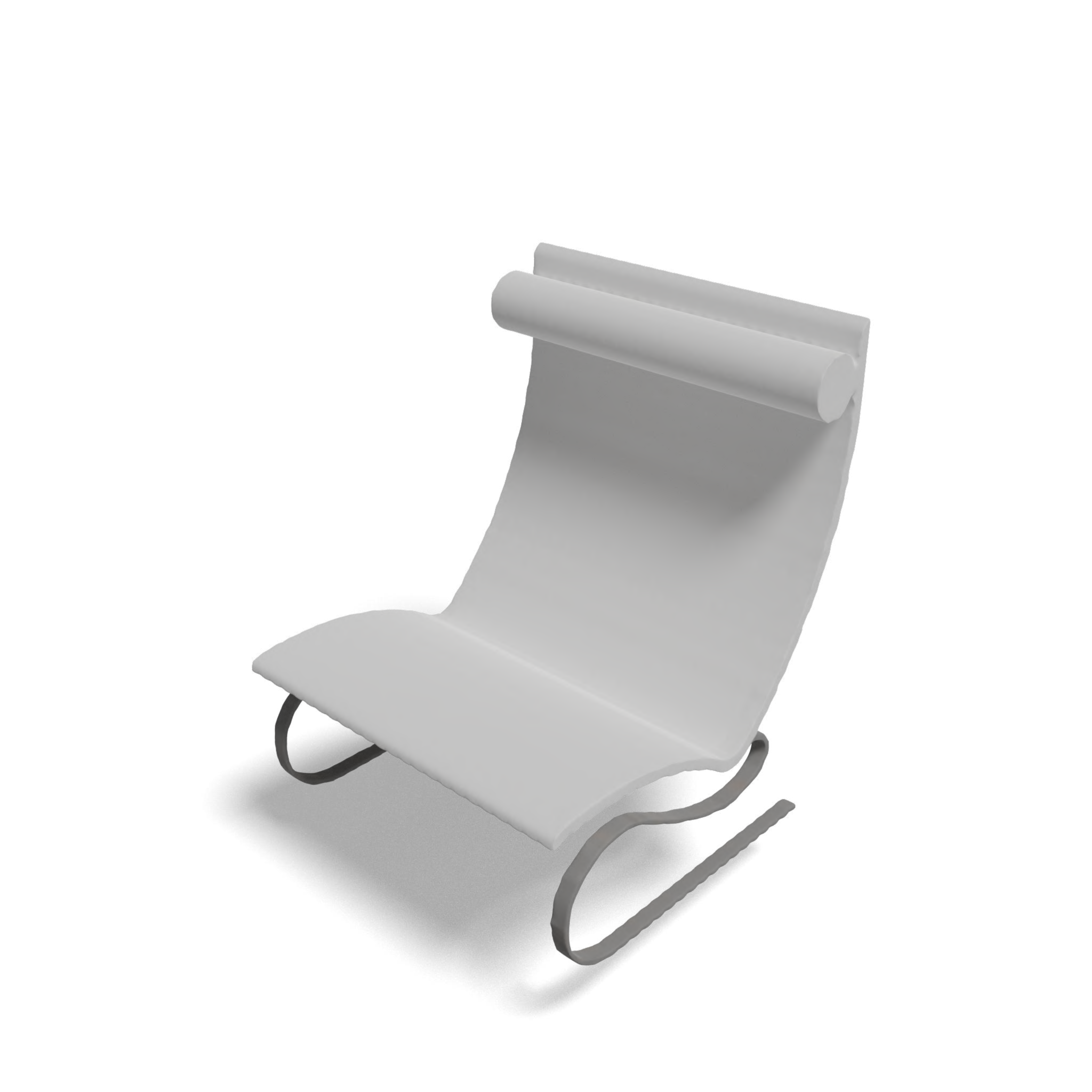}
    \includegraphics[width=0.12\linewidth,trim={205 205 205 205},clip]{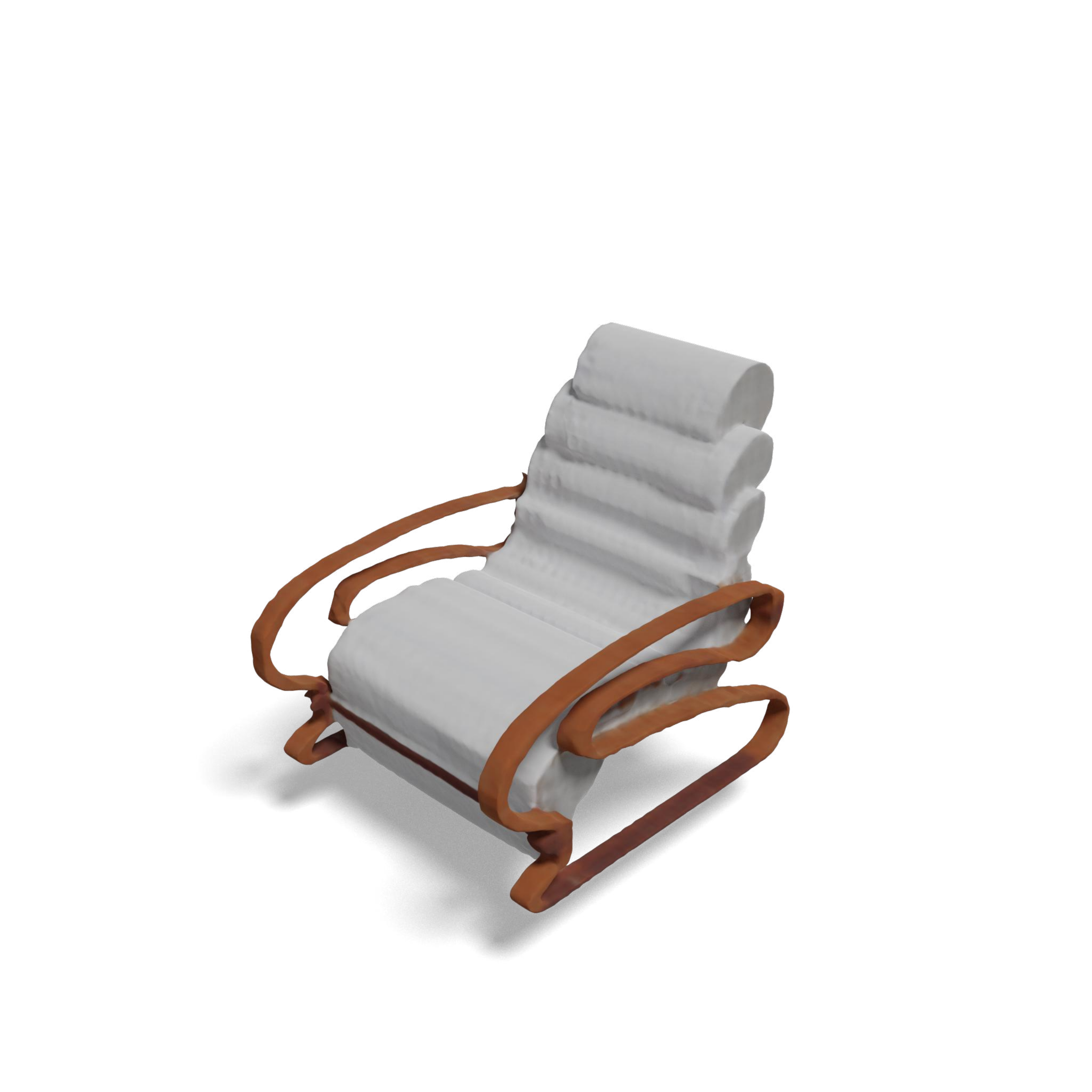}\\
    \vspace{-5pt}
    \includegraphics[width=0.12\linewidth,trim={205 205 205 205},clip]{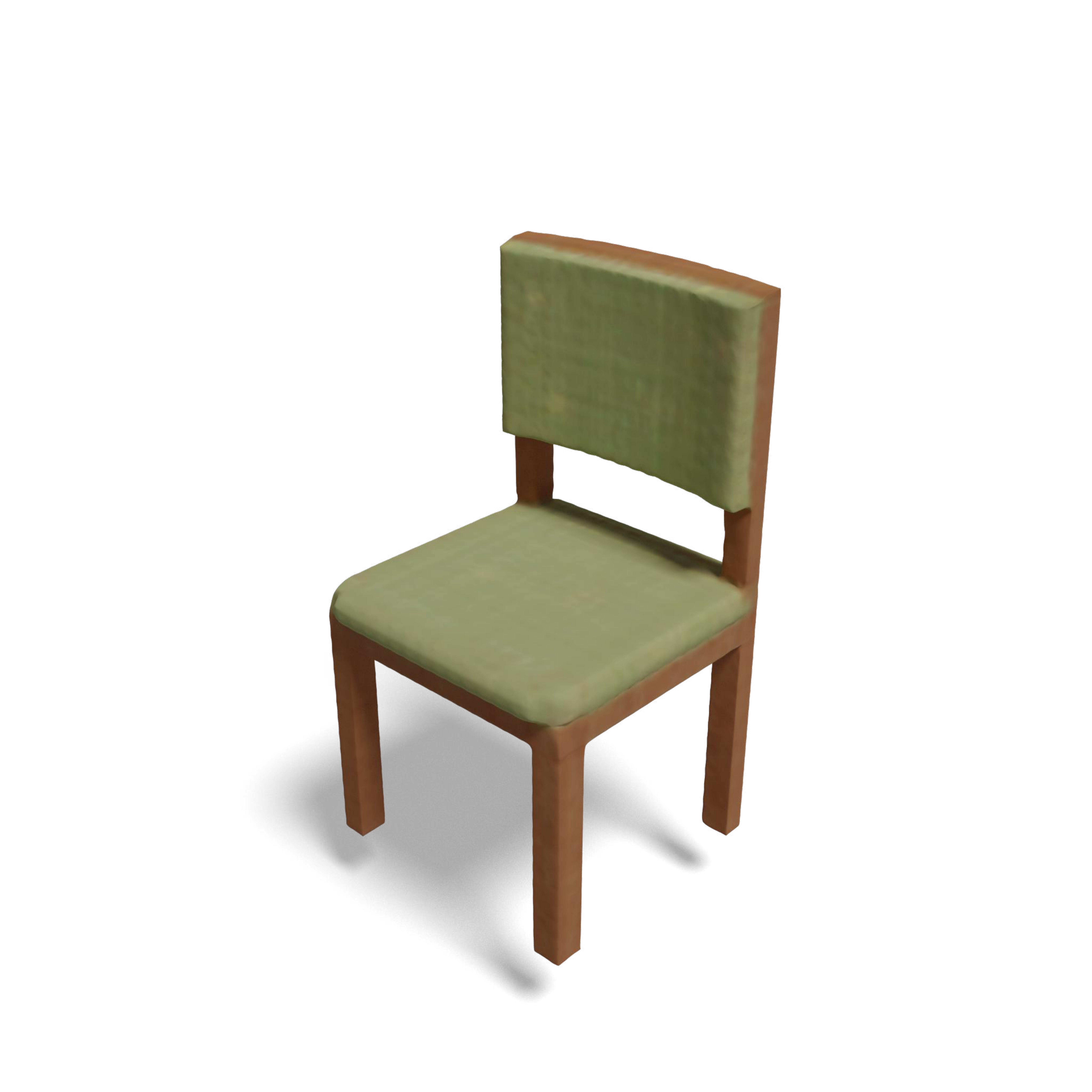}
    \includegraphics[width=0.12\linewidth,trim={205 205 205 205},clip]{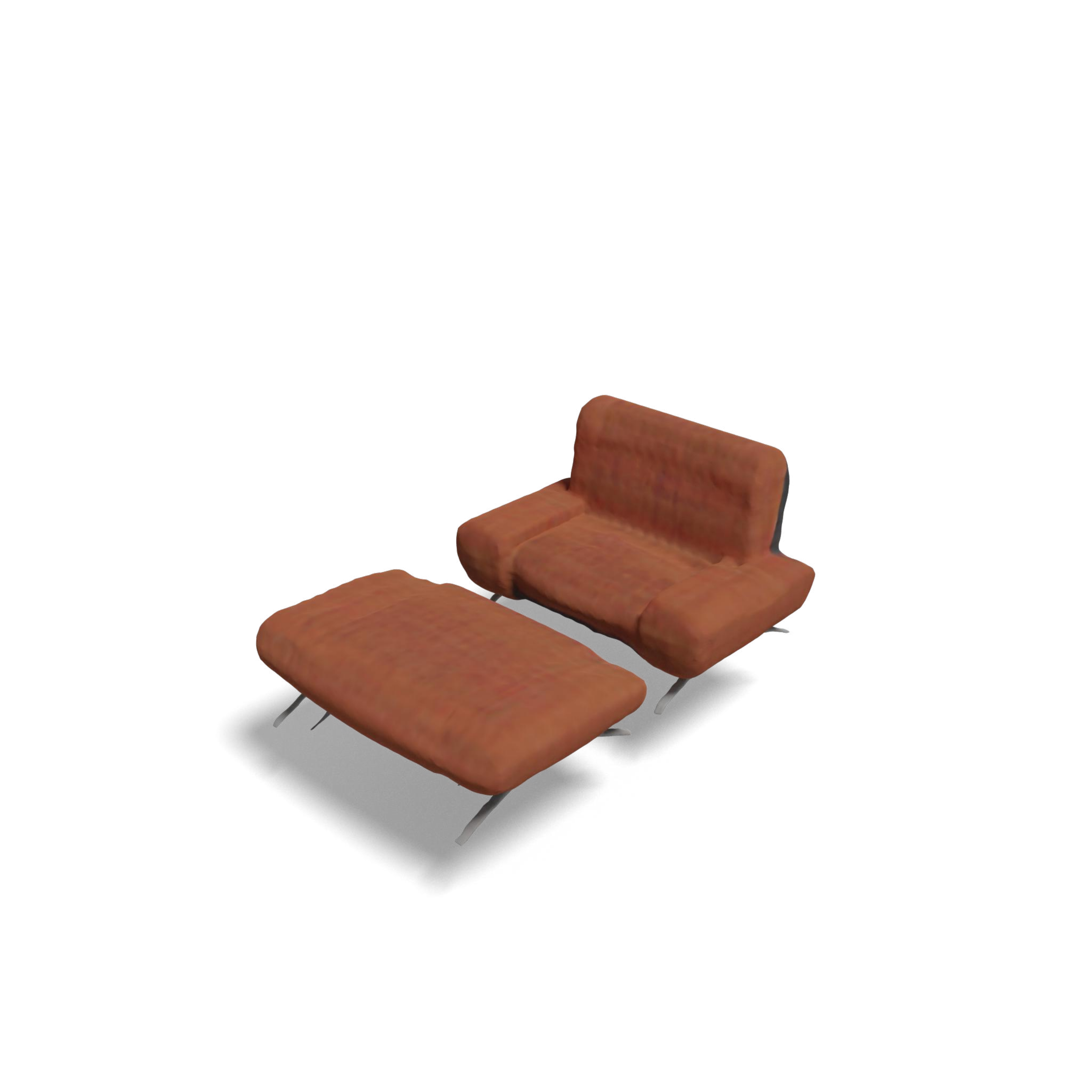}
    \includegraphics[width=0.12\linewidth,trim={205 205 205 205},clip]{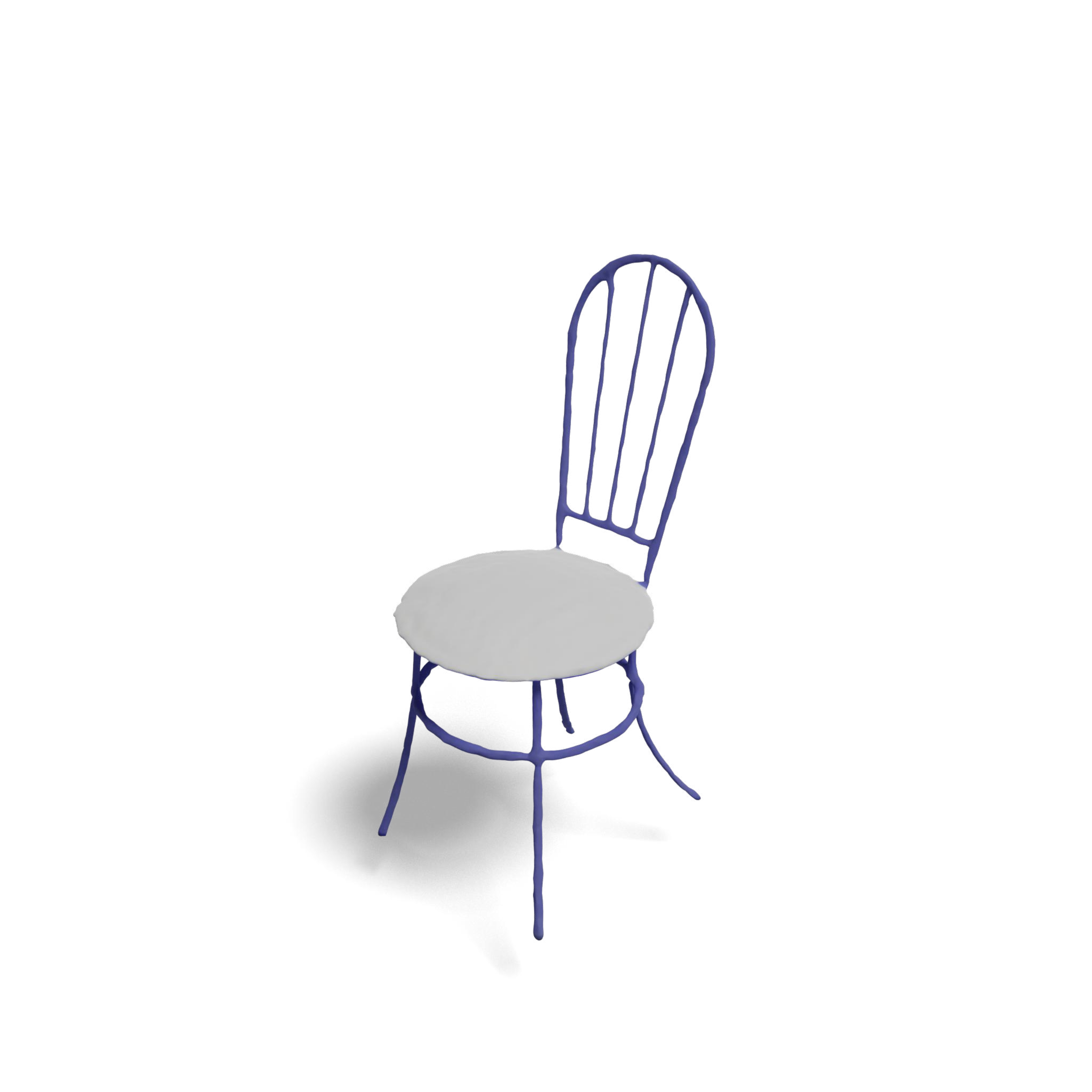}
    \includegraphics[width=0.12\linewidth,trim={205 205 205 205},clip]{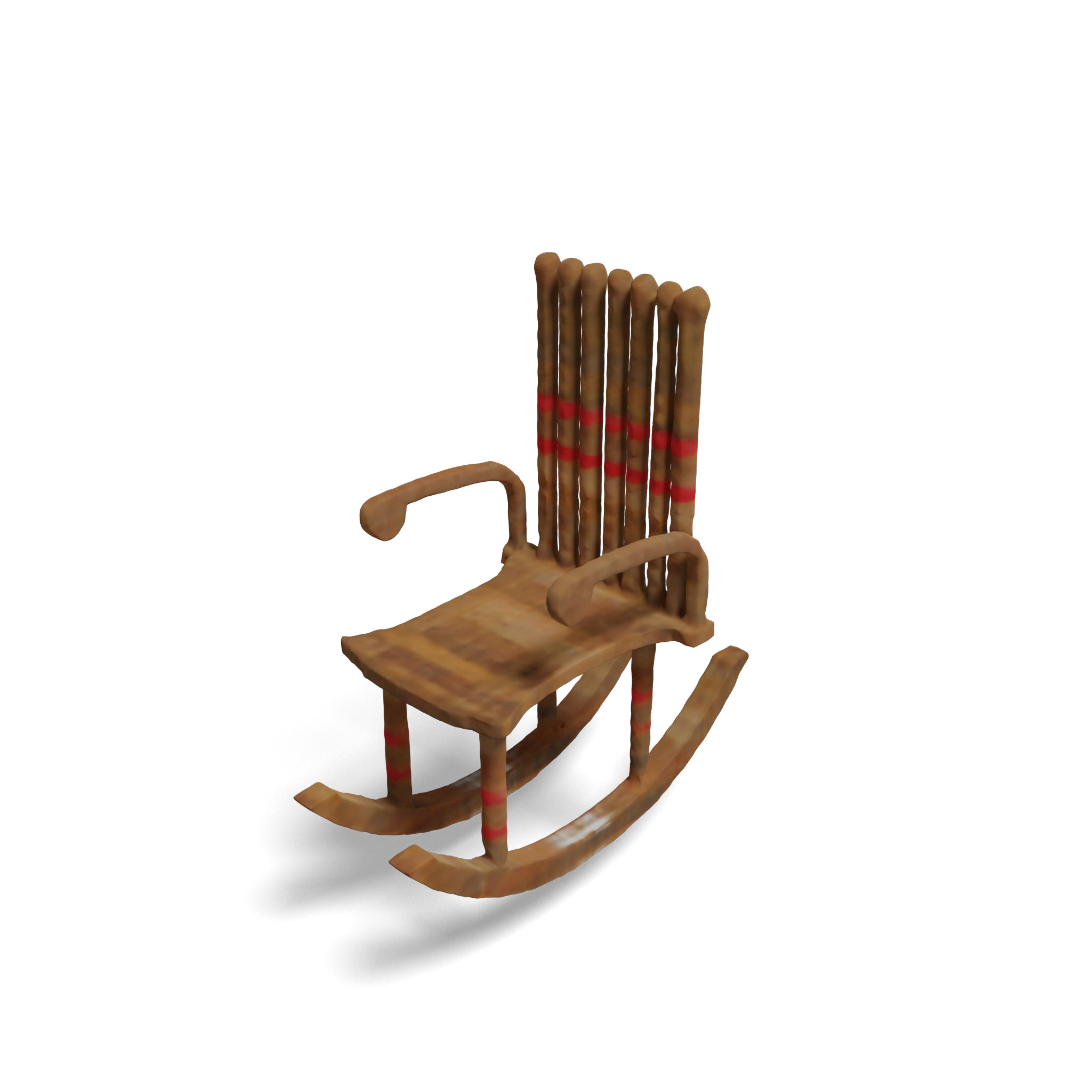}
    \includegraphics[width=0.12\linewidth,trim={205 205 205 205},clip]{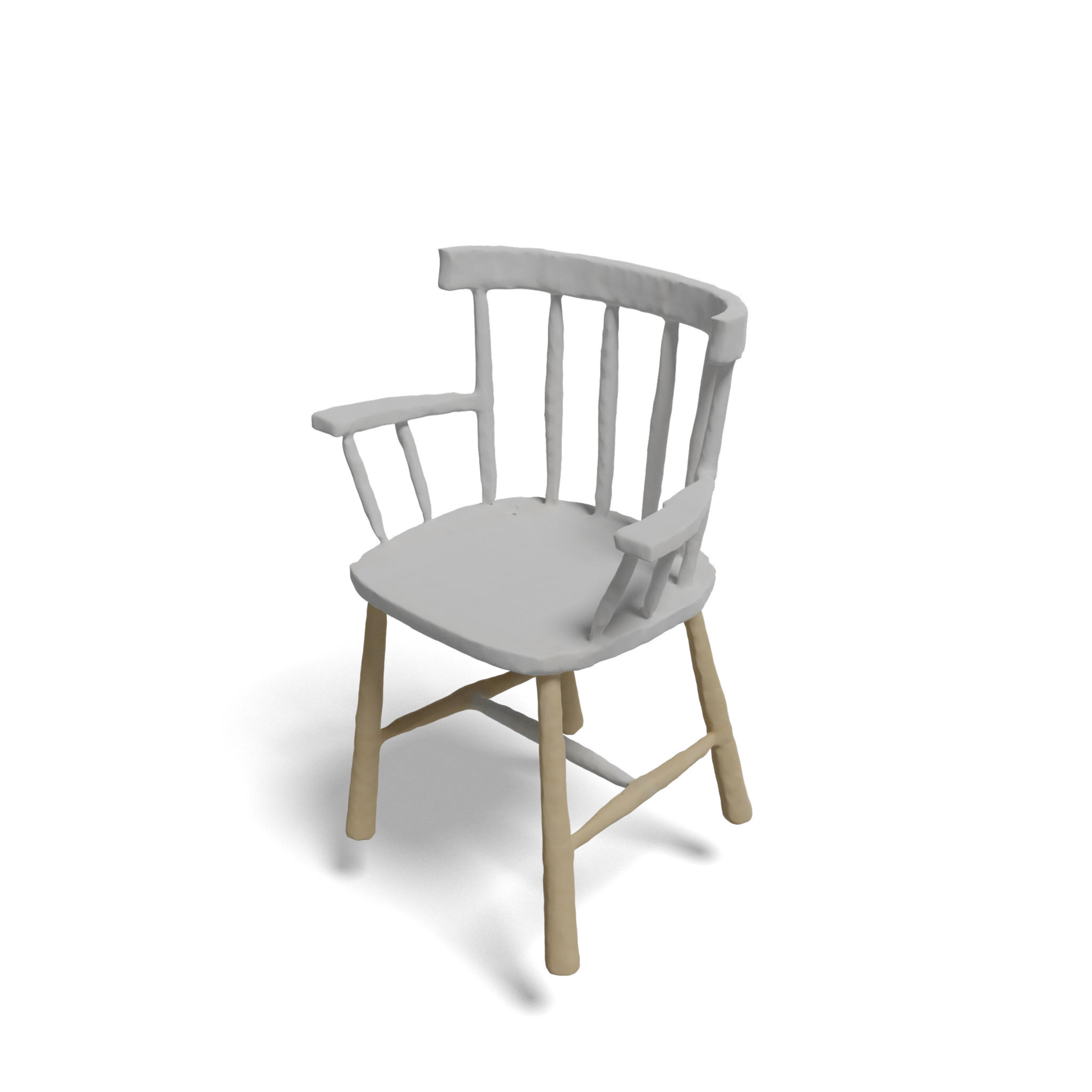}
    \includegraphics[width=0.12\linewidth,trim={205 205 205 205},clip]{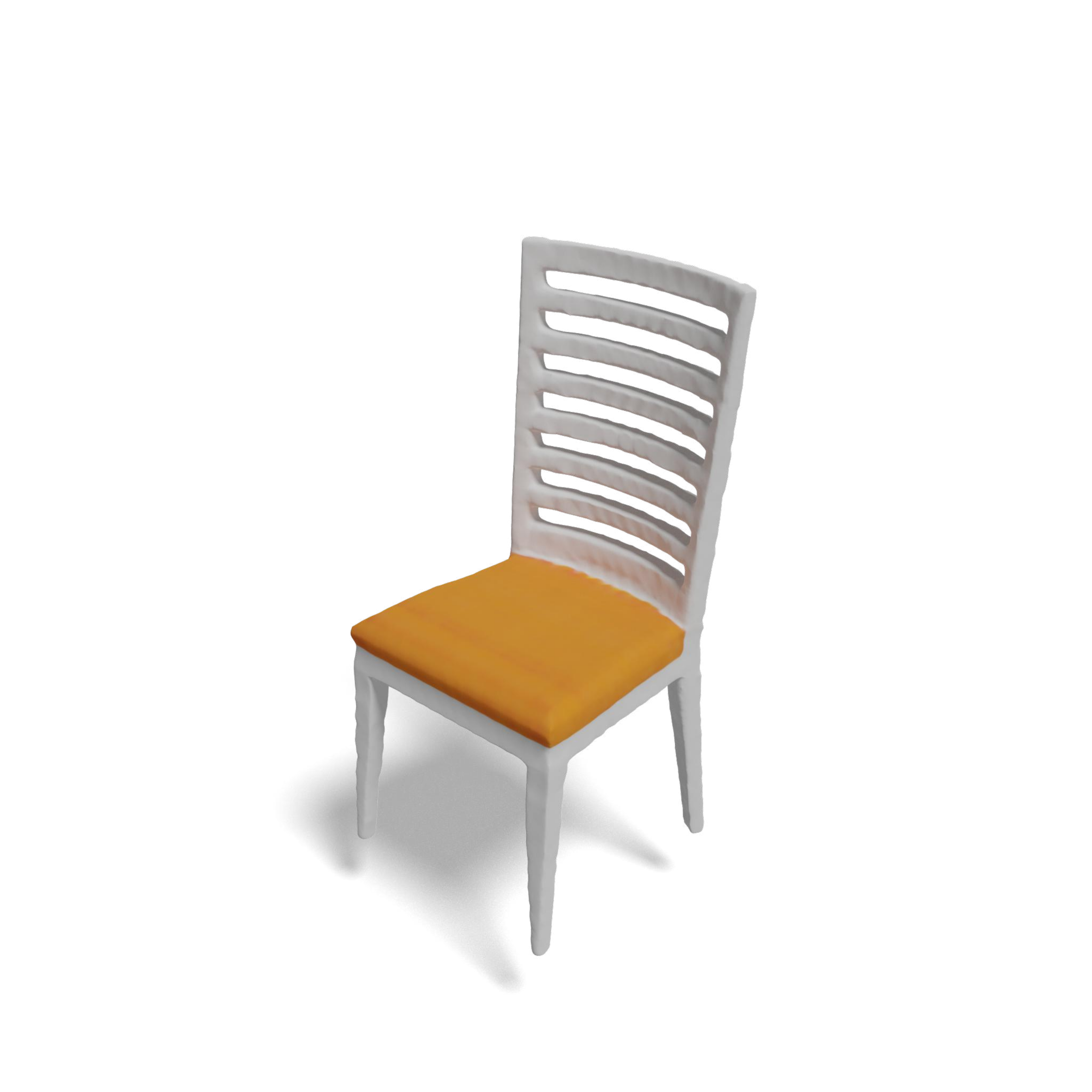}
    \includegraphics[width=0.12\linewidth,trim={205 205 205 205},clip]{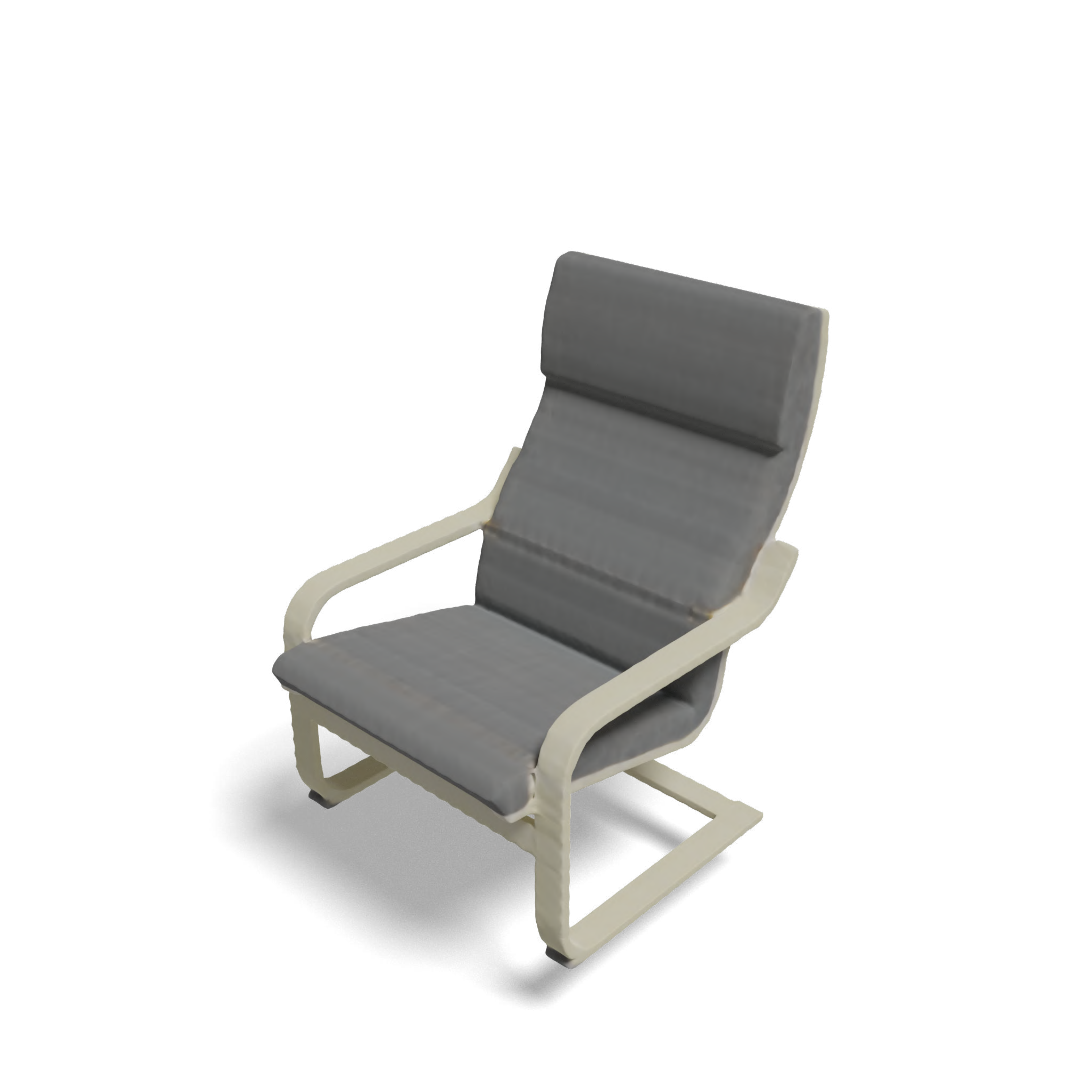}
    \includegraphics[width=0.12\linewidth,trim={205 205 205 205},clip]{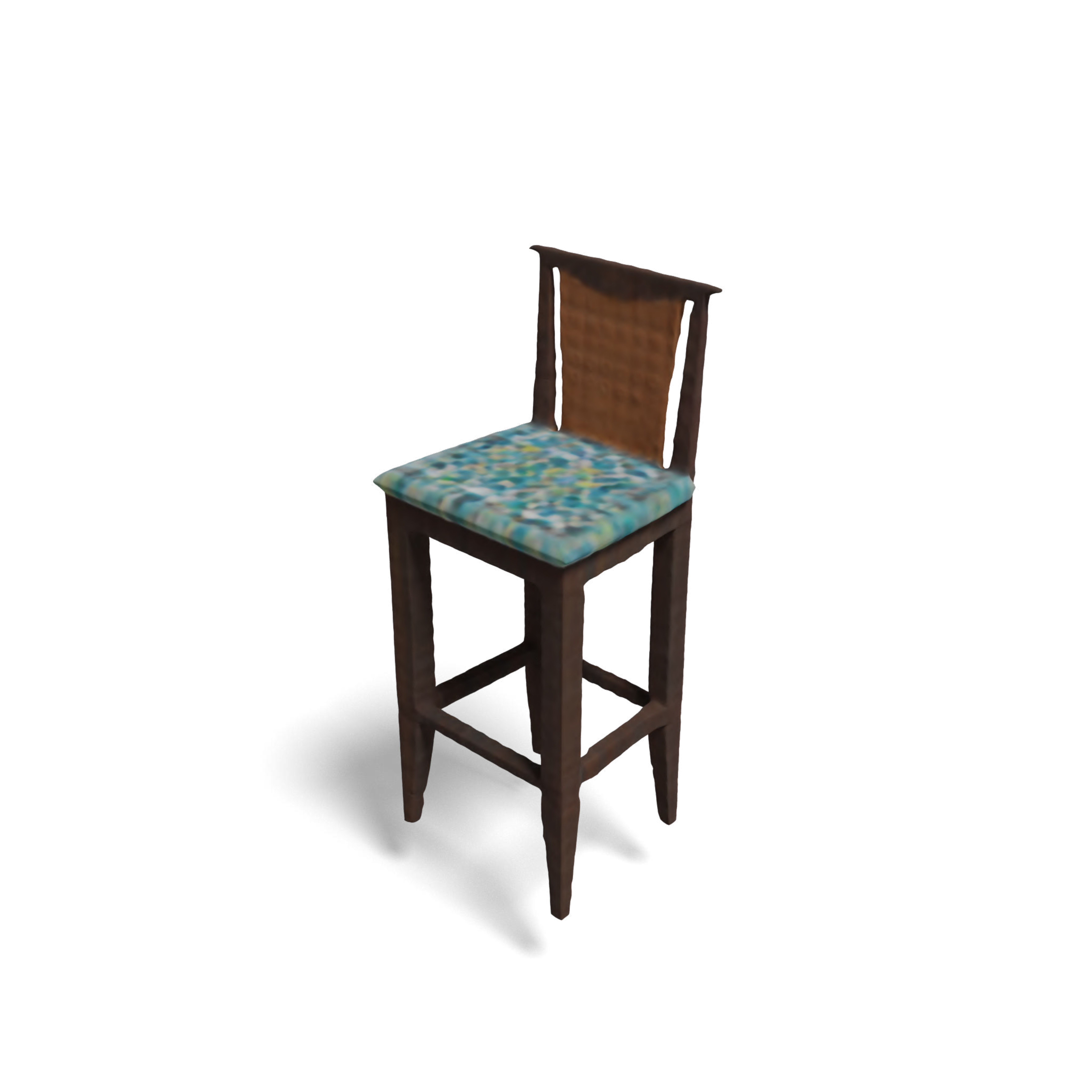}\\
    \vspace{-7pt}
    \includegraphics[width=0.12\linewidth,trim={205 205 205 205},clip]{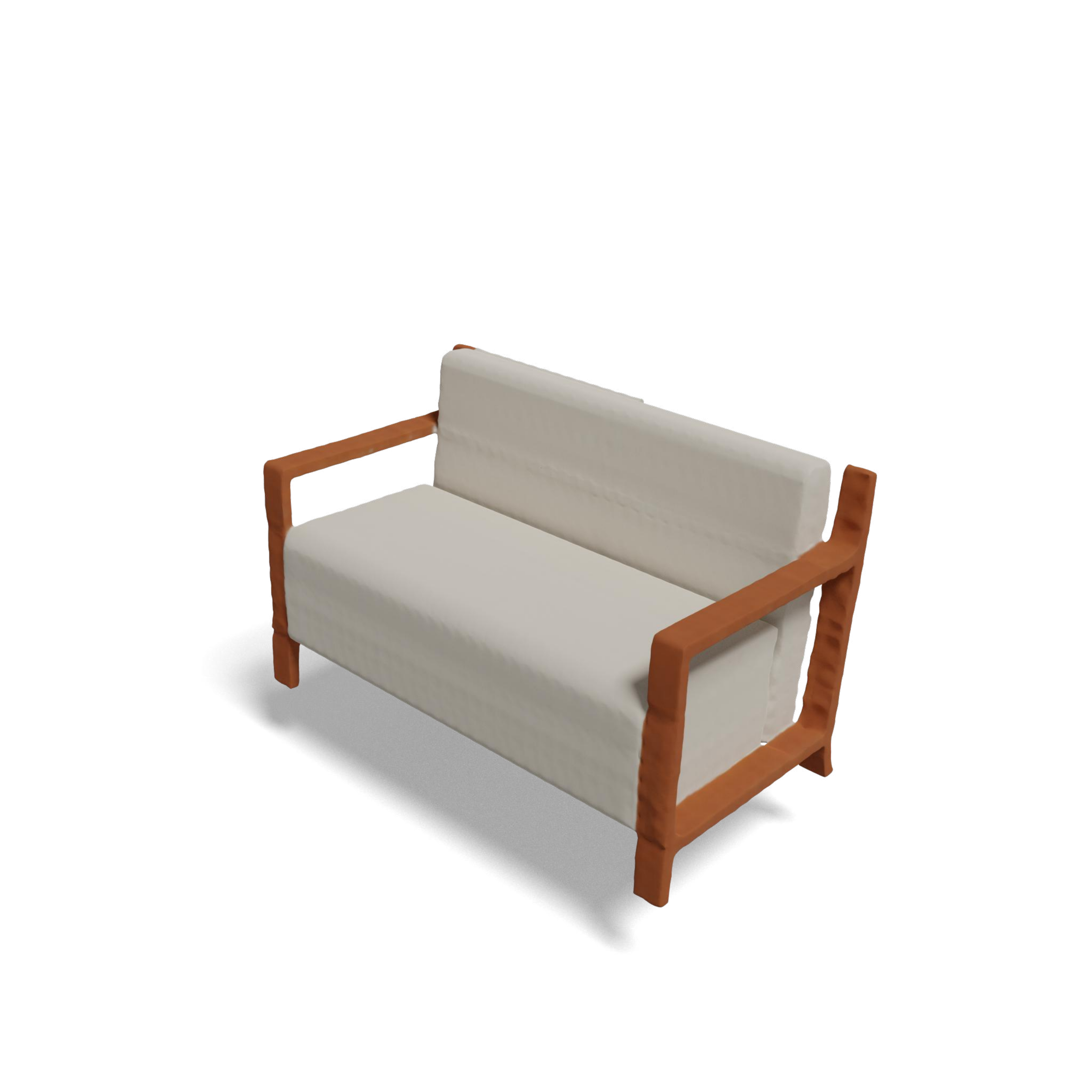}
    \includegraphics[width=0.12\linewidth,trim={205 205 205 205},clip]{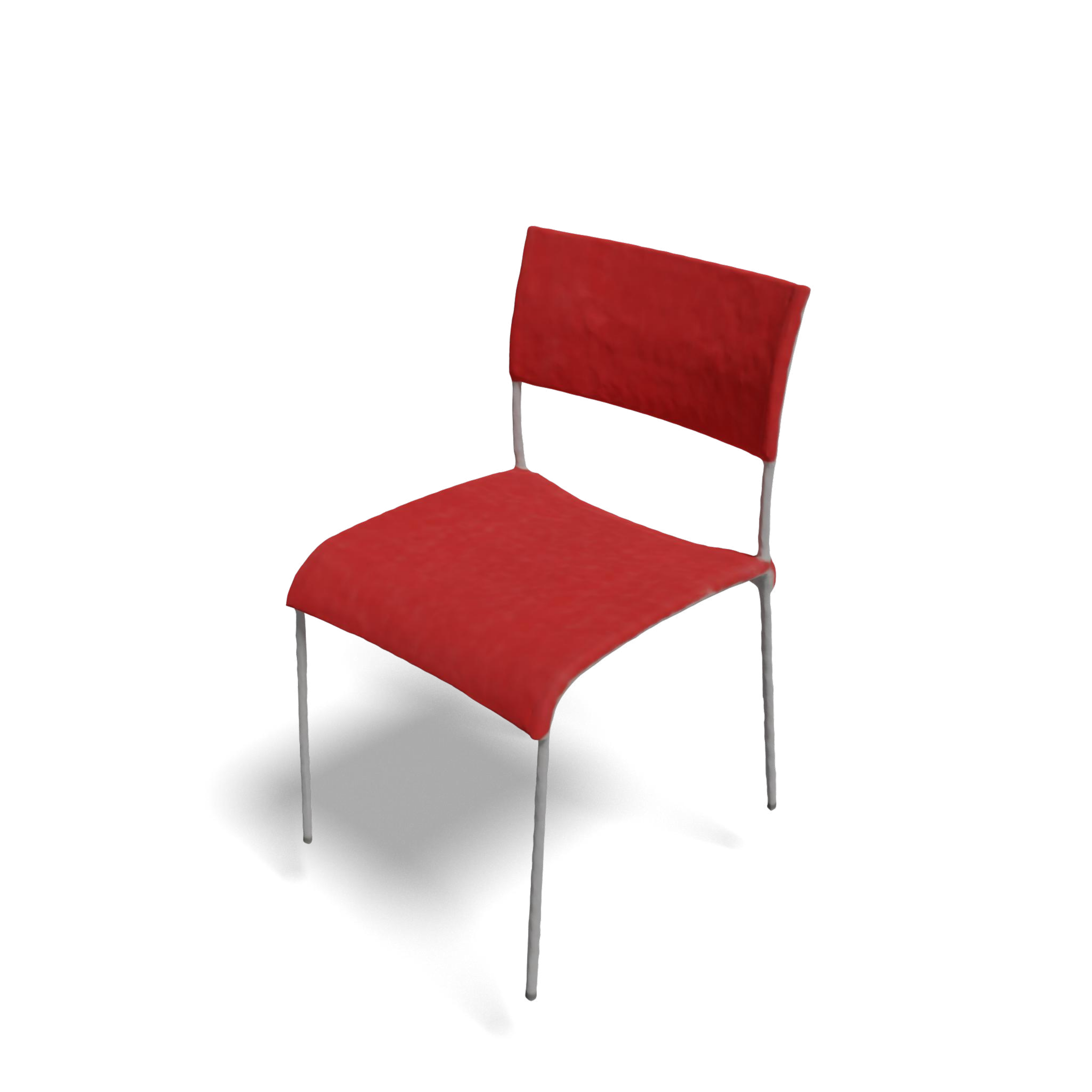}
    \includegraphics[width=0.12\linewidth,trim={205 205 205 205},clip]{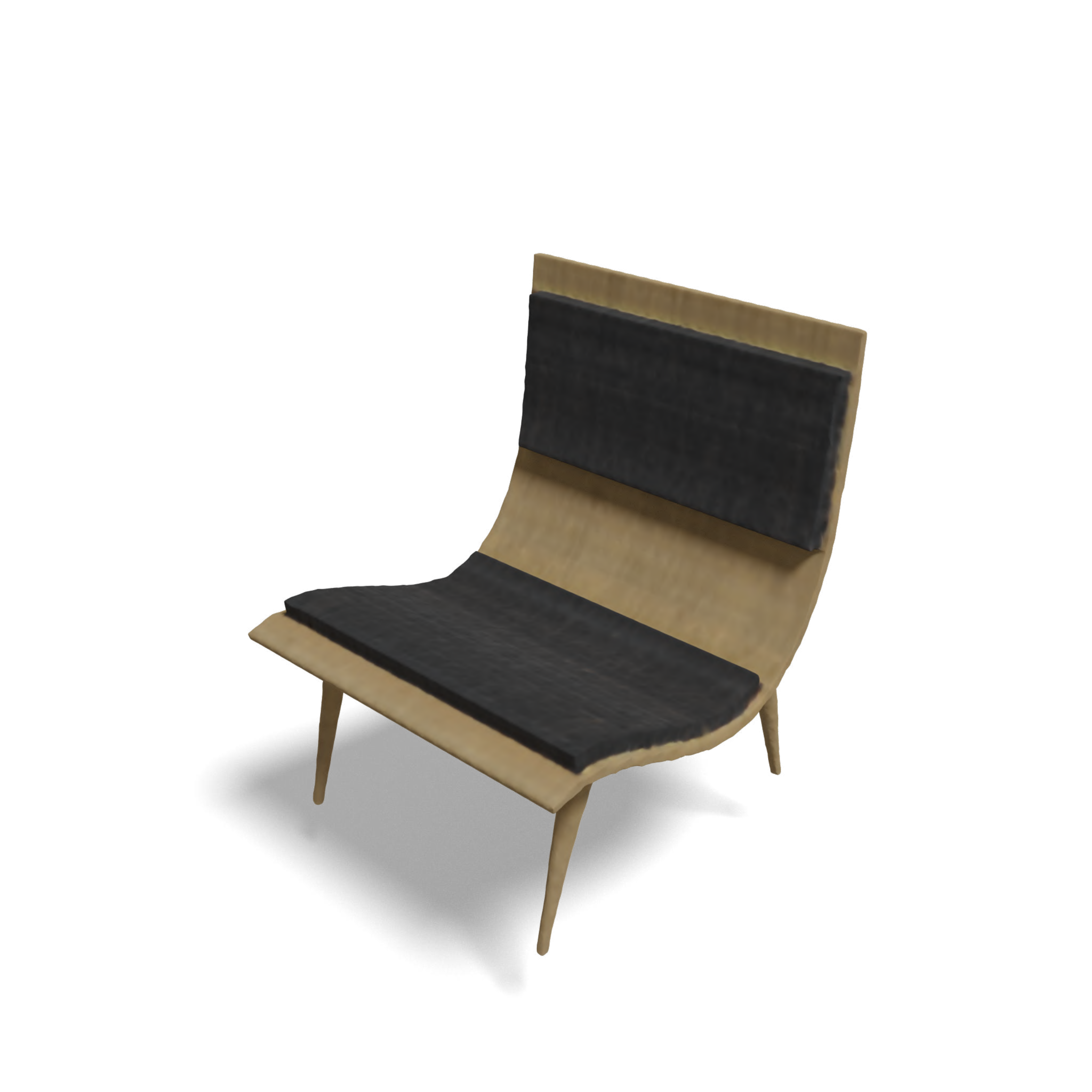}
    \includegraphics[width=0.12\linewidth,trim={205 205 205 205},clip]{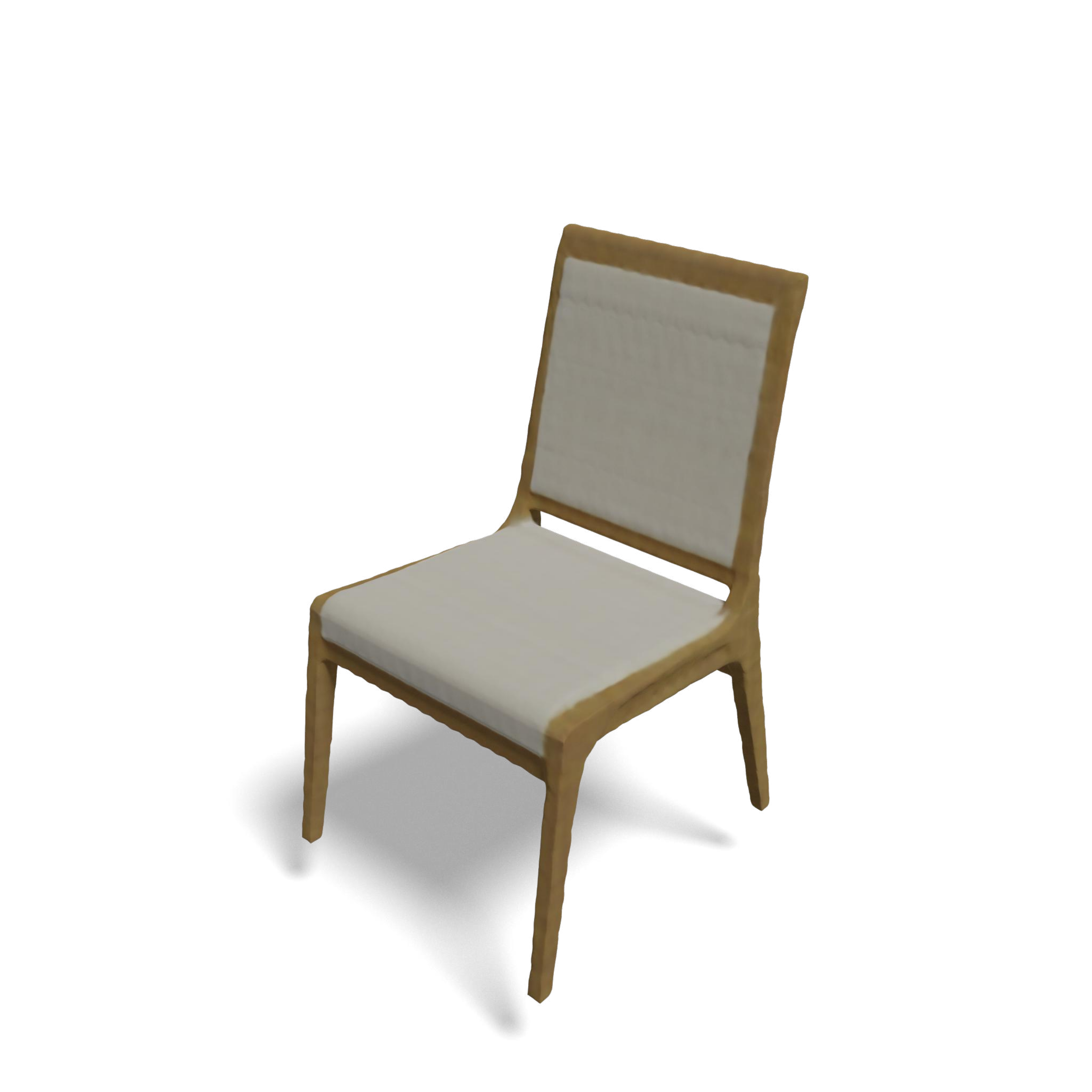}
    \includegraphics[width=0.12\linewidth,trim={205 205 205 205},clip]{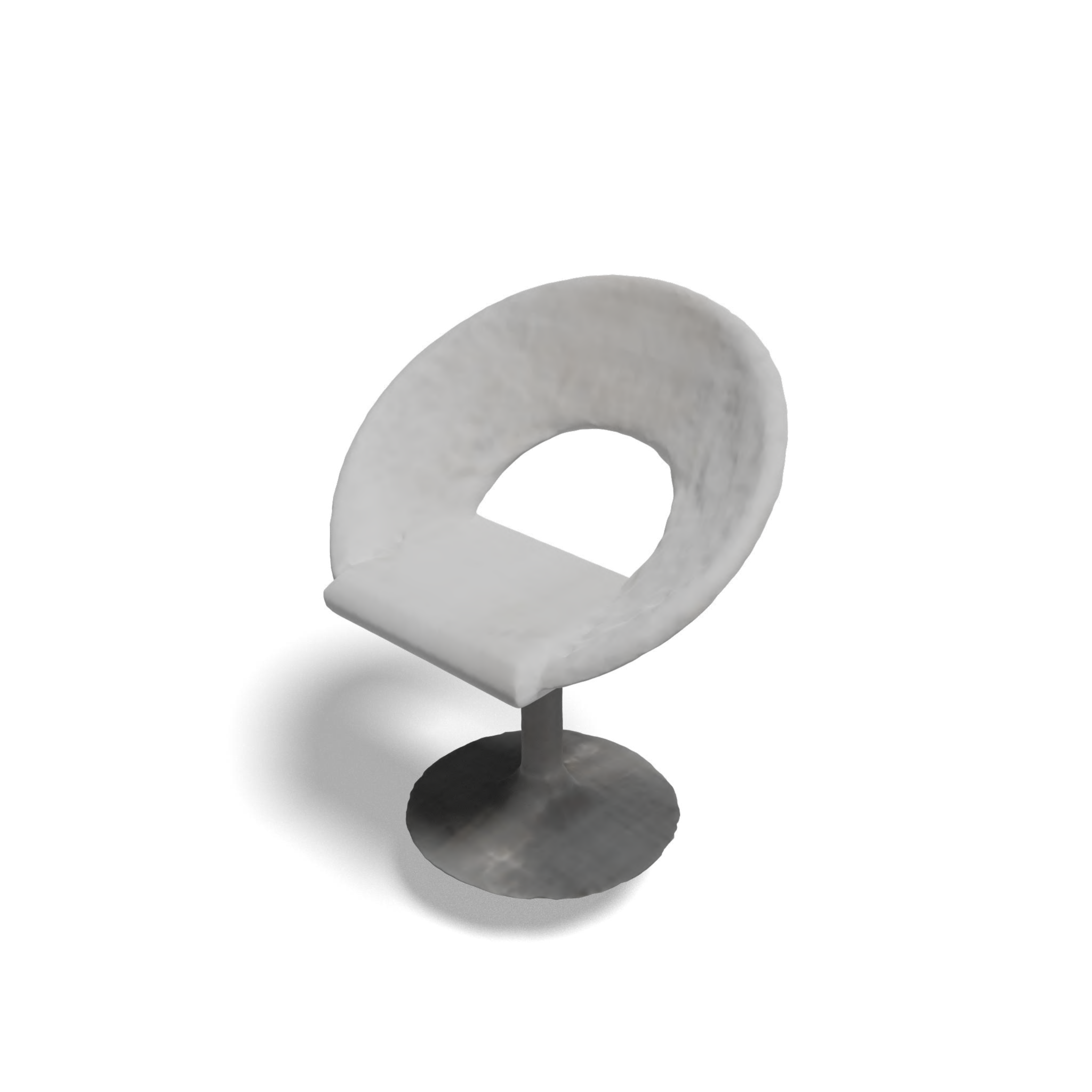}
    % \includegraphics[width=0.12\linewidth,trim={205 205 205 205},clip]{iccv2023AuthorKit/figures/generated_chair_samples/chairs21-min_17.pdf}
    % \includegraphics[width=0.12\linewidth,trim={205 205 205 205},clip]{iccv2023AuthorKit/figures/generated_chair_samples/chairs21-min_18.pdf}
    % \includegraphics[width=0.12\linewidth,trim={205 205 205 205},clip]{iccv2023AuthorKit/figures/generated_chair_samples/chairs21-min_19.pdf}
    % % \includegraphics[width=0.12\linewidth,trim={205 205 205 205},clip]{iccv2023AuthorKit/figures/generated_chair_samples/chairs21-min_20.pdf}
    % \includegraphics[width=0.12\linewidth,trim={205 205 205 205},clip]{iccv2023AuthorKit/figures/generated_chair_samples/chairs21-min_21.pdf}\\
    \includegraphics[width=0.12\linewidth,trim={410 400 410 410},clip]{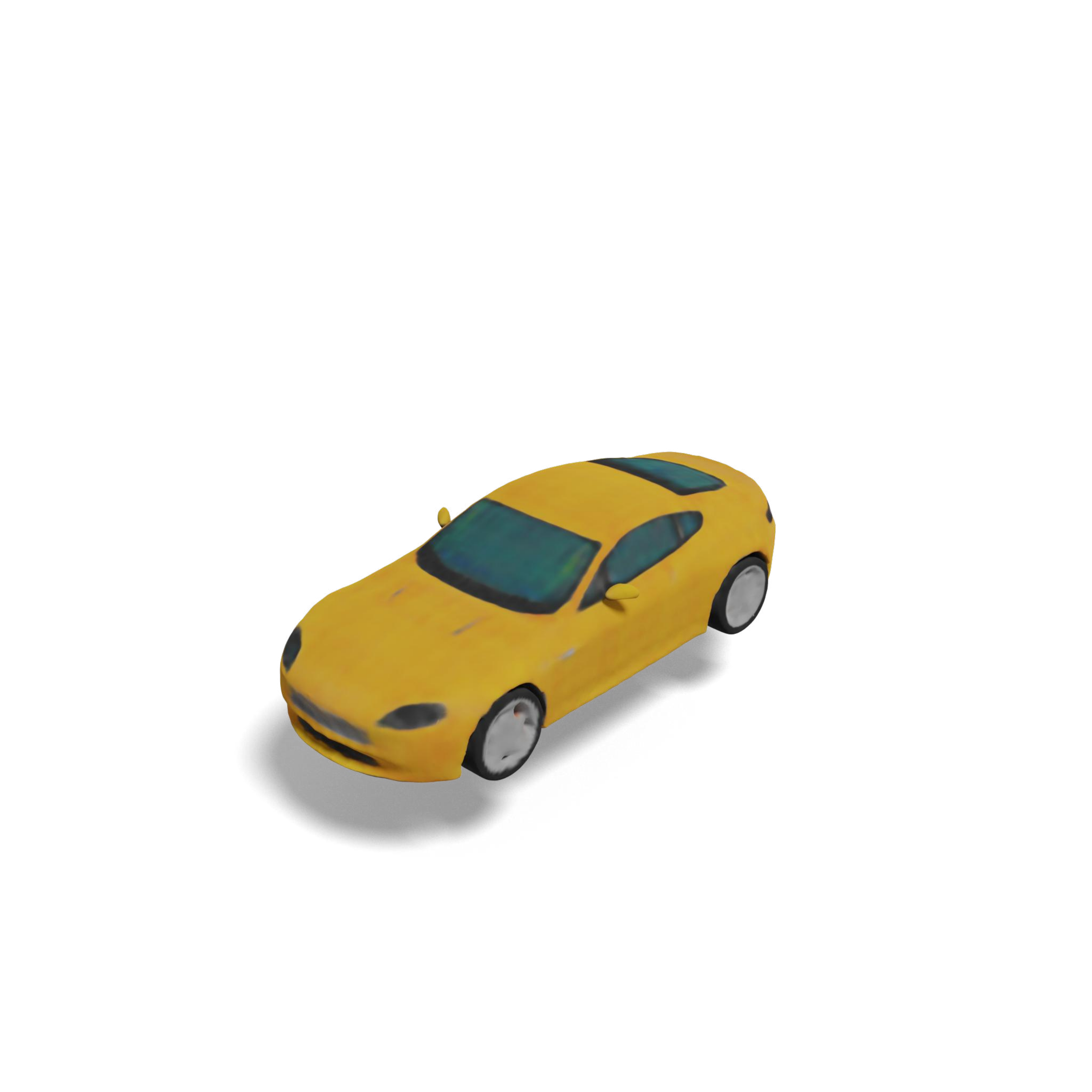}
    \includegraphics[width=0.12\linewidth,trim={410 400 410 410},clip]{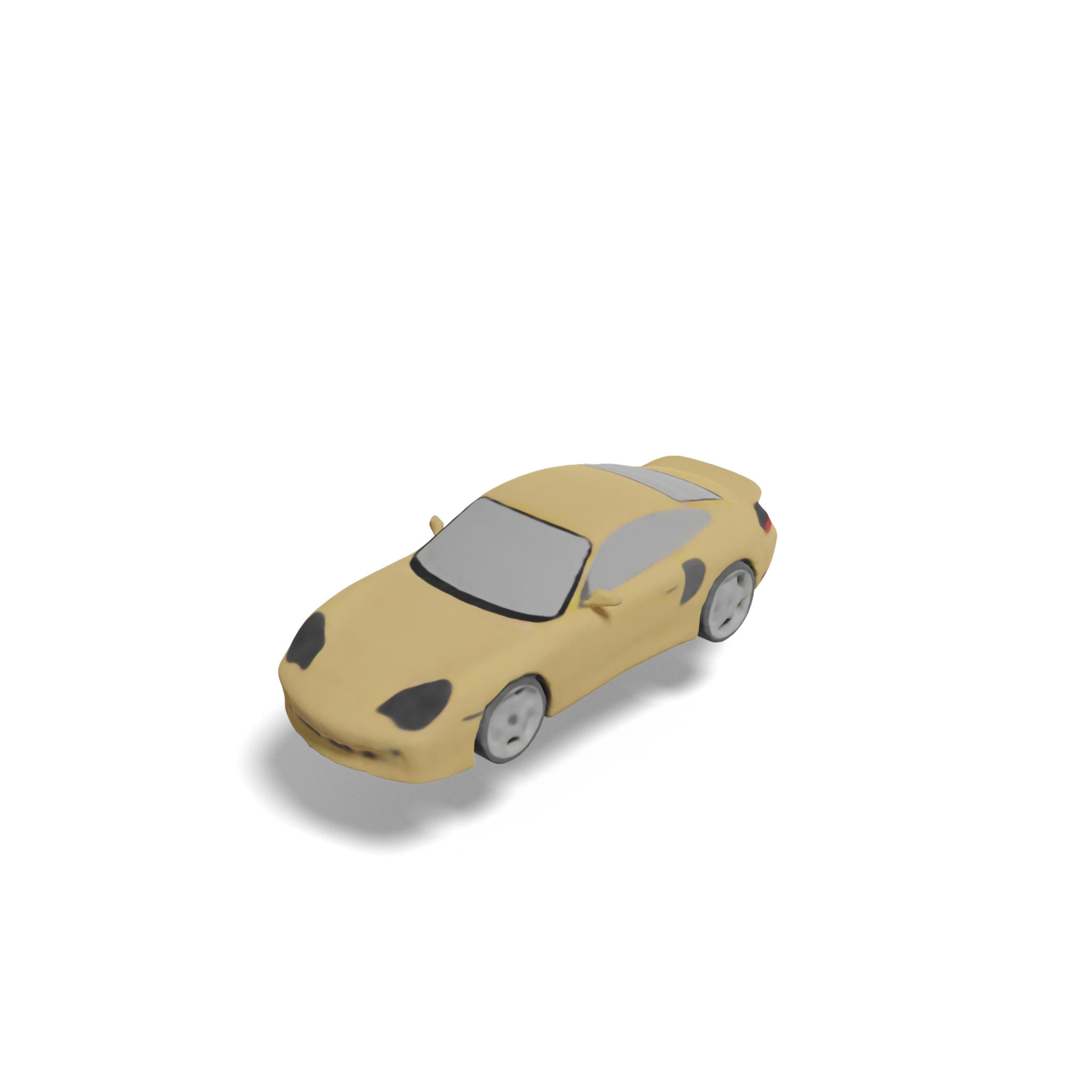}
    \includegraphics[width=0.12\linewidth,trim={410 400 410 410},clip]{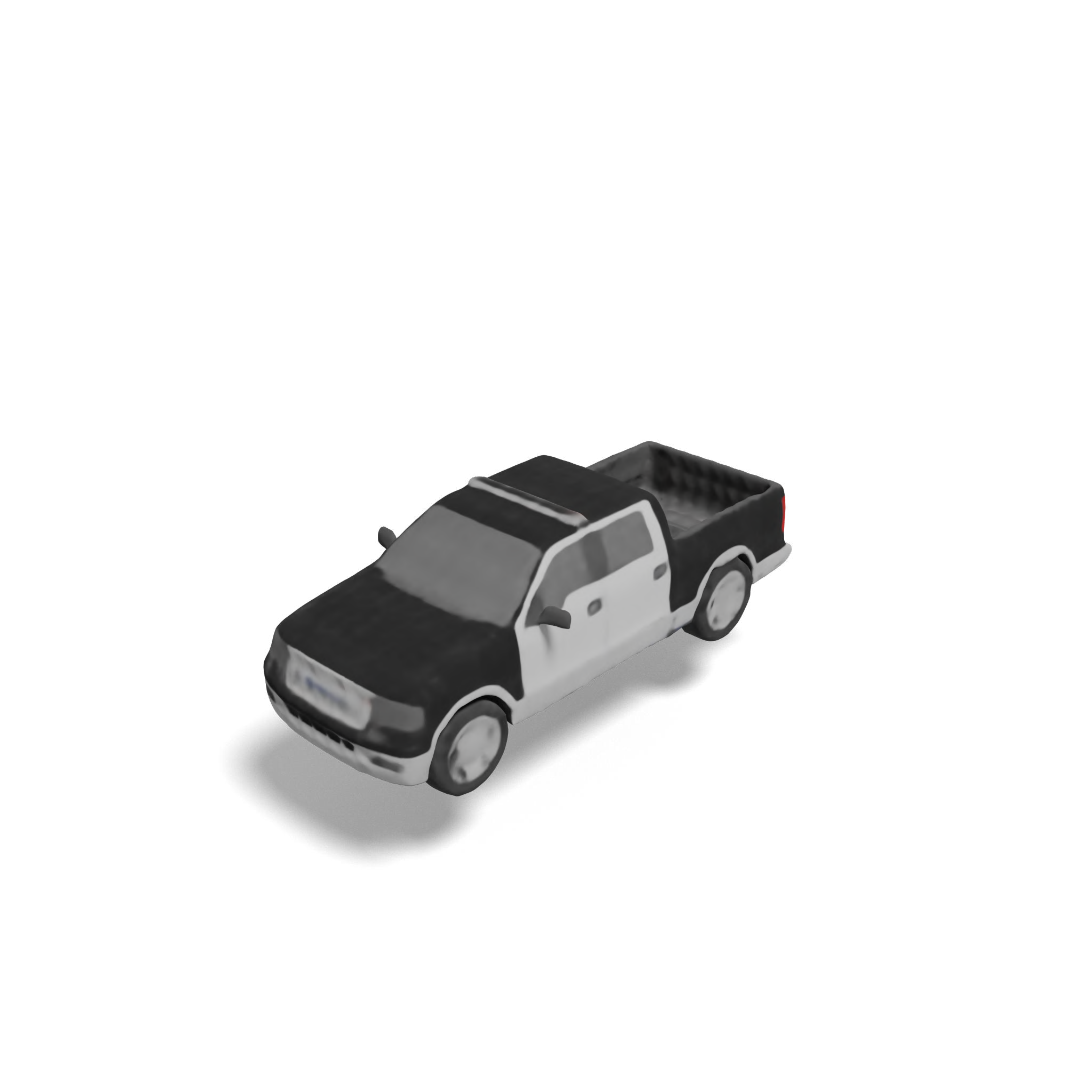}\\
    \vspace{-11pt}
    \includegraphics[width=0.12\linewidth,trim={410 400 410 410},clip]{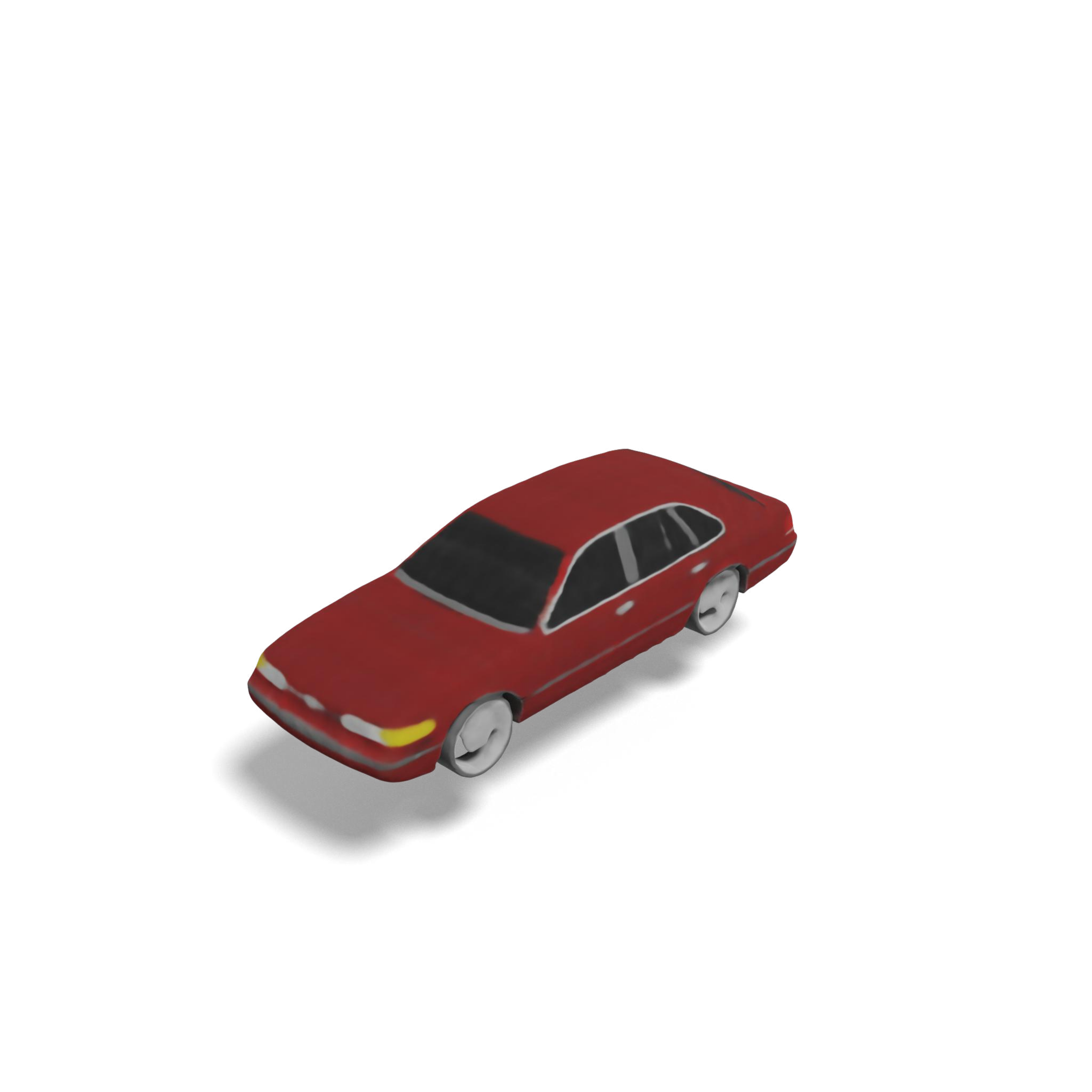}
    \includegraphics[width=0.12\linewidth,trim={410 400 410 410},clip]{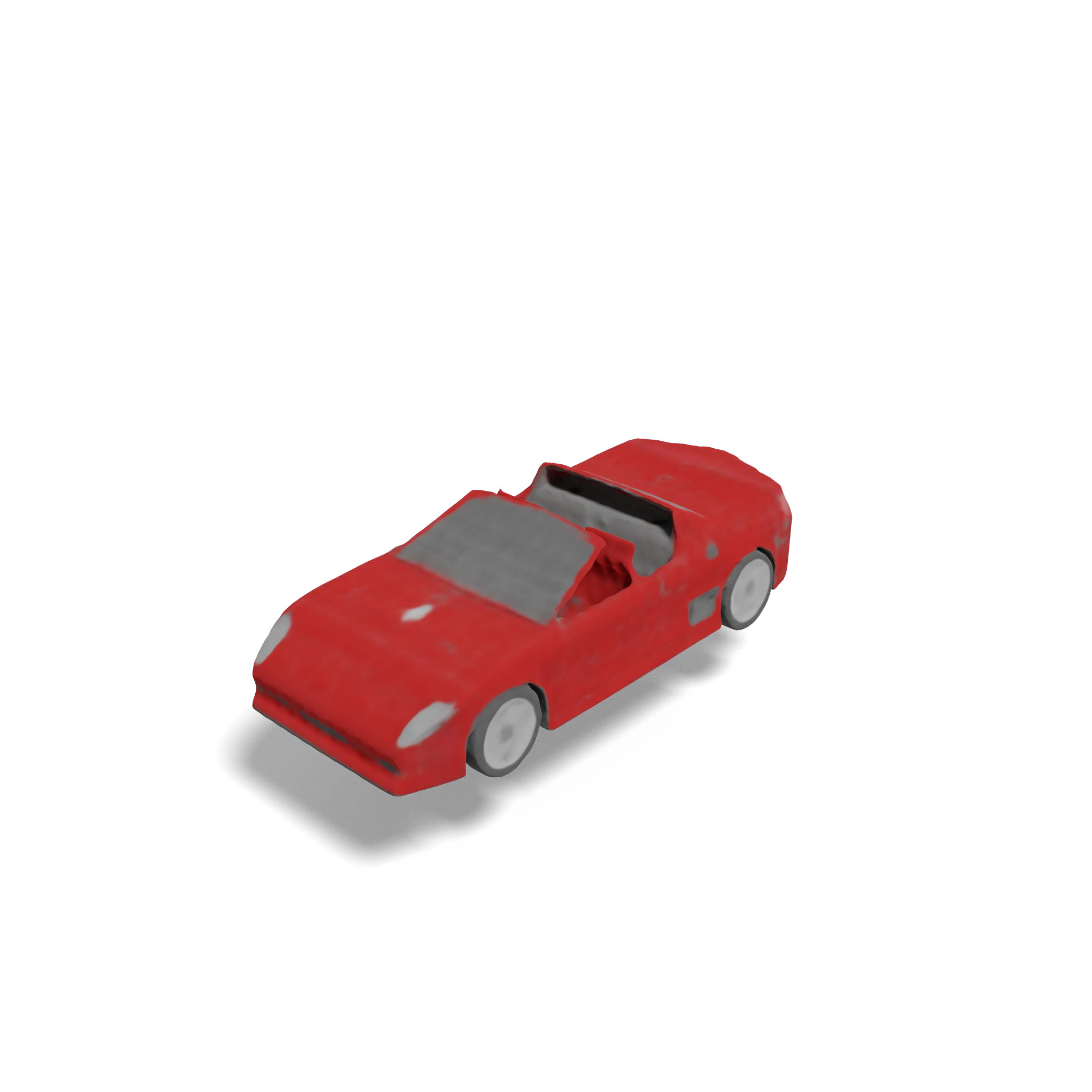}
    \includegraphics[width=0.12\linewidth,trim={410 400 410 410},clip]{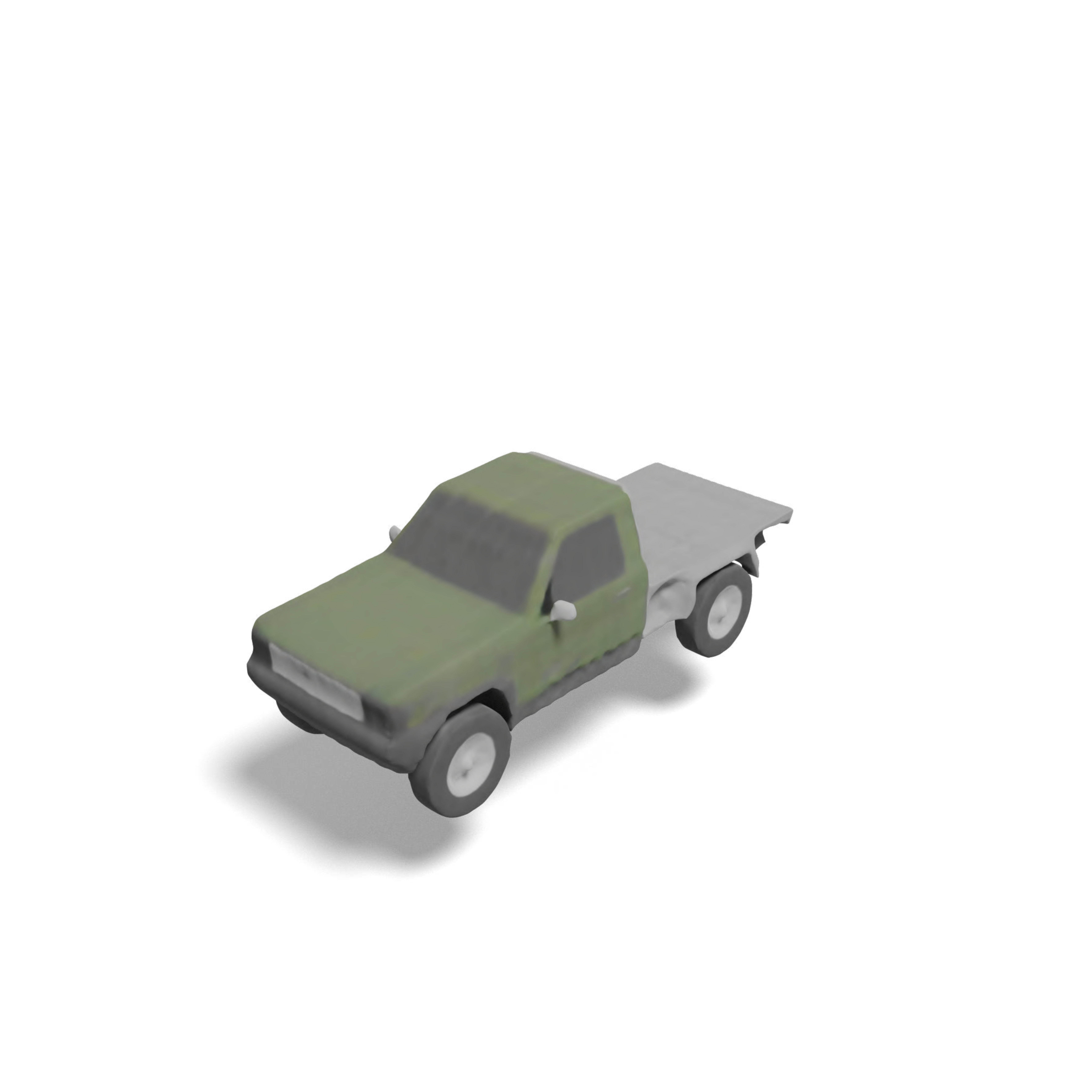}
    \includegraphics[width=0.12\linewidth,trim={410 400 410 410},clip]{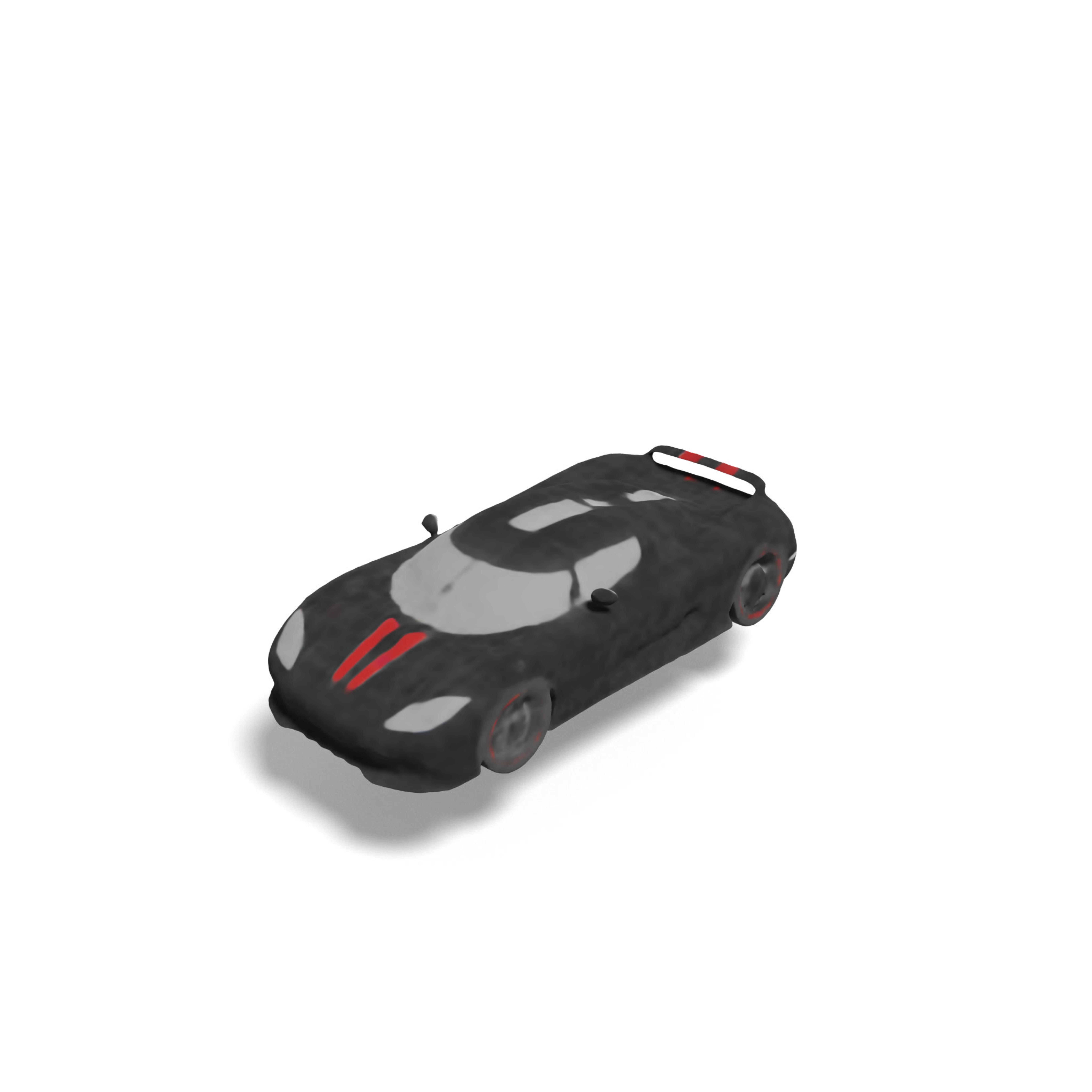}
    \includegraphics[width=0.12\linewidth,trim={410 400 410 410},clip]{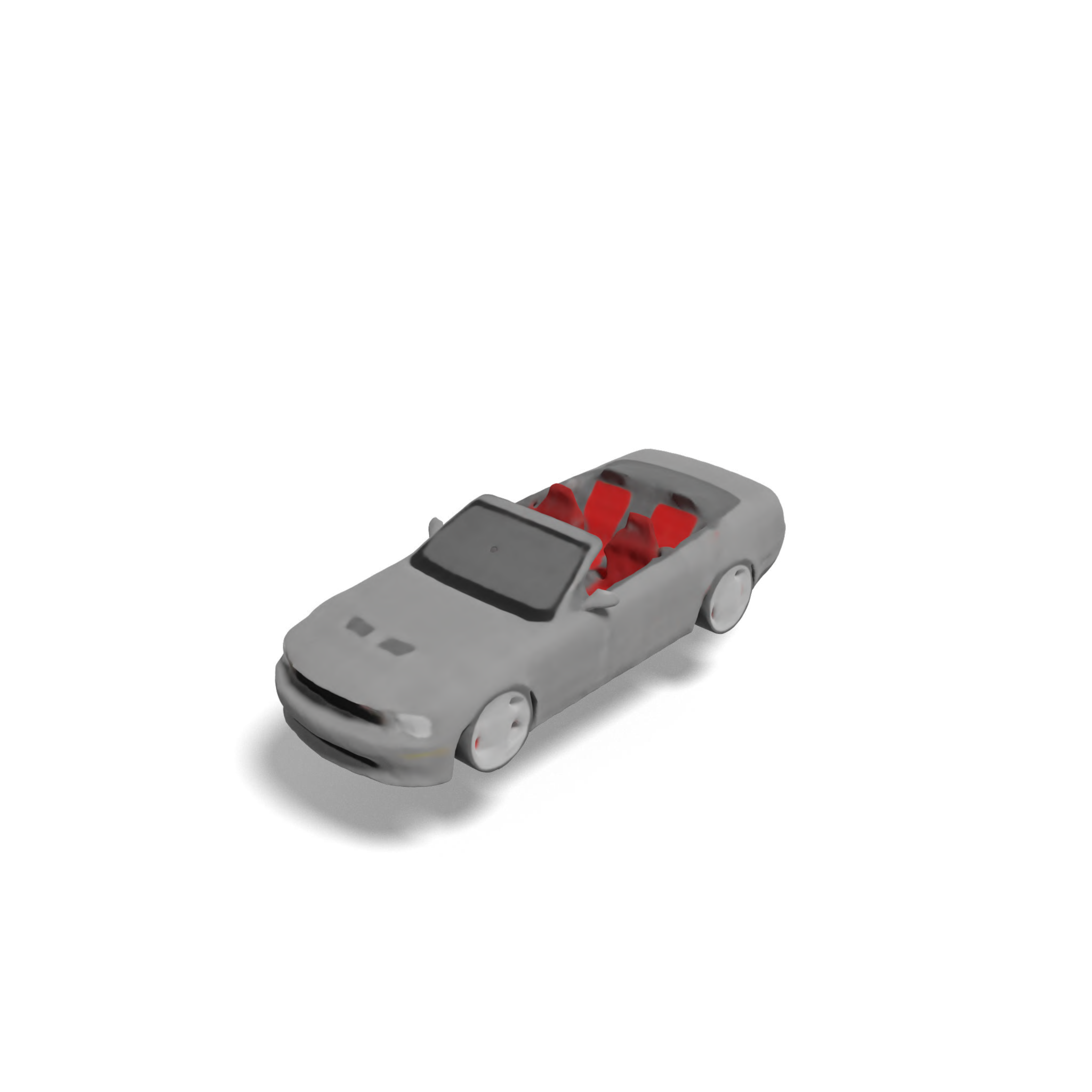}
    \includegraphics[width=0.12\linewidth,trim={410 400 410 410},clip]{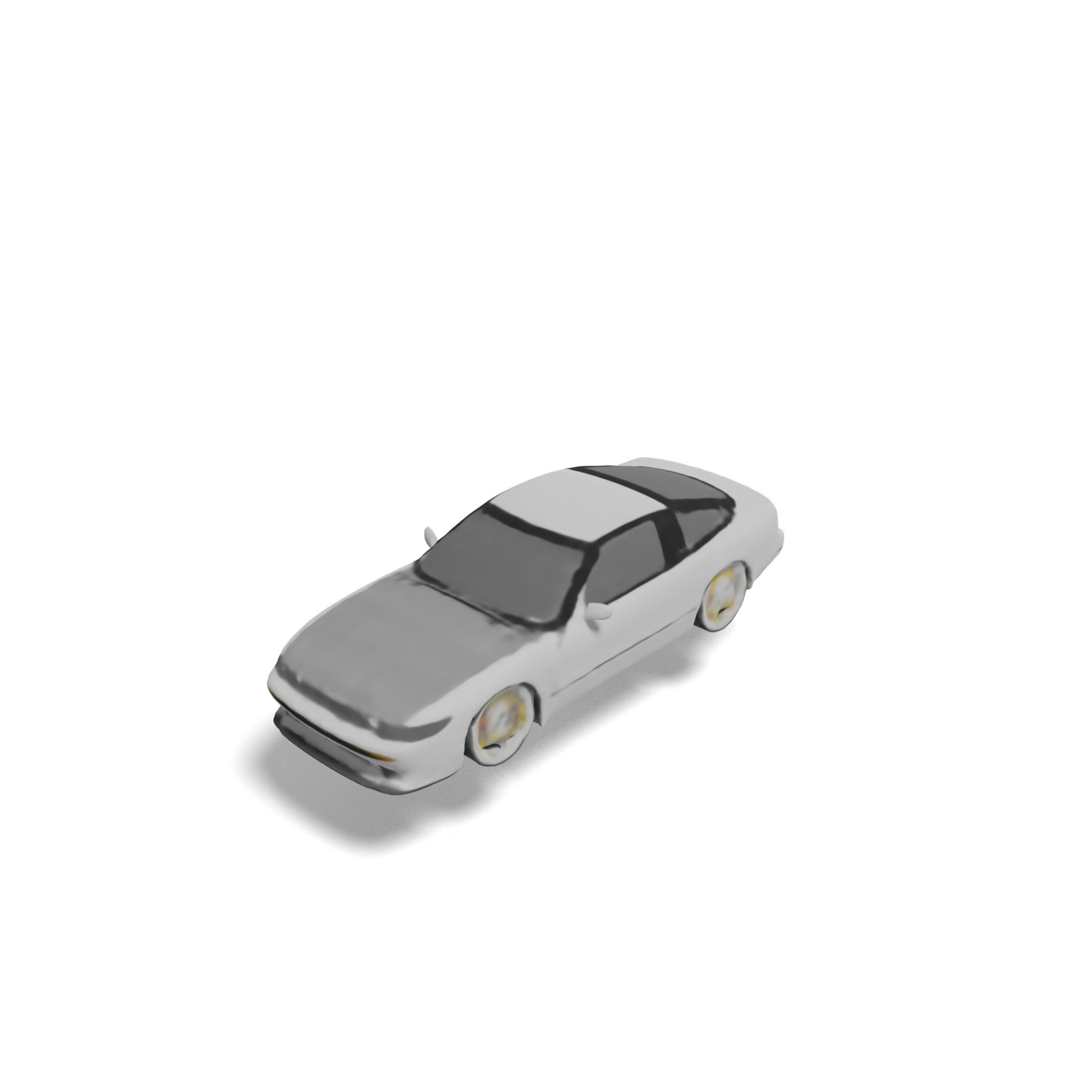}
    \includegraphics[width=0.12\linewidth,trim={410 400 410 410},clip]{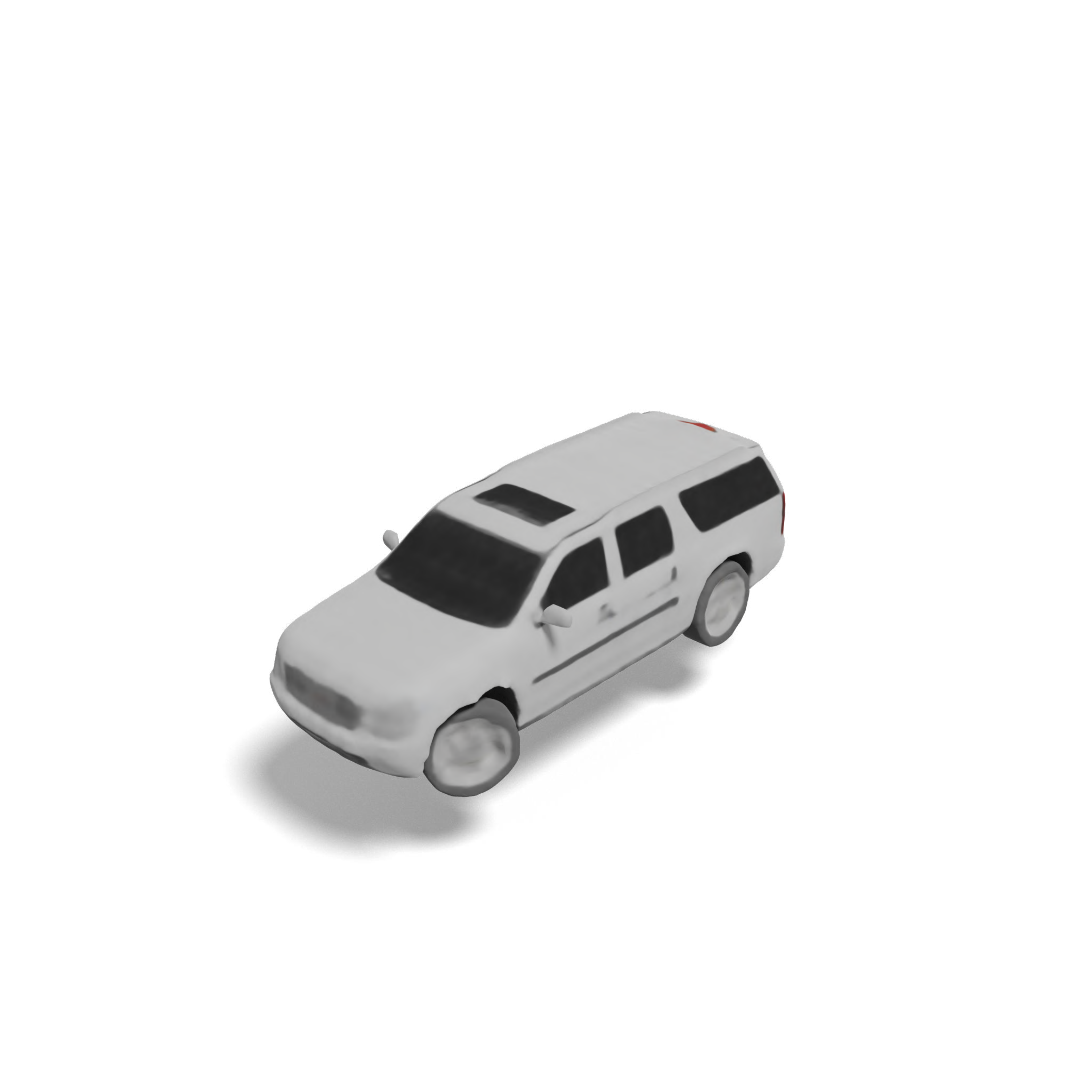}
    \includegraphics[width=0.12\linewidth,trim={410 400 410 410},clip]{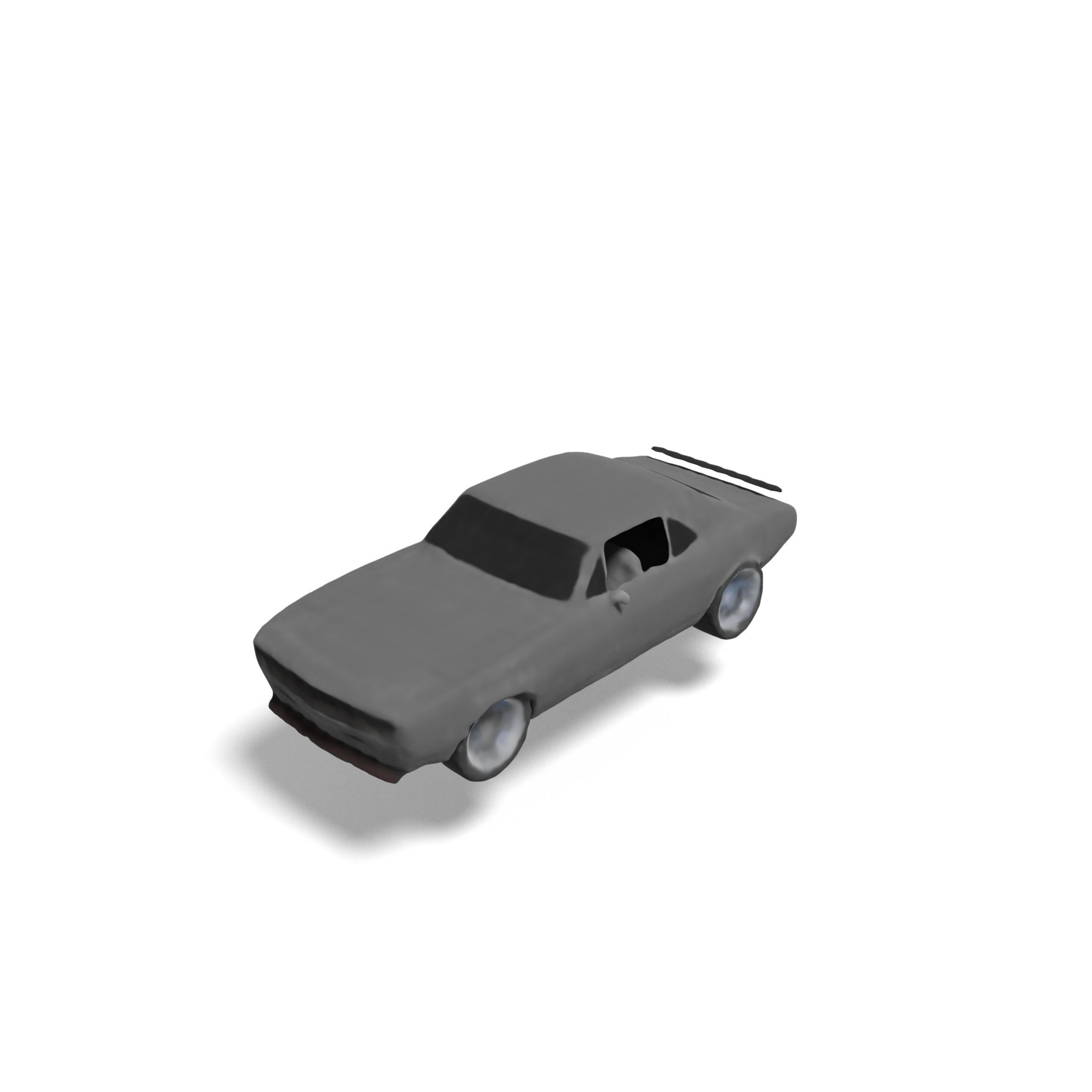}\\
    \vspace{-12pt}
    \includegraphics[width=0.12\linewidth,trim={410 400 410 410},clip]{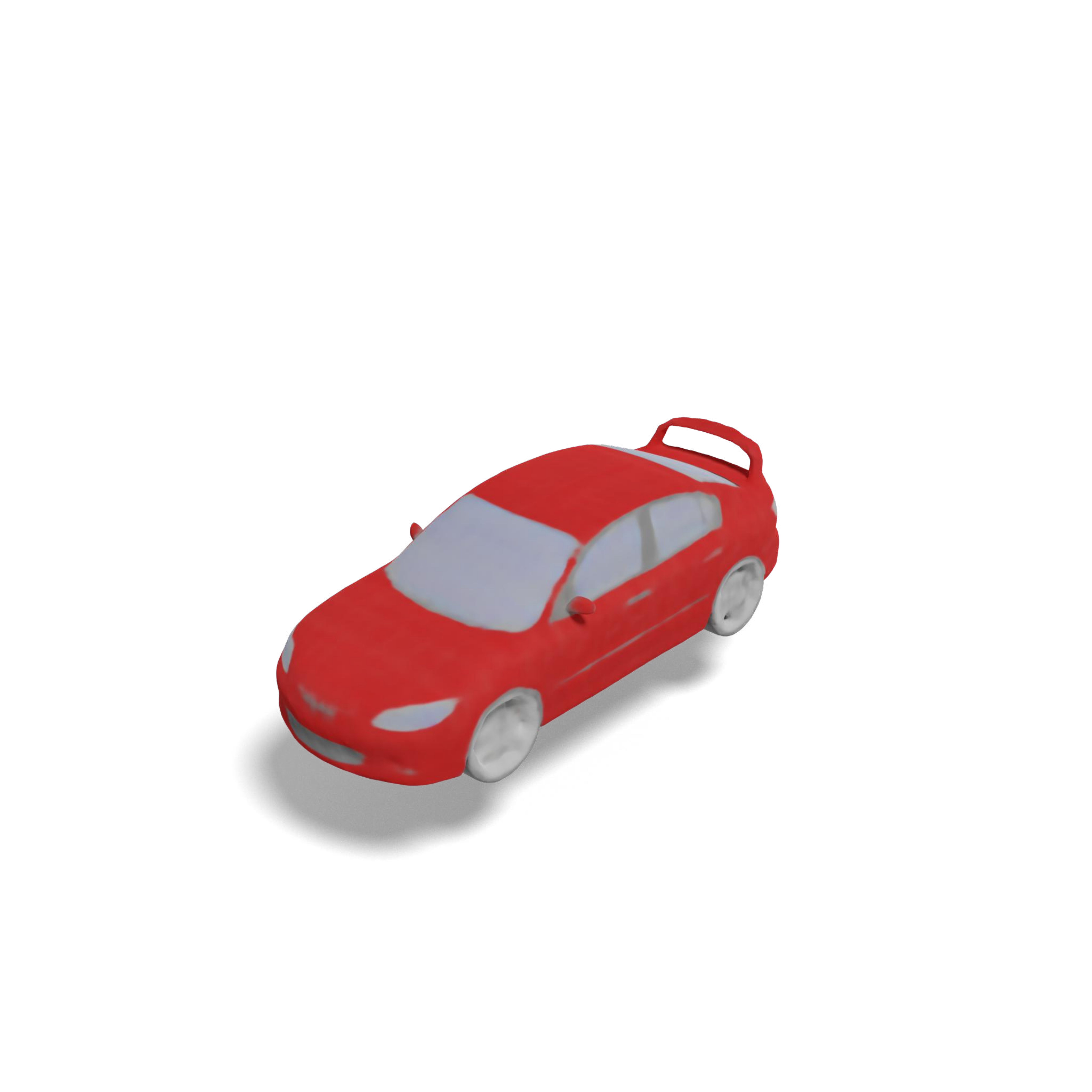}
    \includegraphics[width=0.12\linewidth,trim={410 400 410 410},clip]{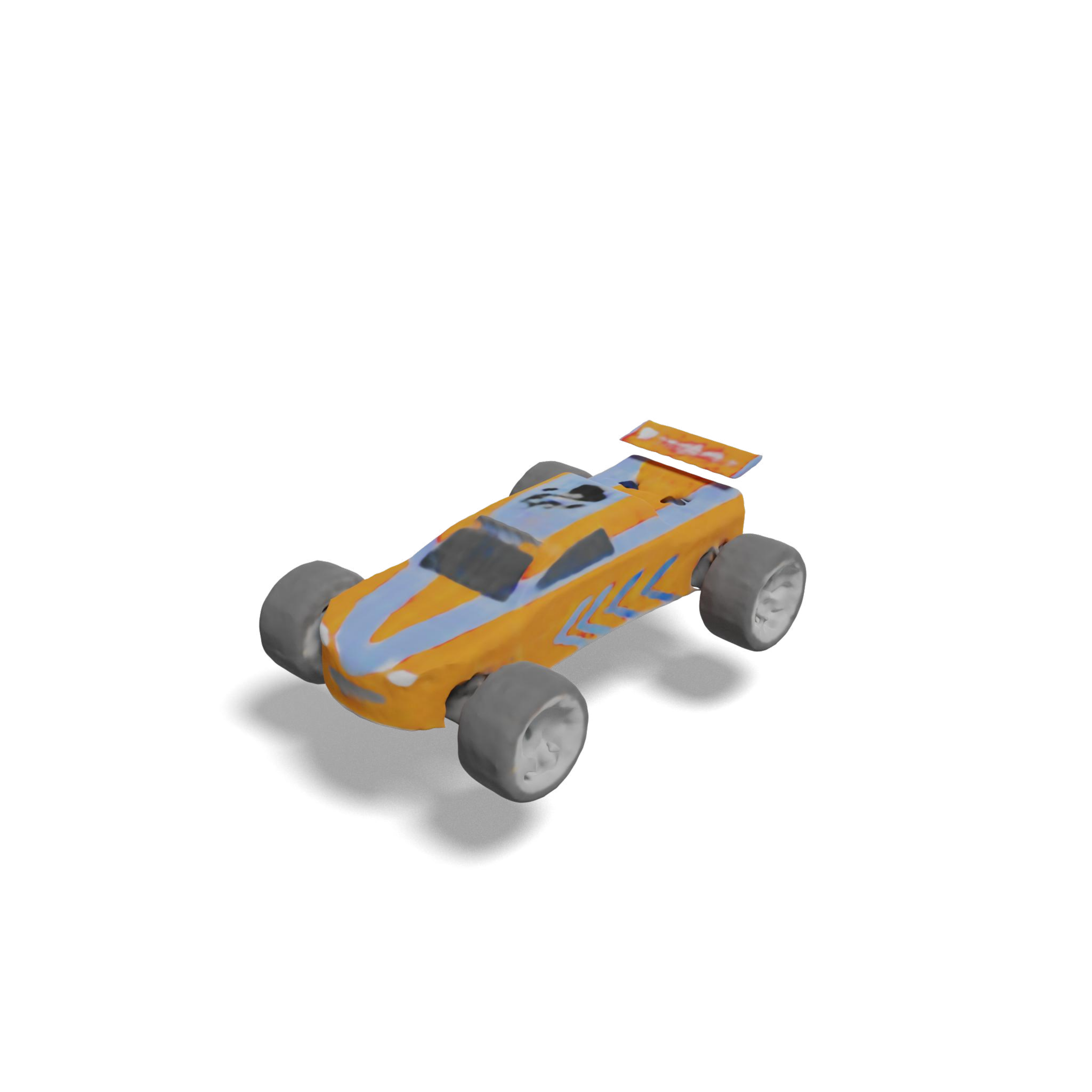}
    \includegraphics[width=0.12\linewidth,trim={410 400 410 410},clip]{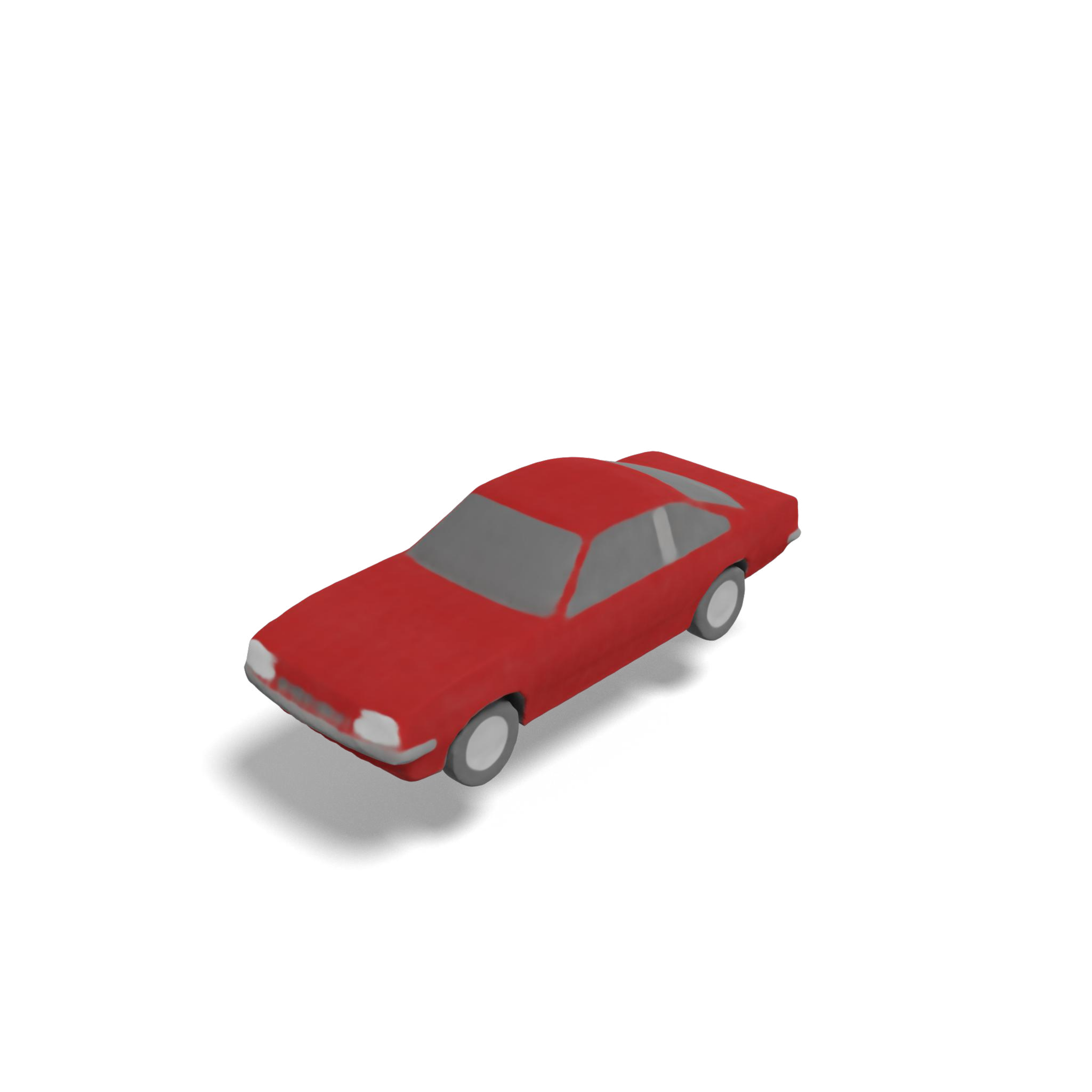}
    \includegraphics[width=0.12\linewidth,trim={410 400 410 410},clip]{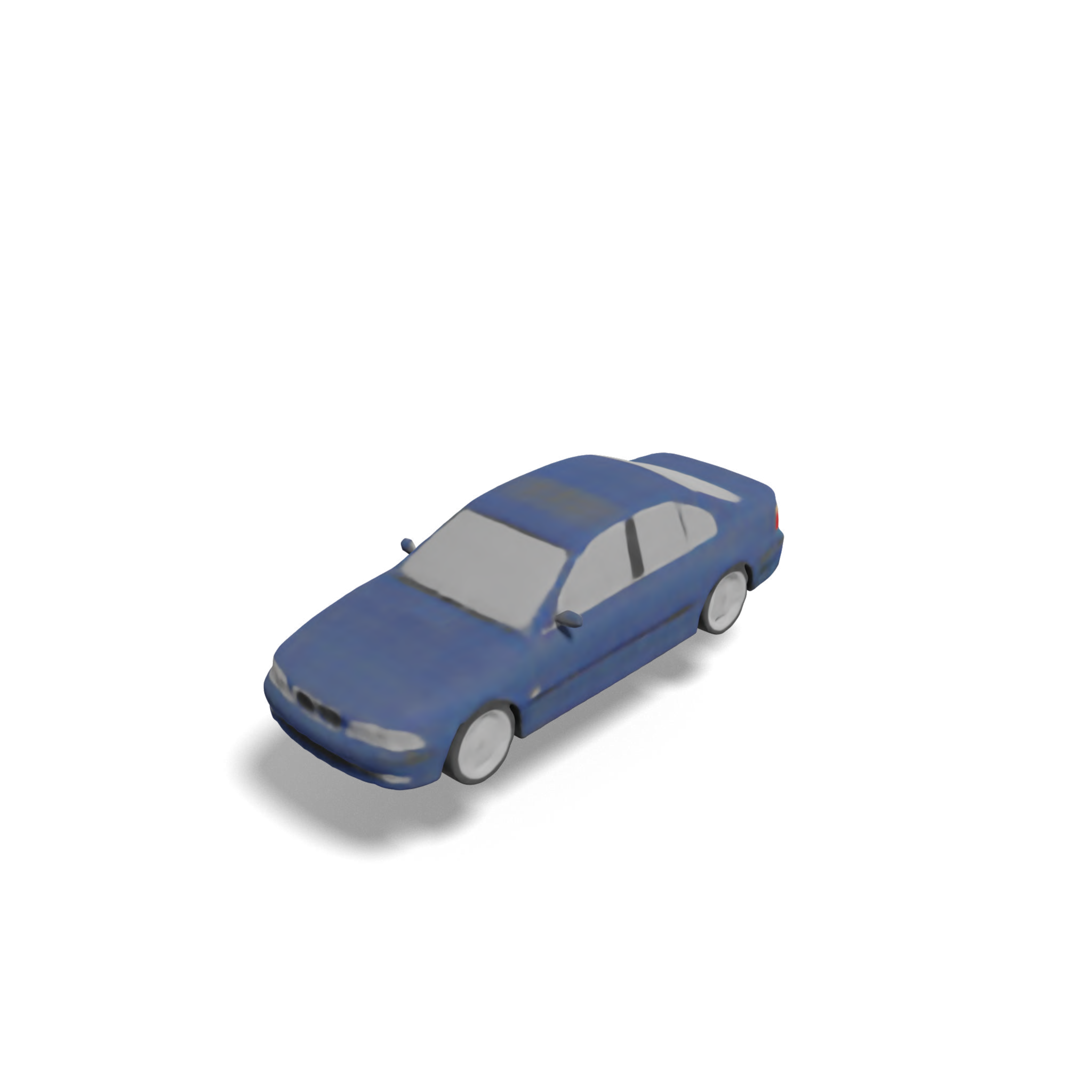}
    \includegraphics[width=0.12\linewidth,trim={410 400 410 410},clip]{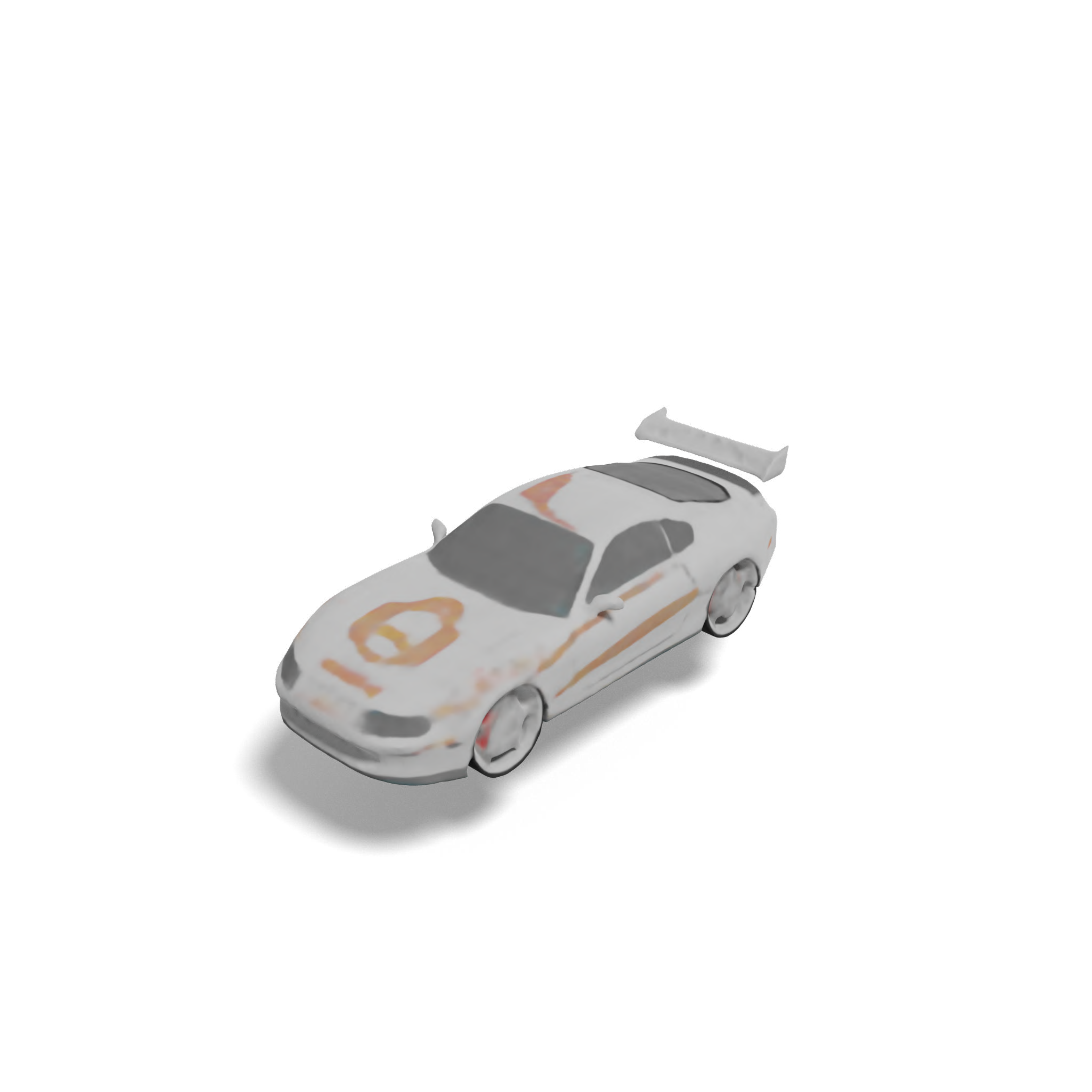}
    \includegraphics[width=0.12\linewidth,trim={410 400 410 410},clip]{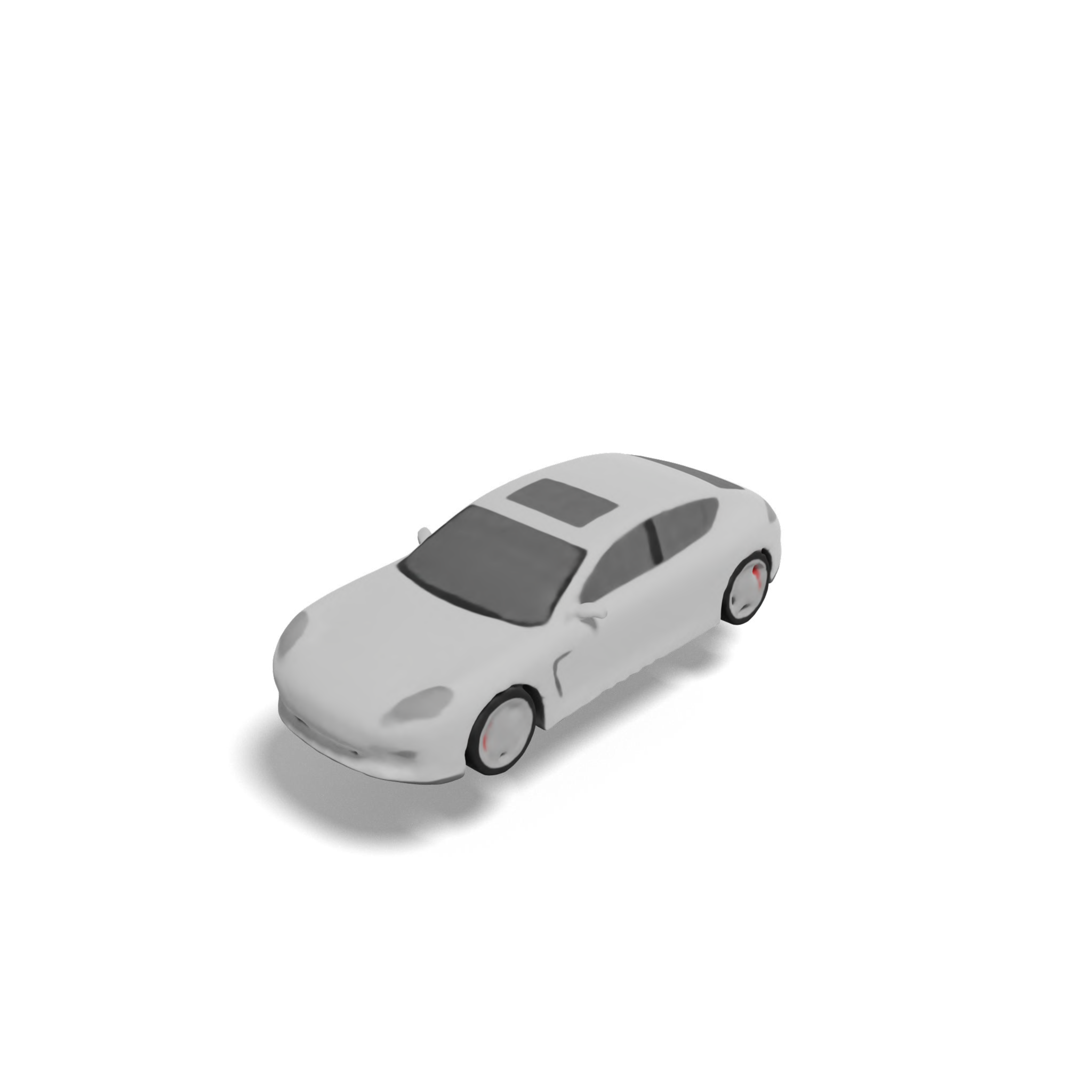}
    \includegraphics[width=0.12\linewidth,trim={410 400 410 410},clip]{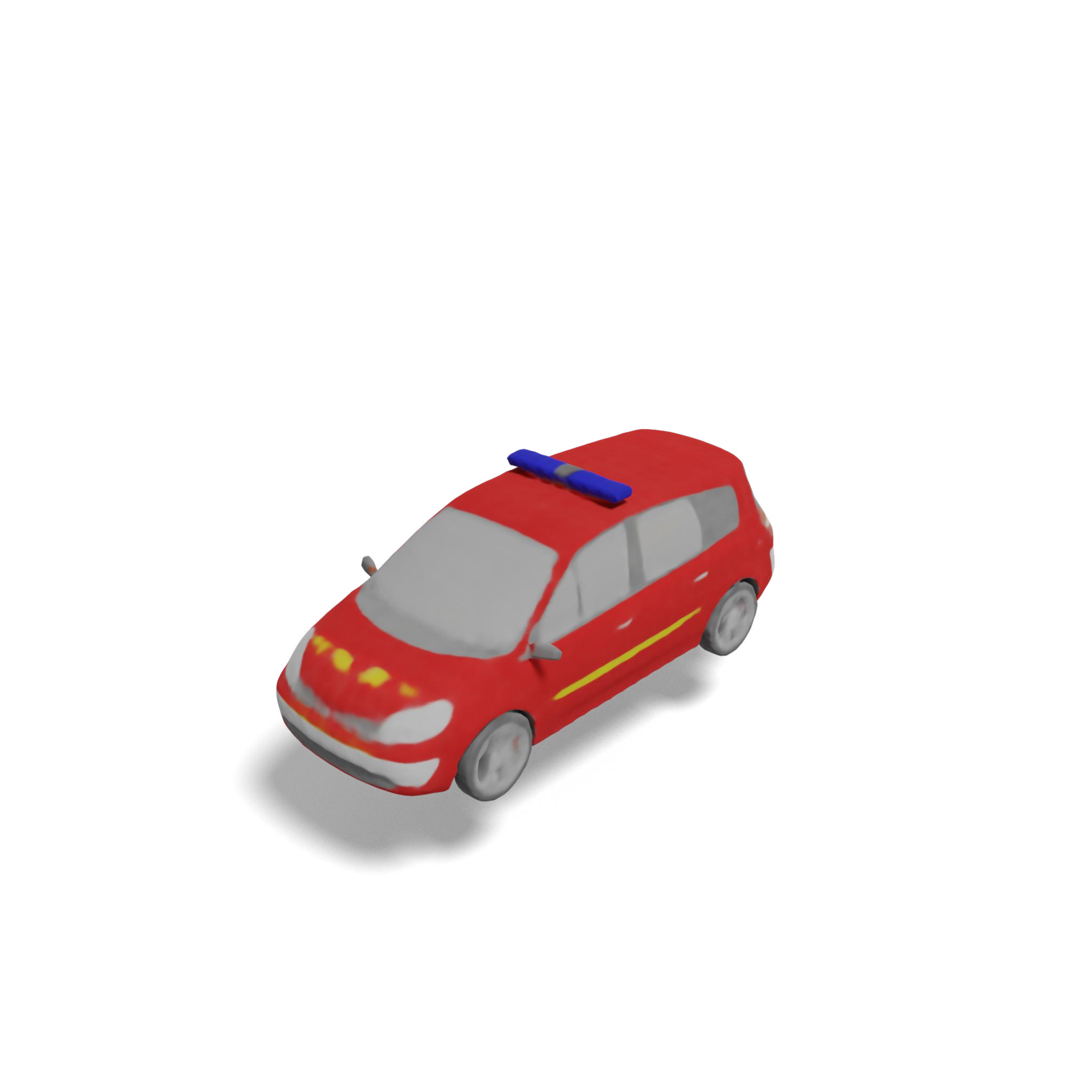}
    \includegraphics[width=0.12\linewidth,trim={410 400 410 410},clip]{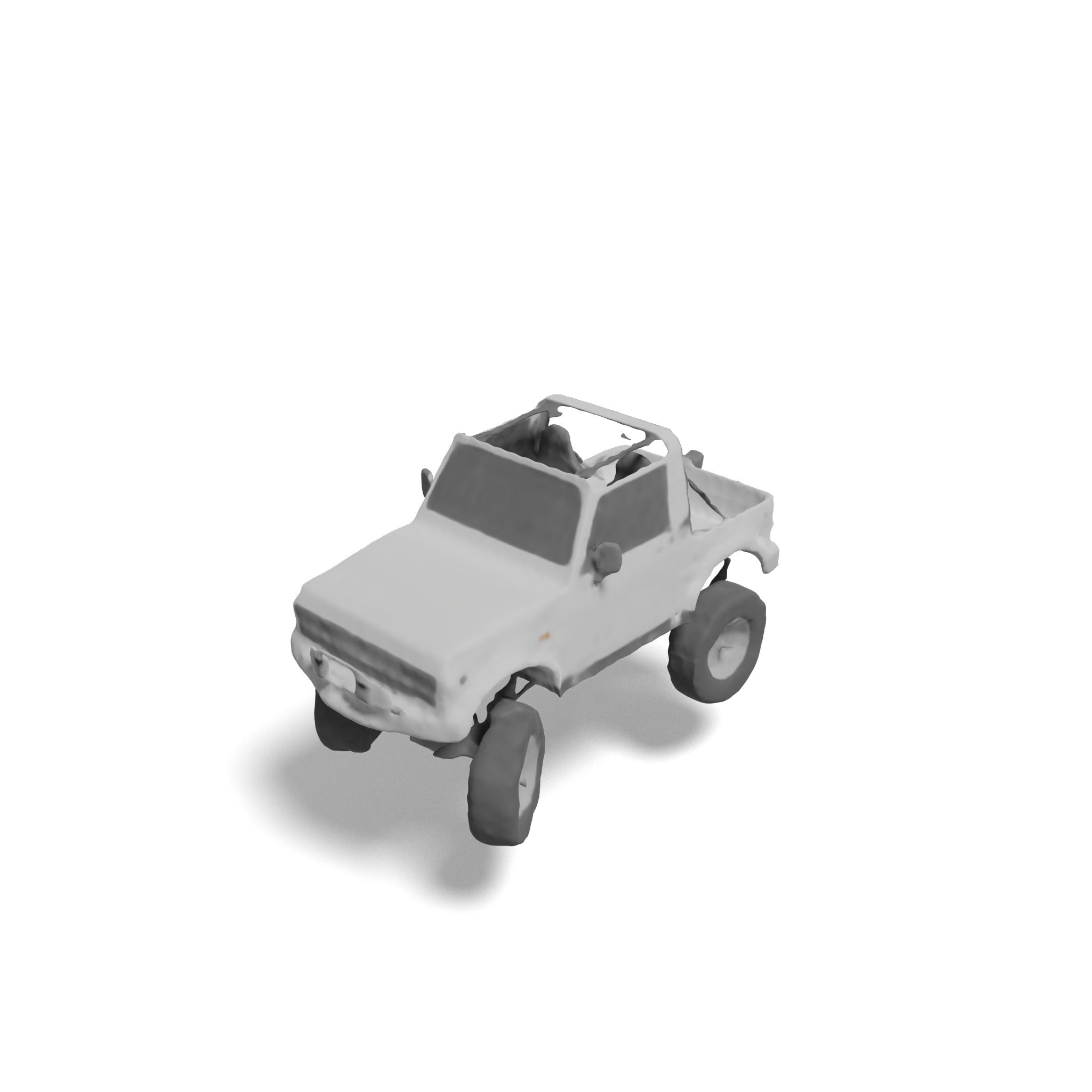}\\ \vspace{-5pt}
    % compress the image a bit to pdf to save compiling time.
    \caption{Generated meshes from our single category unconditional generation models}
    \label{fig:uncond_textured}
\end{figure*}
\begin{table}[t]
\begin{center}
\begin{tabular}{c c c}
\hline
\multirow{2}{*}{Model} & \multicolumn{2}{l}{Shading FiD($\downarrow$)} \\ 
                       & Chairs                & Planes                \\
\hline
              PVD \cite{zhou20213d}     &305.8  &244.4 \\
              SDF-StyleGAN\cite{sdfstylegan}& 36.5   & 65.8  \\
              NFD \cite{NFD}       &26.4  & 32.4\\
              \textbf{3DGen}  & \textbf{20.5}& \textbf{24.9}  \\ \bottomrule
\end{tabular}
\end{center}
\caption{Unconditional mesh geometry generation}
\label{tab:uncond_gen}
\end{table}

\begin{table}[t]
\begin{center}
\begin{tabular}{c c c c}
\toprule
\multirow{2}{*}{Model} && FiD($\downarrow$)& \\
      & Cars & Chairs  & Tables \\
\midrule
GET3D$_{flat}$ &  53.80&	36.12	&	41.40 \\\
GET3D$_{lit}$ &  40.26 &	27.42&	42.20 \\\
\textbf{3DGen} & \textbf{36.27} &	\textbf{11.07} &	\textbf{12.34}\\
\bottomrule
\end{tabular}
\end{center}
\caption{Unconditional textured mesh generation. GET3D$_{flat}$ is trained on our renders and data splits. GET3D$_{lit}$ is trained on GET3D style renders but our data.}
\label{tab:uncond_texture_gen}
\end{table}
\subsection{Unconditional generation}
We take the image conditioned diffusion model from above and finetune it for single category unconditional generation by zeroing out the image embedding. Additionally we also finetune the VAE decoder so that it can specialize in capturing category specific details.
We follow the data splits, evaluation scripts in previous works \cite{sdfstylegan, NFD} and report FiD calculated on shading renders across the chairs and planes categories. As seen in Table \ref{tab:uncond_gen}, our model outperforms strong baselines including recent GAN~\cite{sdfstylegan} and diffusion models (PVD~\cite{zhou20213d}, NFD~\cite{NFD}). Qualitative samples in Figure \ref{fig:uncond_examples} showcase the high quality and diverse outputs generated by our model.

\subsection{Pre-training on Objaverse}\label{sec:objaverse}
% Pre-training models on large sets of data consistently shows big improvements in both accuracy and robustness of downstream fine-tuned models across domains like computer vision and language processing. 
Pre-training on large datasets has proven to be an effect recipe for achieving significant performance boost across a wide range of vision and language tasks.
We show how a similar effect can be achieved in 3D mesh generation by pre-training our model on the large Objaverse dataset which is much more diverse than the commonly used benchmark.
We pre-train both our VAE and the image conditional diffusion model on the combined Objaverse and 3D Warehouse data. The diffusion model is further finetuned on the 3D Warehouse dataset. To measure the impact we report improvements in both the average Chamfer Distance and FiD metrics in Table \ref{tab:cond_gen}. The pre-trained model is more faithful to the input image and also has improved visual quality as evident from the lower FiD scores. Notably, we observe that pre-training can bring more significant improvements on low-resource categories as demonstrated by the FiD scores.
% \vspace{-0.3cm}
\paragraph{Text conditioned generation} Our model architecture allows using text to condition generation by swapping the image embedding with a text embedding coming from the bi-encoder similar to the approach in \cite{clip_forge}. As visible in Figure \ref{fig:textcond_examples} our model is able to generate well formed objects matching the text description. We use our best image conditioned diffusion(geometry only) model for this experiment. Though this simple form of text prompting is not always robust it showcases the quality and flexibility of our model.

\subsection{Textured Mesh Generation}

As described in Section~\ref{sec:data}, we extend our VAE model for texture prediction on the mesh surface. When training the textured-version of the VAE model, 20k colored points are sampled from the pre-computed point cloud as the input $x$ of VAE encoder $\psi_{enc}$ and another set of 60k point are sampled as $c^{gt}_{x}$ for calculating $L_{color}$. We sample a large number of points particularly for the reference set as we observe that this can substantially improve texture quality.

Our texture diffusion model uses a $128\times128$ triplane resolution as we found this improved fidelity. Besides the above we follow the same training setup and hyper-parameters as in our mesh geometry only model for both VAE and diffusion. We train unconditional and conditional textured mesh generation models.
\paragraph{Unconditional Generation} For unconditioned generation, we train a single category VAE and diffusion model on the car, chair and table categories separately. We compare our model with GET3D by computing the FID scores between renders of the generated textured meshes and ground truth meshes. 
Following GET3D's evaluation protocol, the combined set of all train, val, and test renders are used for this evaluation.  We use GET3D's lighting and rendering setup with a set of randomly sampled camera angels to calculate FID scores. This can slightly favor GET3D output, since the same renders were used during training. Also unlike GET3D we don't do any category specific hyper-parameter tuning. Despite this, our 3DGen model outperforms GET3D on all three categories. In particular, we improve FID from $27.4$ to $11.1$ on chairs and from $42.2$ to $12.3$ on tables (Table~\ref{tab:uncond_texture_gen}).  As shown in Figure~\ref{fig:uncond_textured}, the meshes generated by our model are high quality and capture both geometry and texture details.
% Since FID scores are sensitive to the render method, for a fair comparison we render all meshes using Blender with the same lighting setup as in the GET3D code and choose random camera locations on a sphere looking towards the origin. 
In addition to GET3D$_{lit}$ trained on renders with realistic lighting, we also compare to a GET3D$_{flat}$ model trained on our renders with flat lighting for a like for like comparison, this even more so shows the advantage of 3DGen.
% \vspace{-0.2cm}
\begin{figure*}[t]
    \centering
    \includegraphics[width=0.012\linewidth,trim={0 50 0 0 },clip]{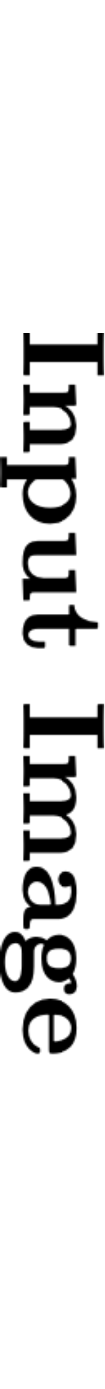}
    \includegraphics[width=0.118\linewidth,trim={30 30 30 30},clip]{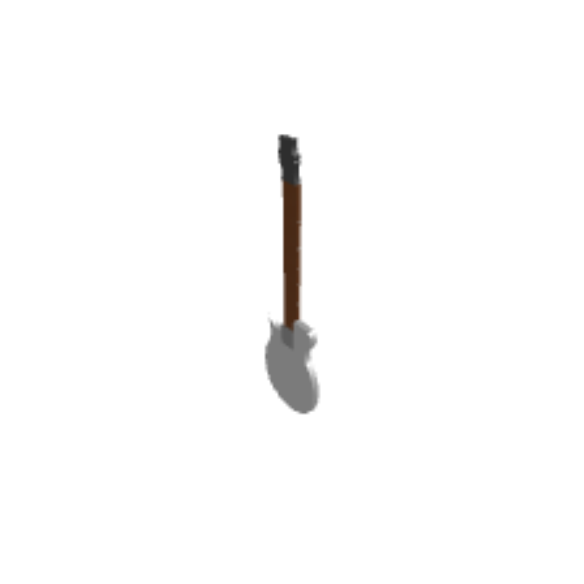}
    \includegraphics[width=0.118\linewidth,trim={30 30 30 30},clip]{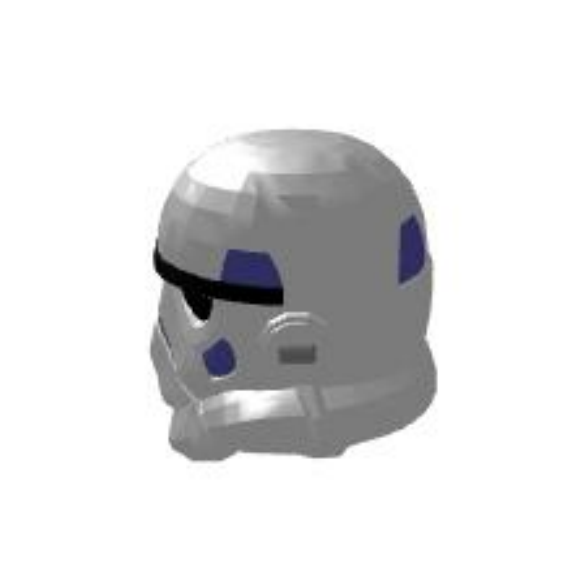}
    \includegraphics[width=0.118\linewidth,trim={30 30 30 30},clip]{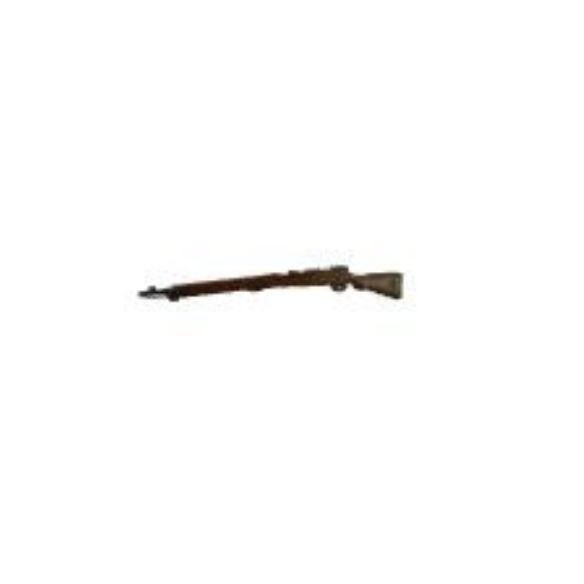}
    \includegraphics[width=0.118\linewidth,trim={30 30 30 30},clip]{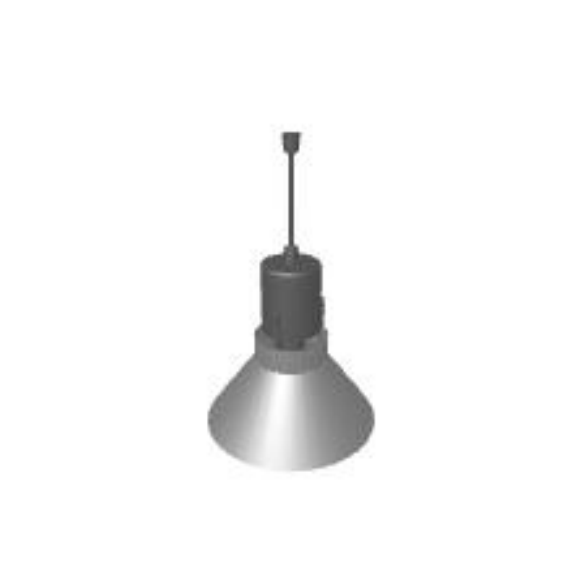}
        \includegraphics[width=0.118\linewidth,trim={30 30 30 30},clip]{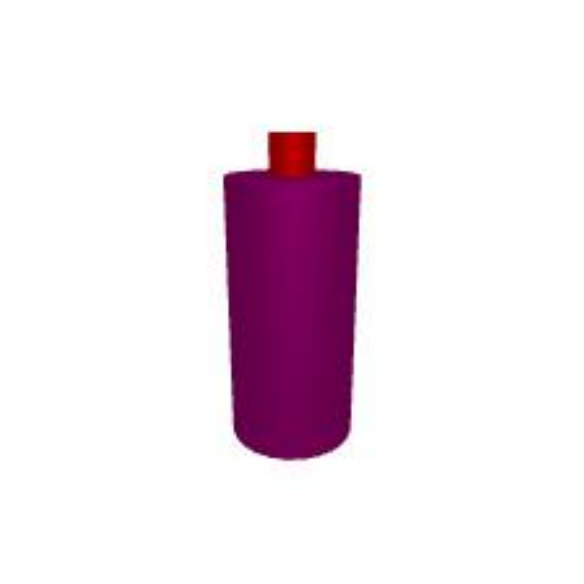}
   \includegraphics[width=0.118\linewidth,trim={30 30 30 30},clip]{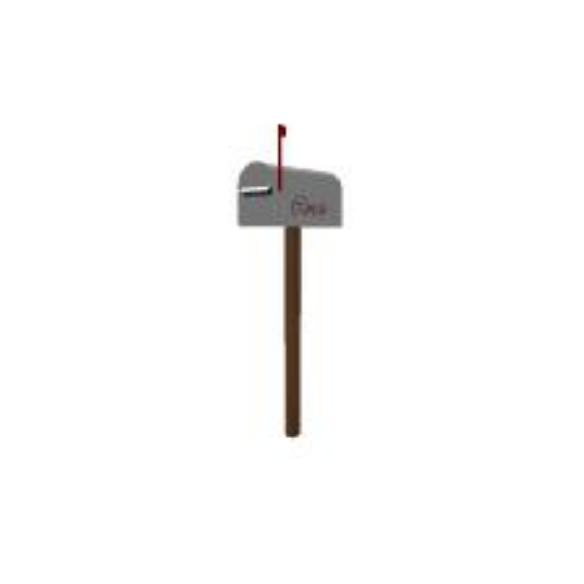}
       \includegraphics[width=0.118\linewidth,trim={30 30 30 30},clip]{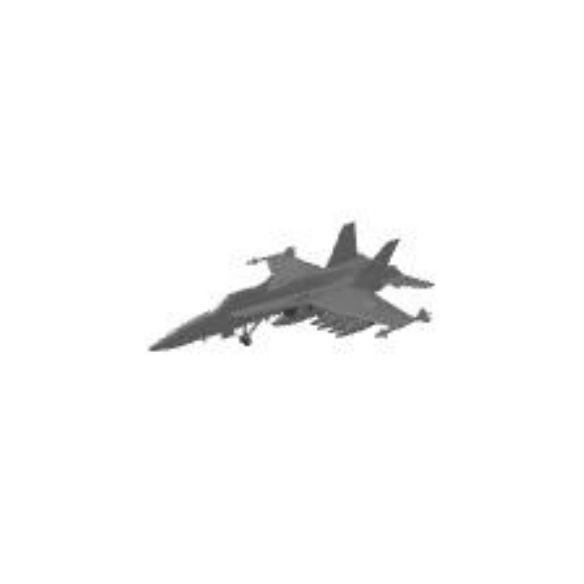}
    \includegraphics[width=0.118\linewidth,trim={30 30 30 30},clip]{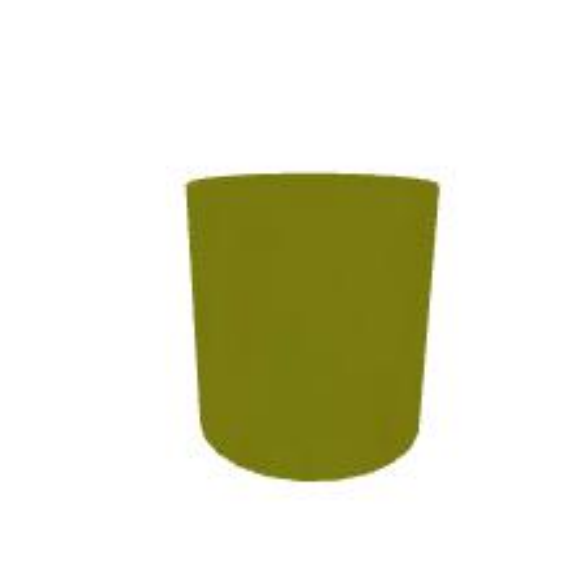} \\ \vspace{-10pt}
    \includegraphics[width=0.012\linewidth,trim={0 0 0 0 },clip]{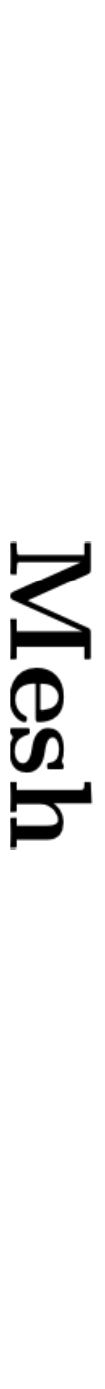}
    \includegraphics[width=0.118\linewidth,trim={400 300 400 300},clip]{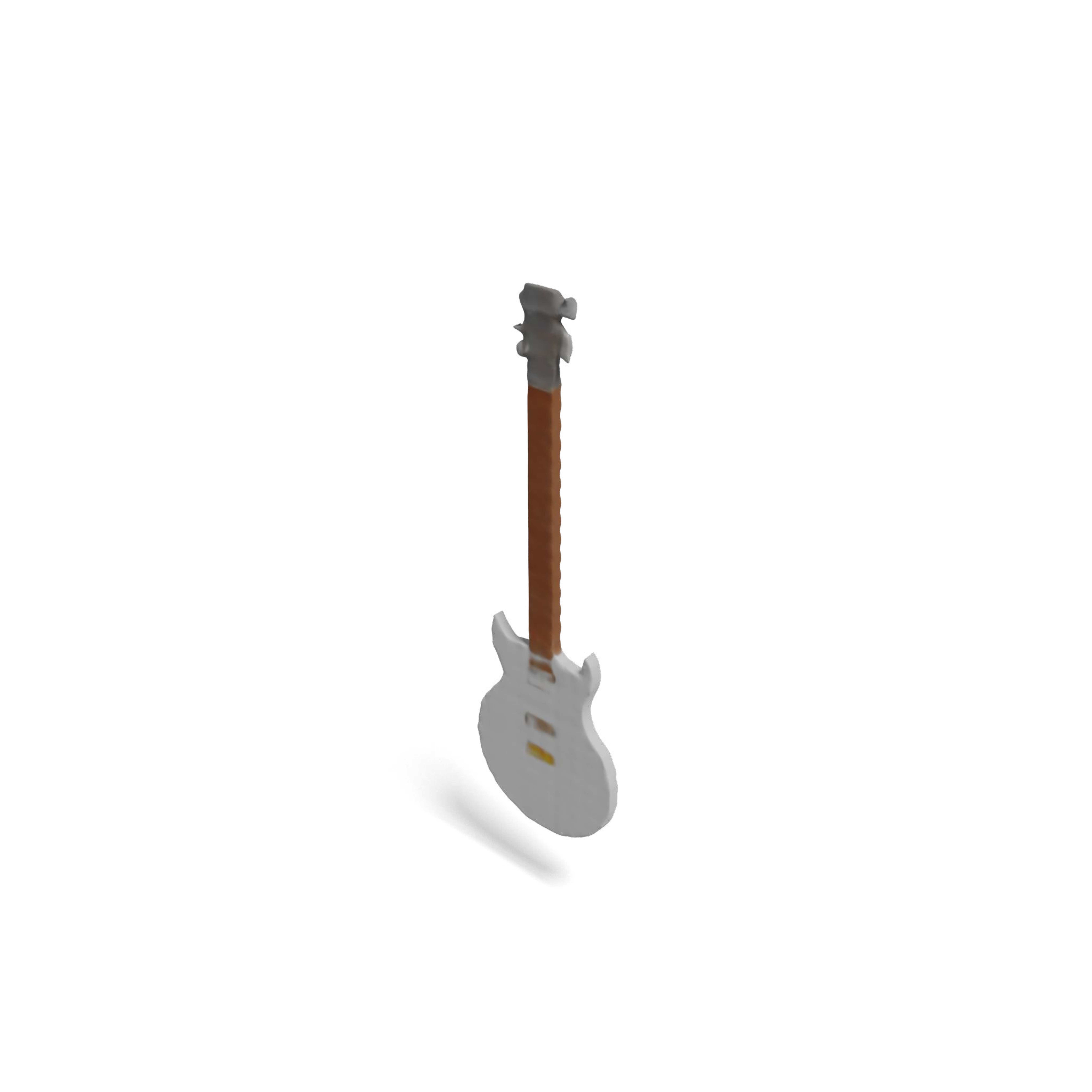}
    \includegraphics[width=0.118\linewidth,trim={400 200 400 400},clip]{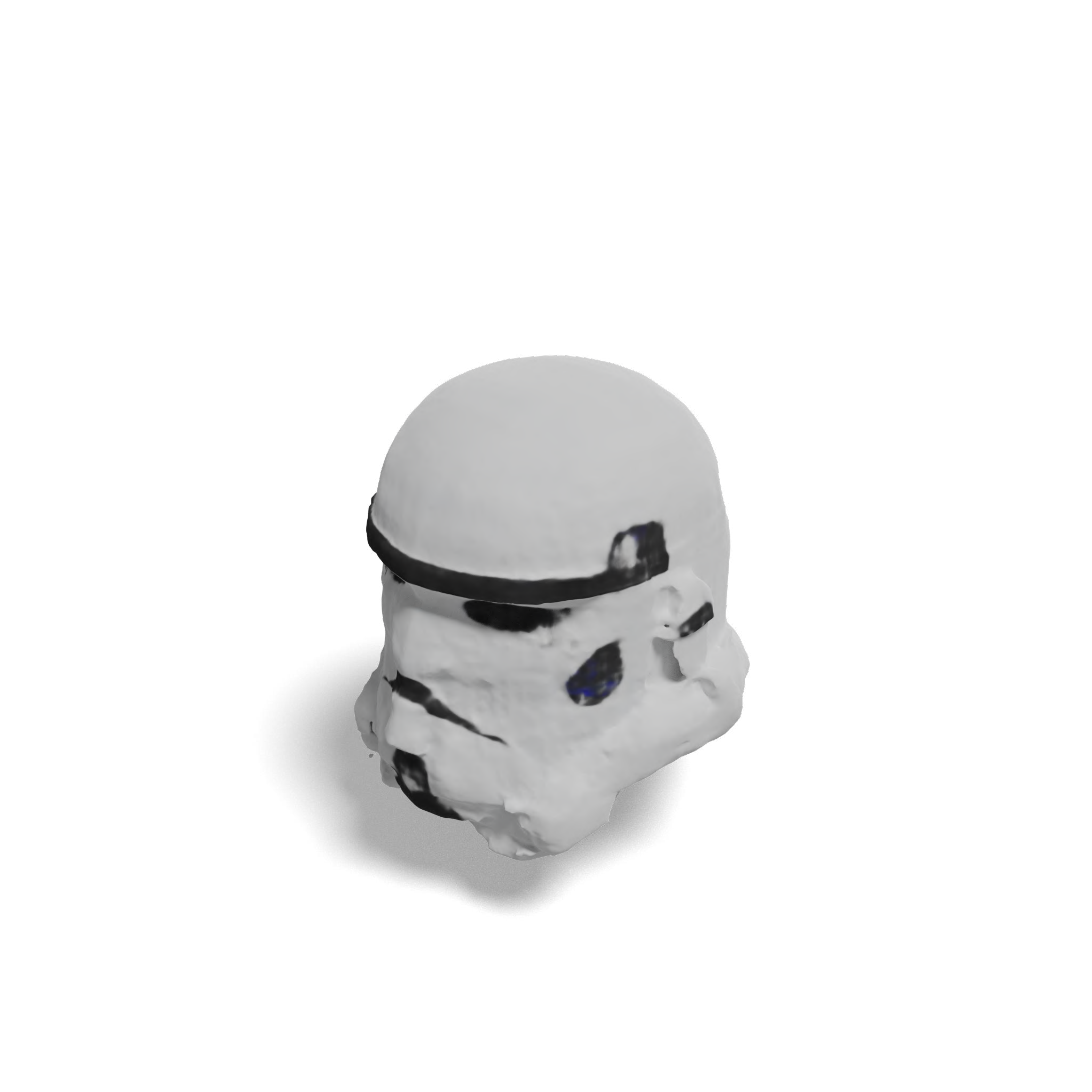}
    \includegraphics[width=0.118\linewidth,trim={300 400 300 900},clip]{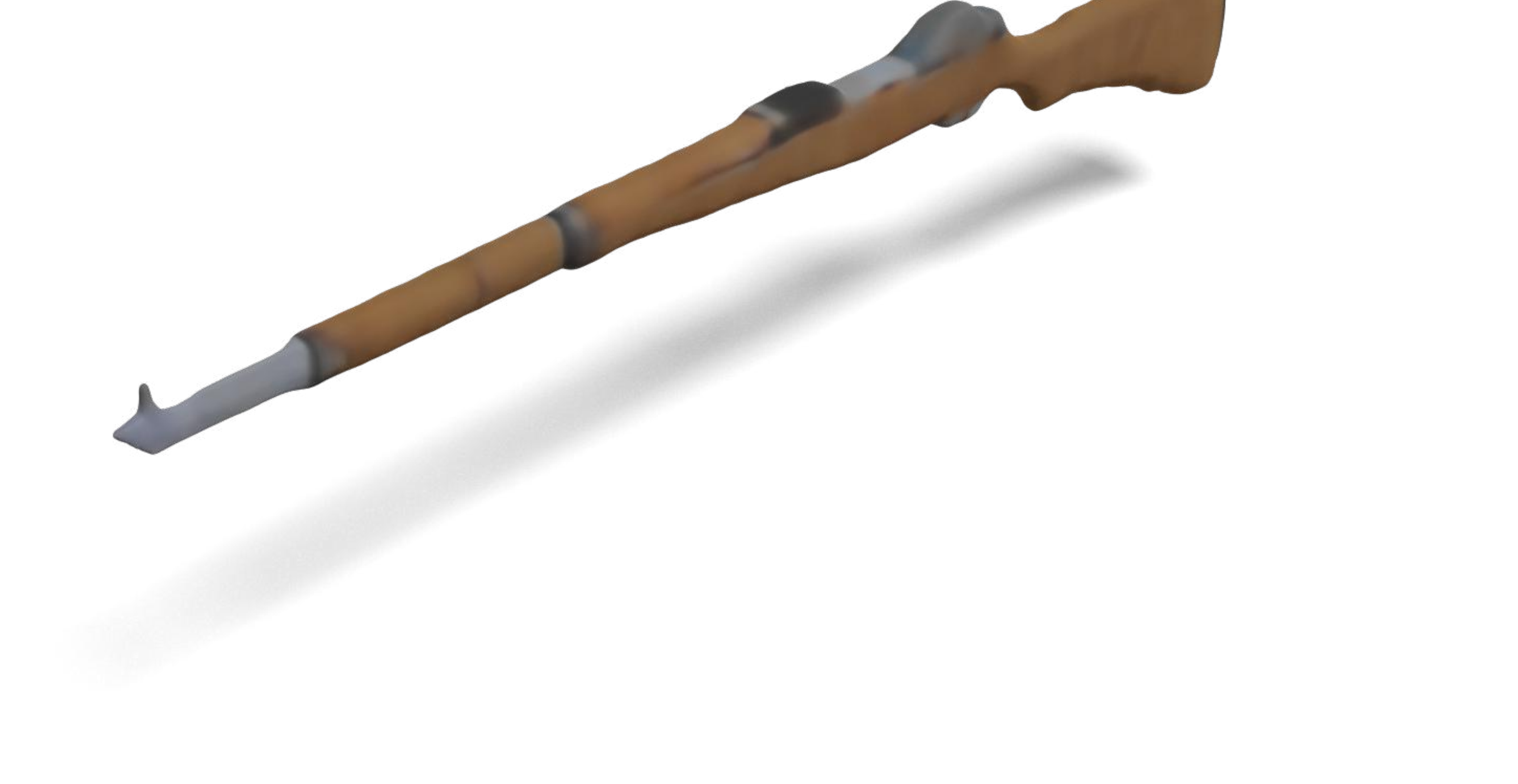}
    \includegraphics[width=0.118\linewidth,trim={300 200 300 320},clip]{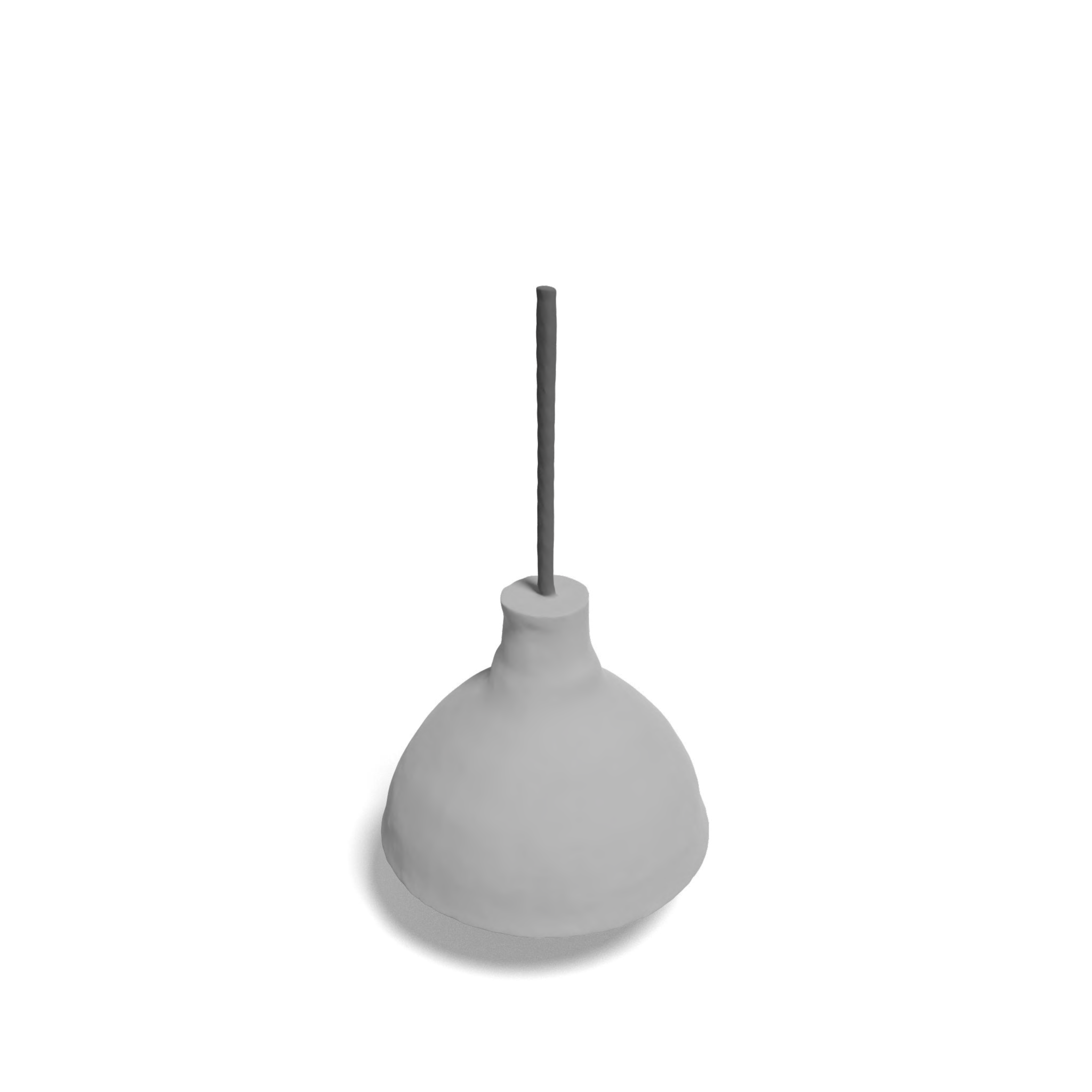}
    \includegraphics[width=0.118\linewidth,trim={300 300 300 300},clip]{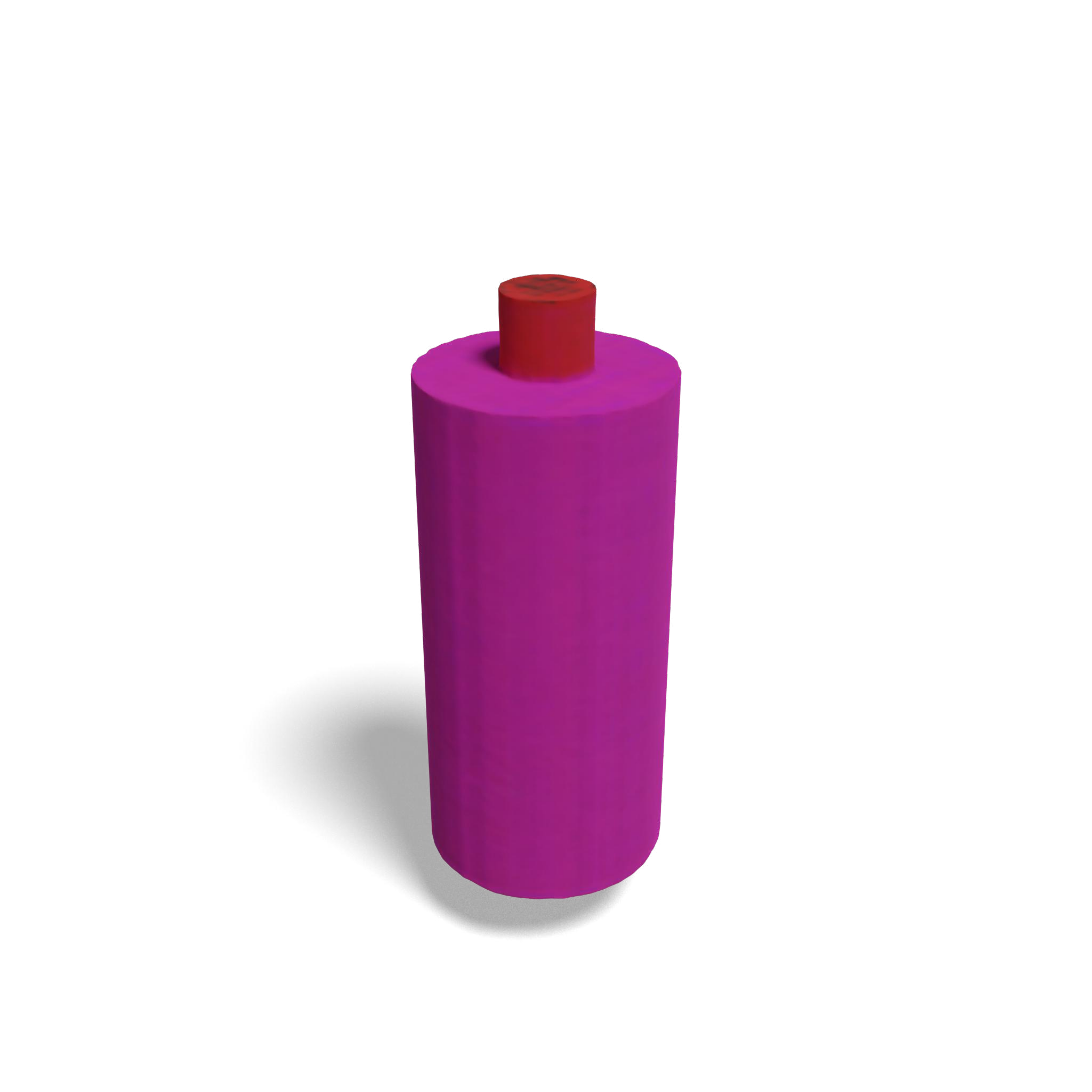}
    \includegraphics[width=0.118\linewidth,trim={420 300 420 300},clip]{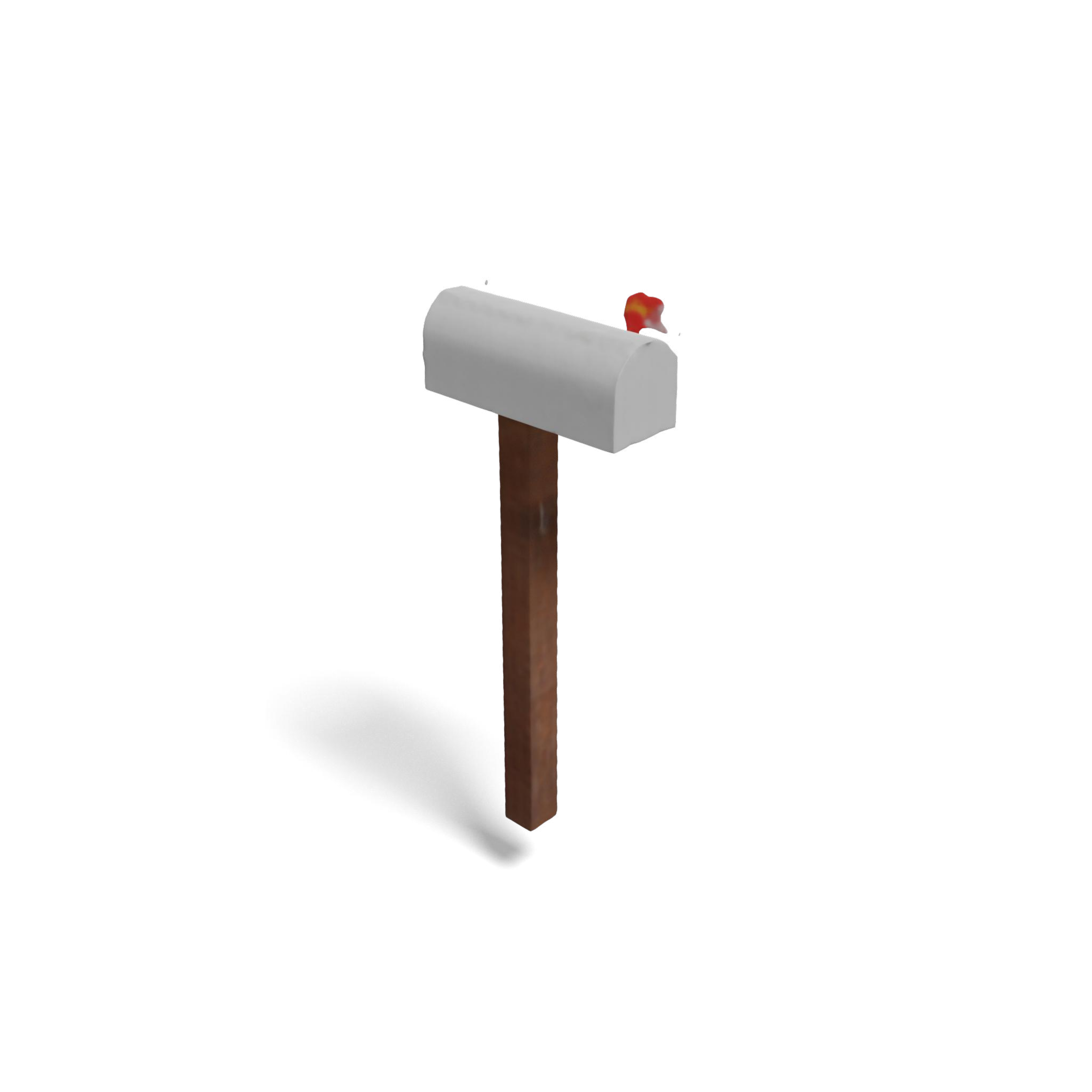}
    \includegraphics[width=0.118\linewidth,trim={600 500 600 600},clip]{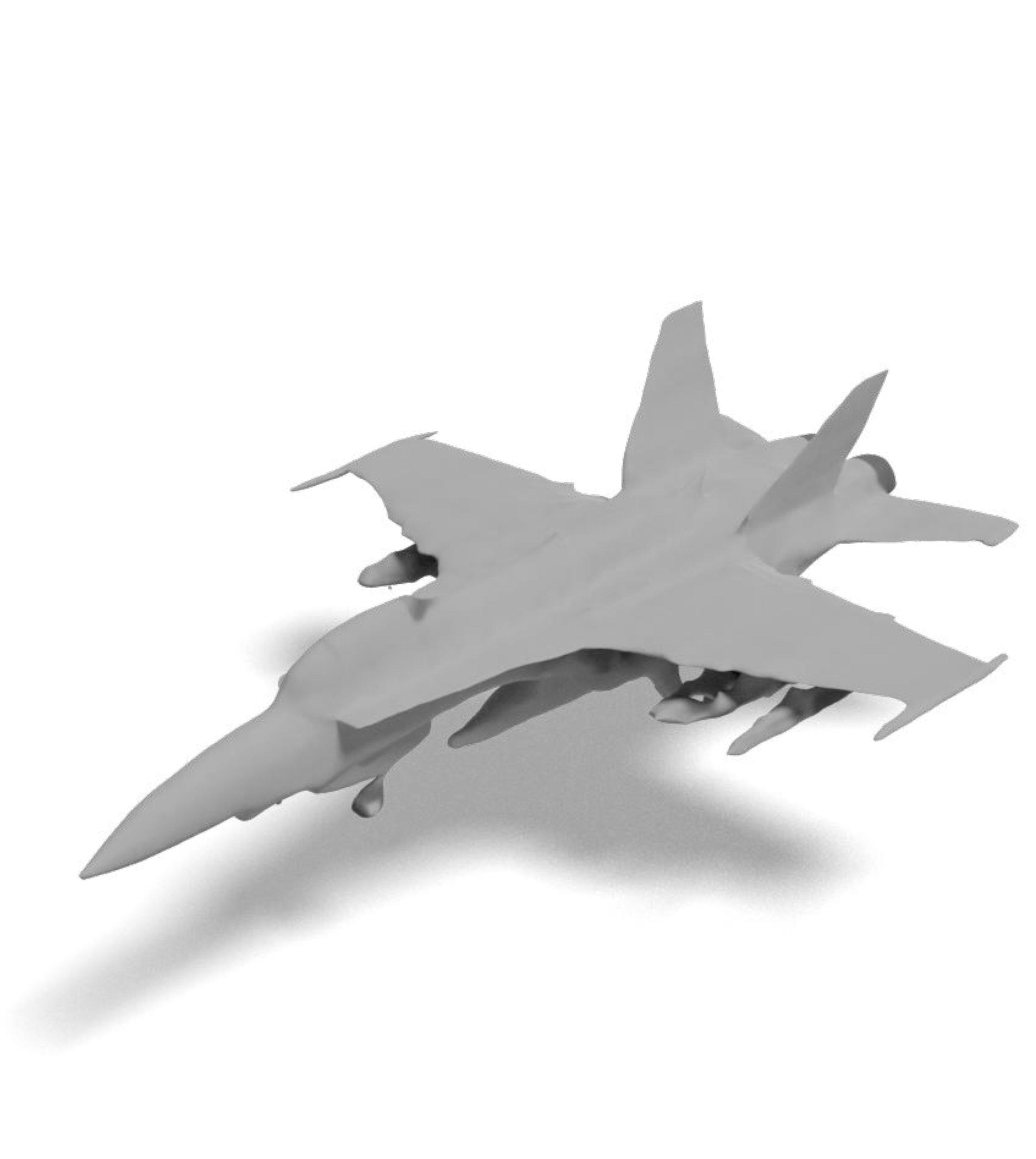}
    \includegraphics[width=0.118\linewidth,trim={300 300 300 300},clip]{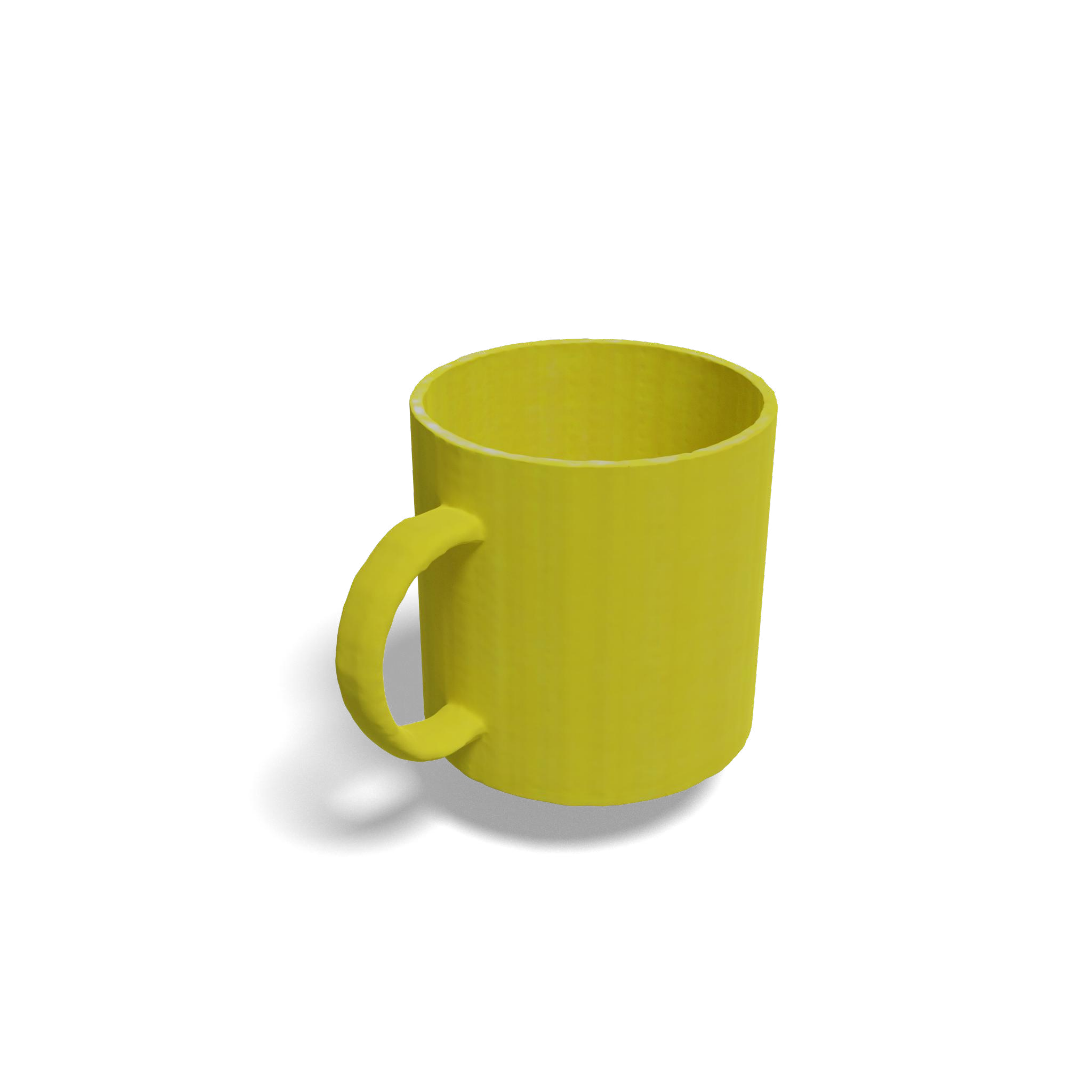}\\

    \includegraphics[width=0.012\linewidth,trim={0 50 0 0 },clip]{iccv2023AuthorKit/figures/cond_gen_tex_labels/input_image.pdf}
    \includegraphics[width=0.118\linewidth,trim={30 30 30 30},clip]{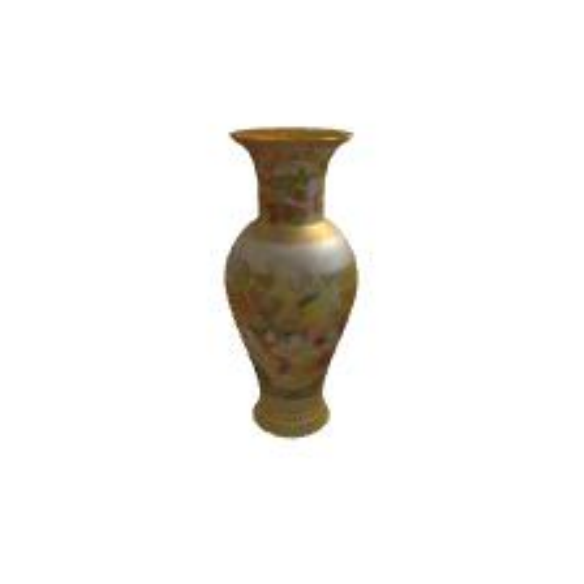}
    \includegraphics[width=0.118\linewidth,trim={30 30 30 30},clip]{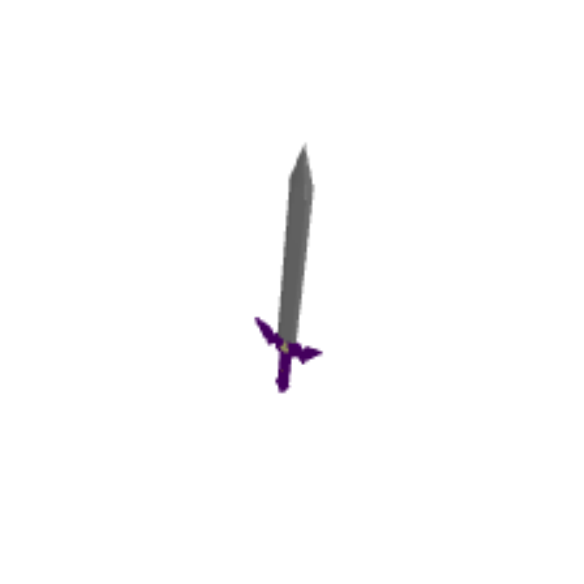}
     \includegraphics[width=0.118\linewidth,trim={30 30 30 30},clip]{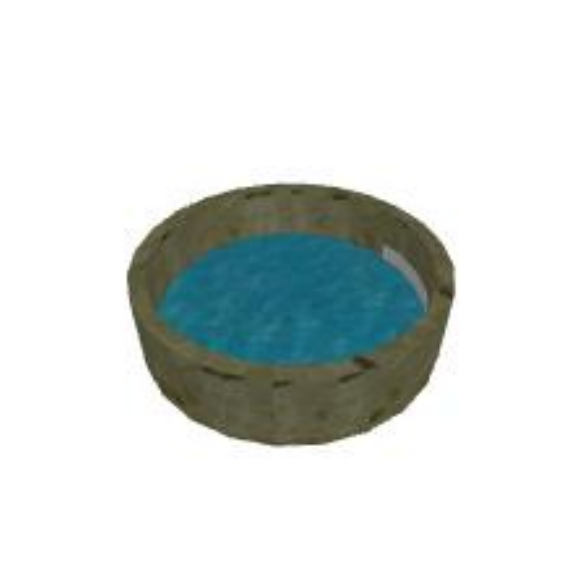}
    \includegraphics[width=0.118\linewidth,trim={30 30 30 30},clip]{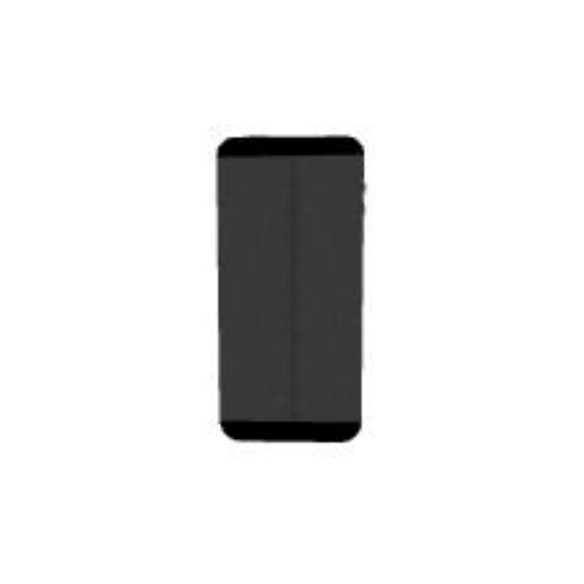}
    \includegraphics[width=0.118\linewidth,trim={30 30 30 30},clip]{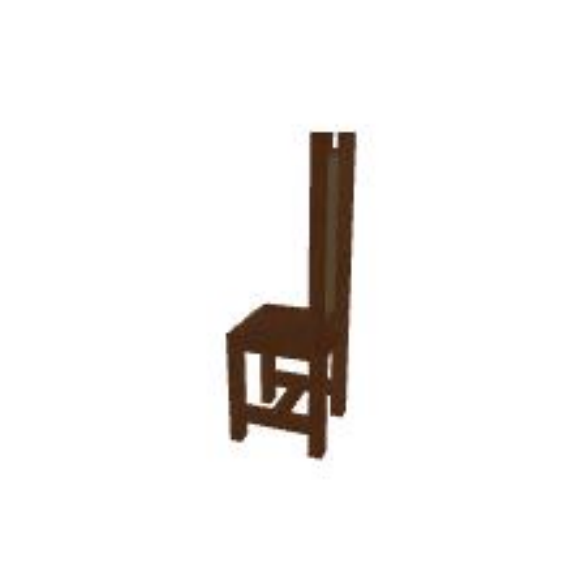}
    \includegraphics[width=0.118\linewidth,trim={30 30 30 30},clip]{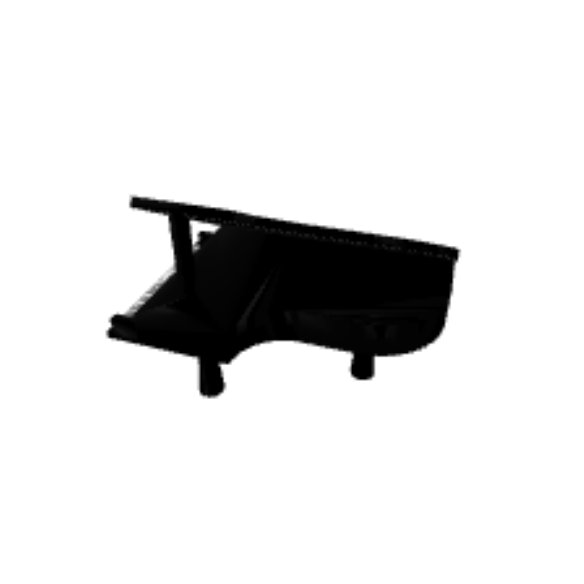}
    \includegraphics[width=0.118\linewidth,trim={30 30 30 30},clip]{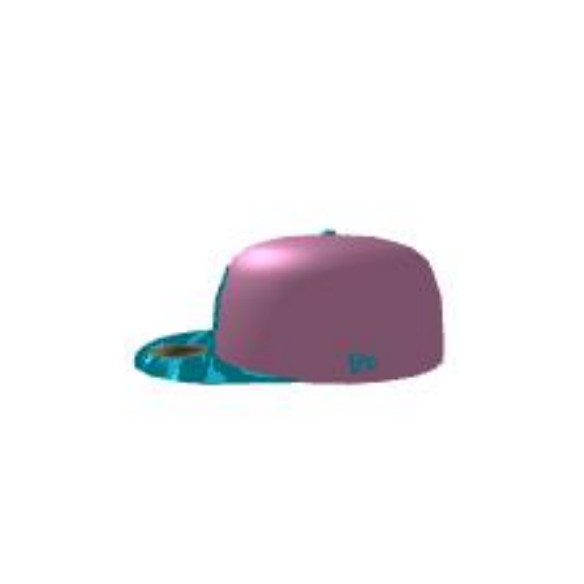}
    \includegraphics[width=0.118\linewidth,trim={30 30 30 30},clip]{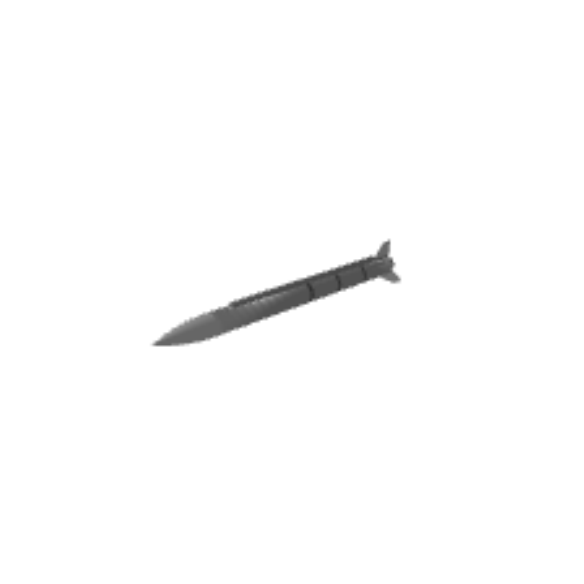}\\\vspace{-7pt}
    \includegraphics[width=0.012\linewidth,trim={0 0 0 0 },clip]{iccv2023AuthorKit/figures/cond_gen_tex_labels/mesh.pdf}
    \includegraphics[width=0.118\linewidth,trim={500 300 500 300},clip]{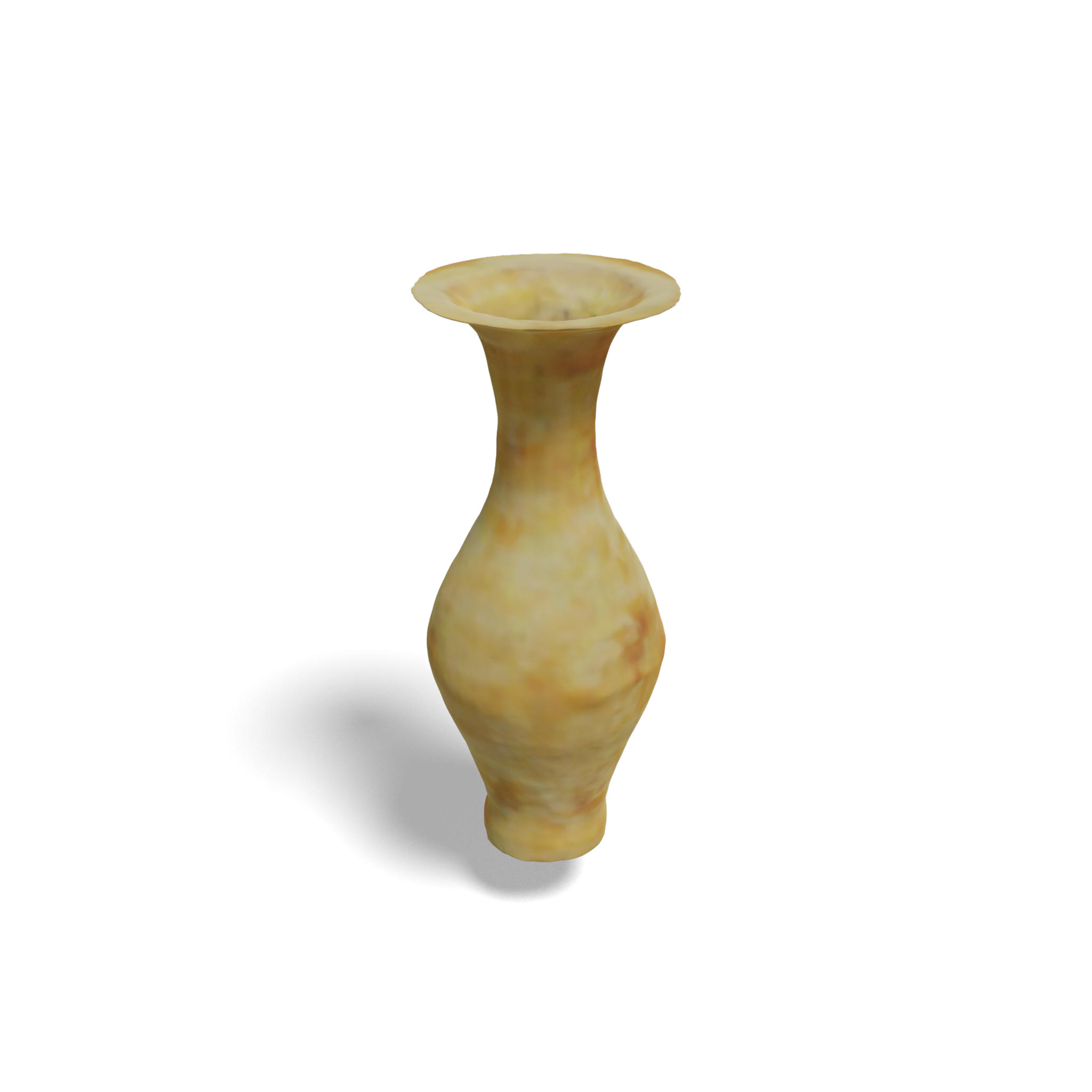}
    \includegraphics[width=0.118\linewidth,trim={500 300 500 300},clip]{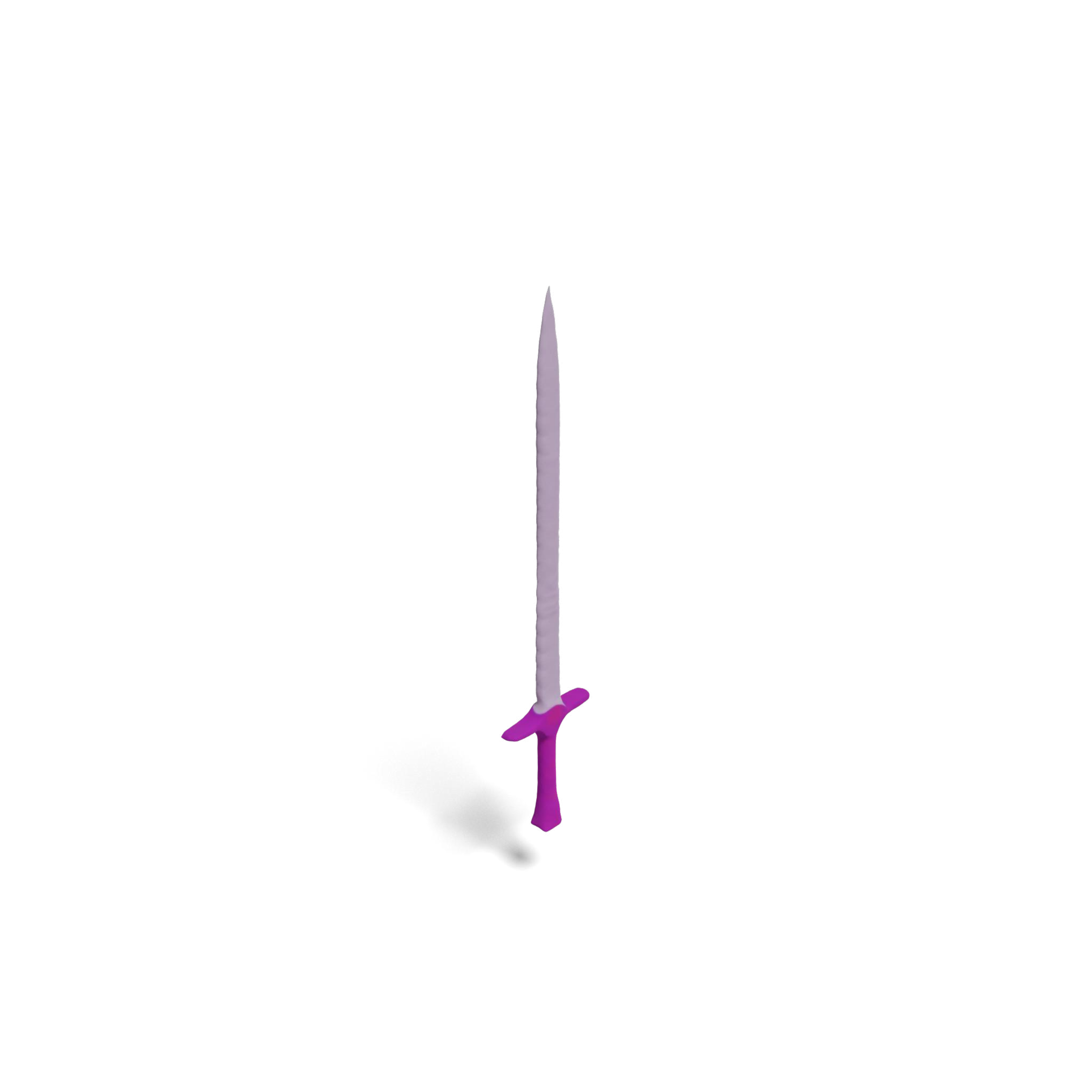}
    \includegraphics[width=0.118\linewidth,trim={500 500 500 650},clip]{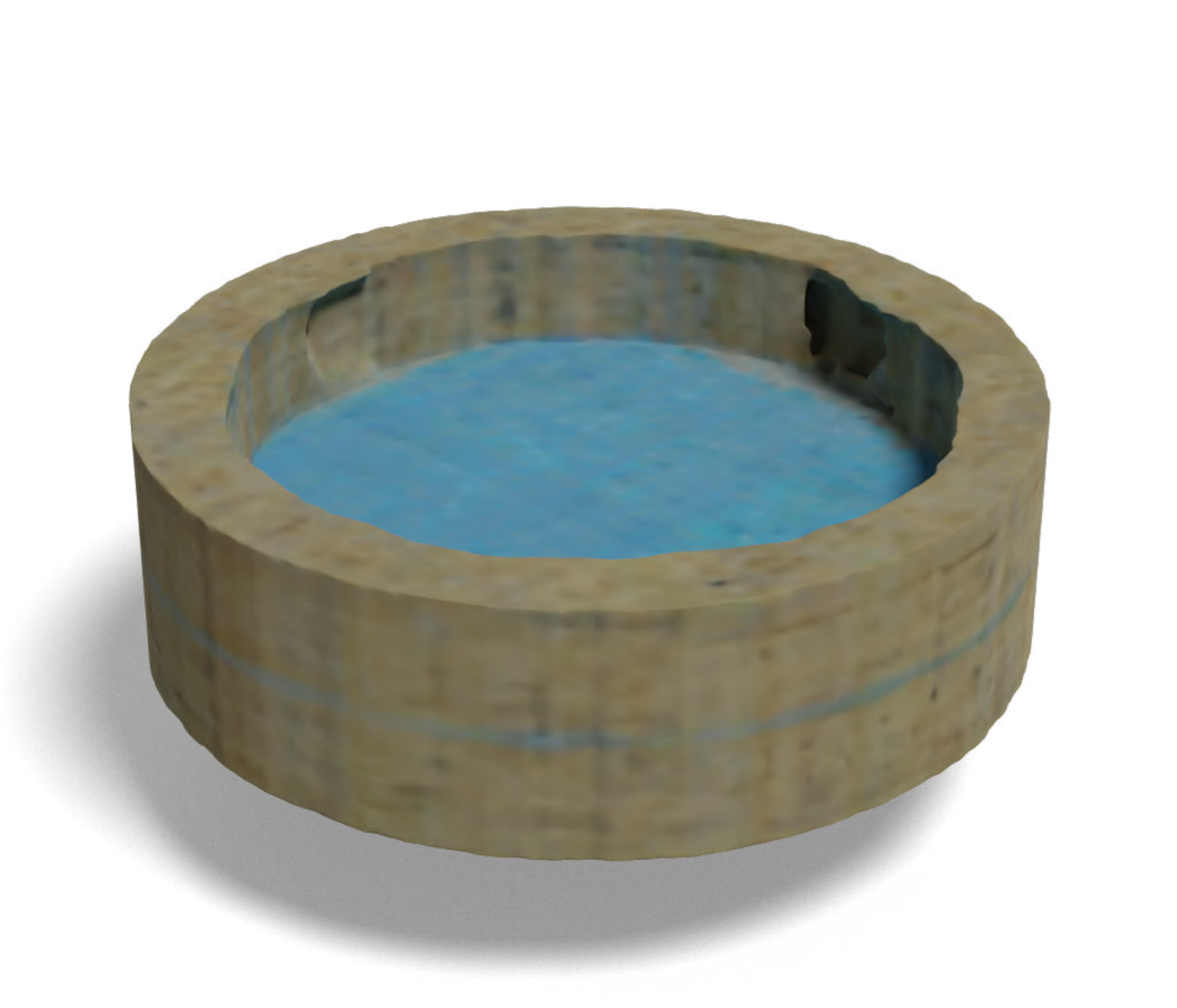}
    \includegraphics[width=0.118\linewidth,trim={450 300 450 300},clip]{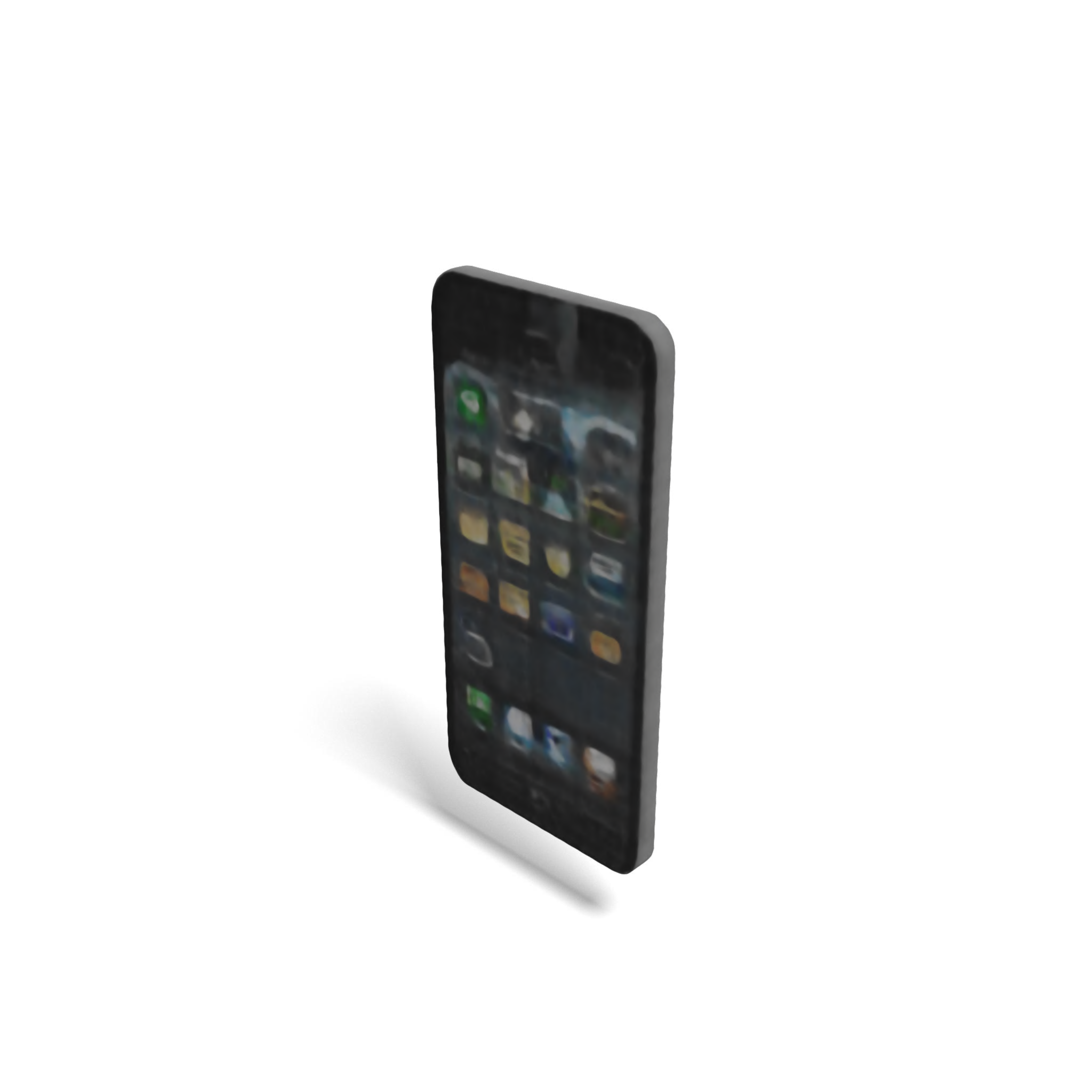}
    \includegraphics[width=0.118\linewidth,trim={450 300 450 300},clip]{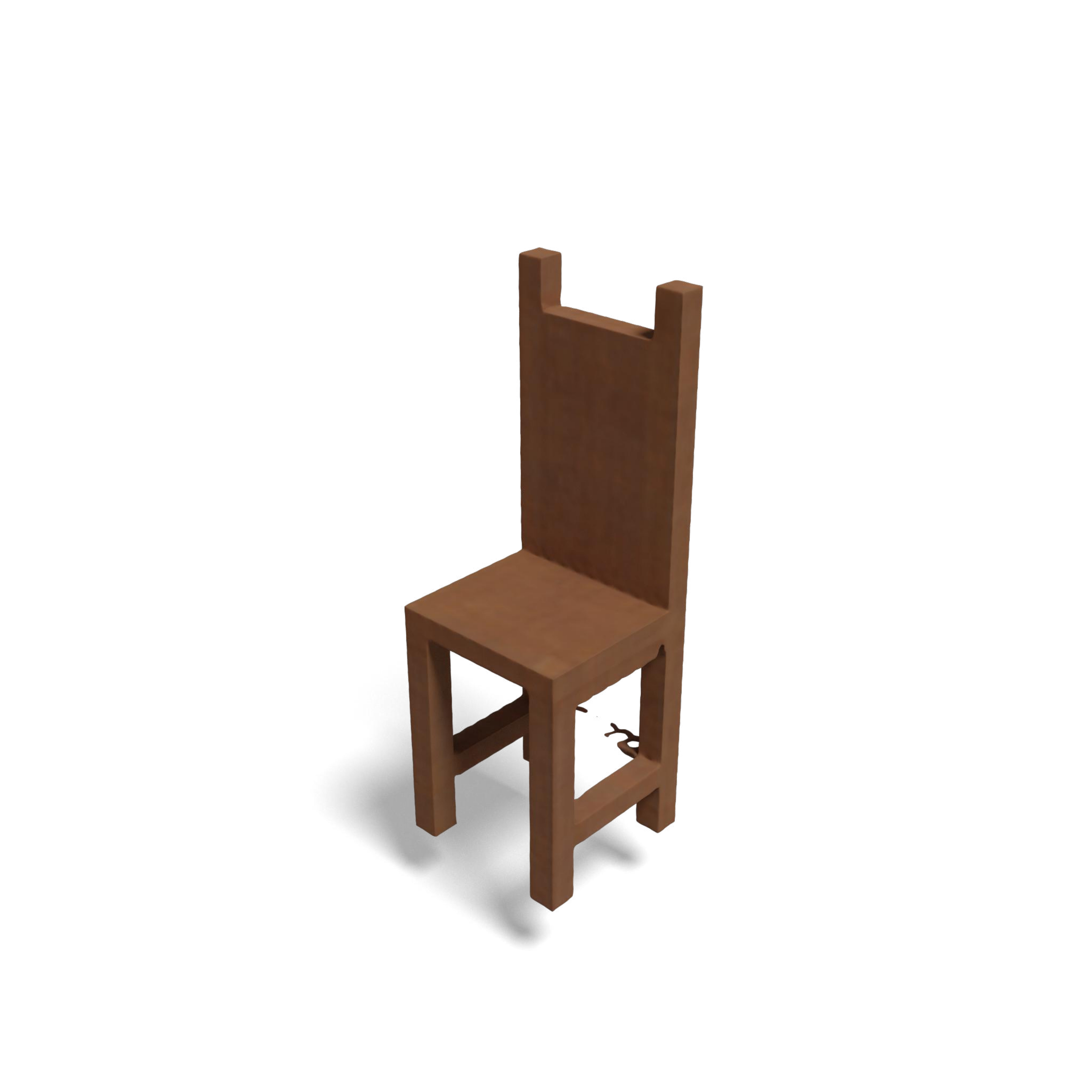}
    \includegraphics[width=0.118\linewidth,trim={400 280 400 580},clip]{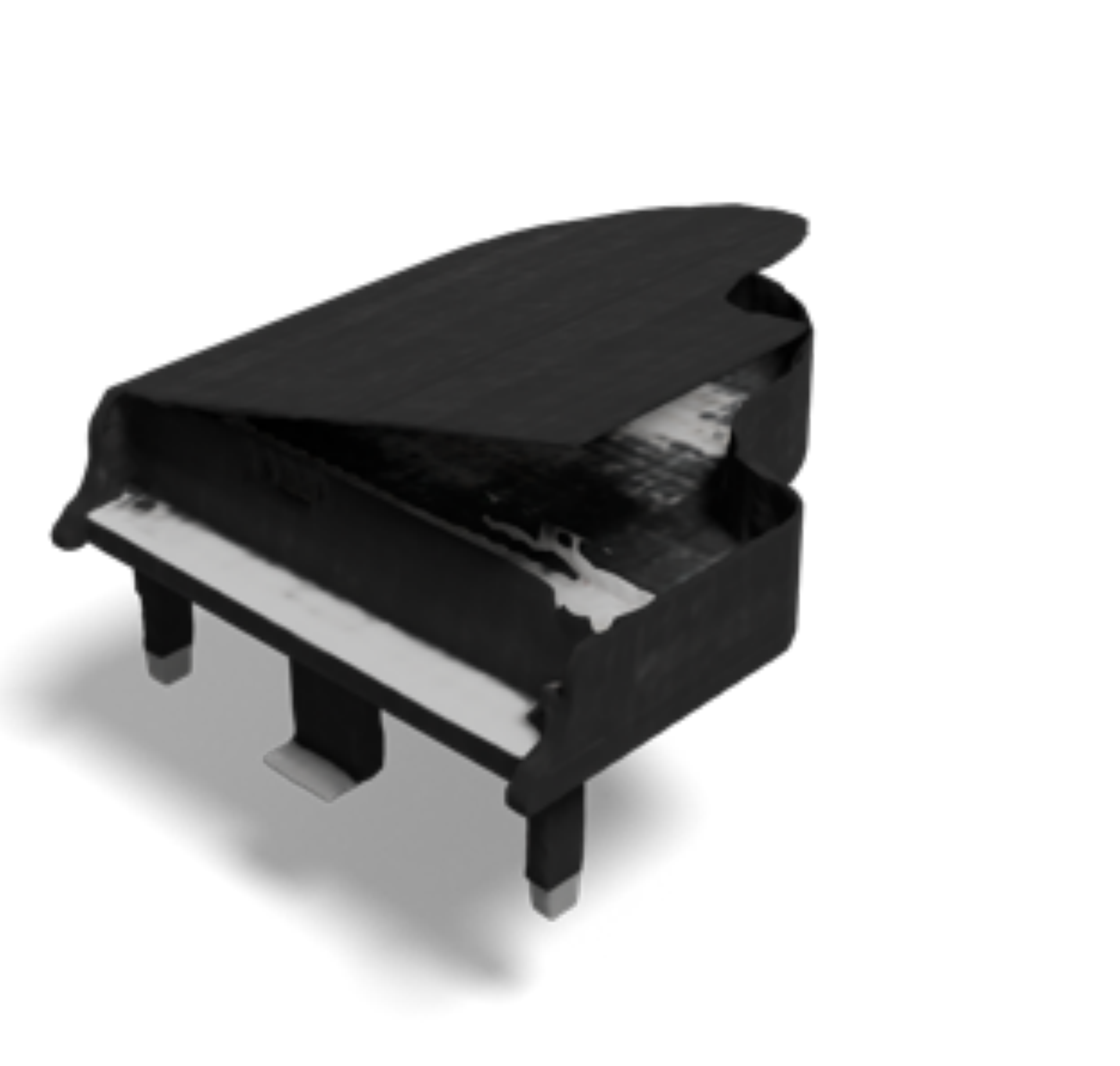}
    \includegraphics[width=0.118\linewidth,trim={500 400 500 820},clip]{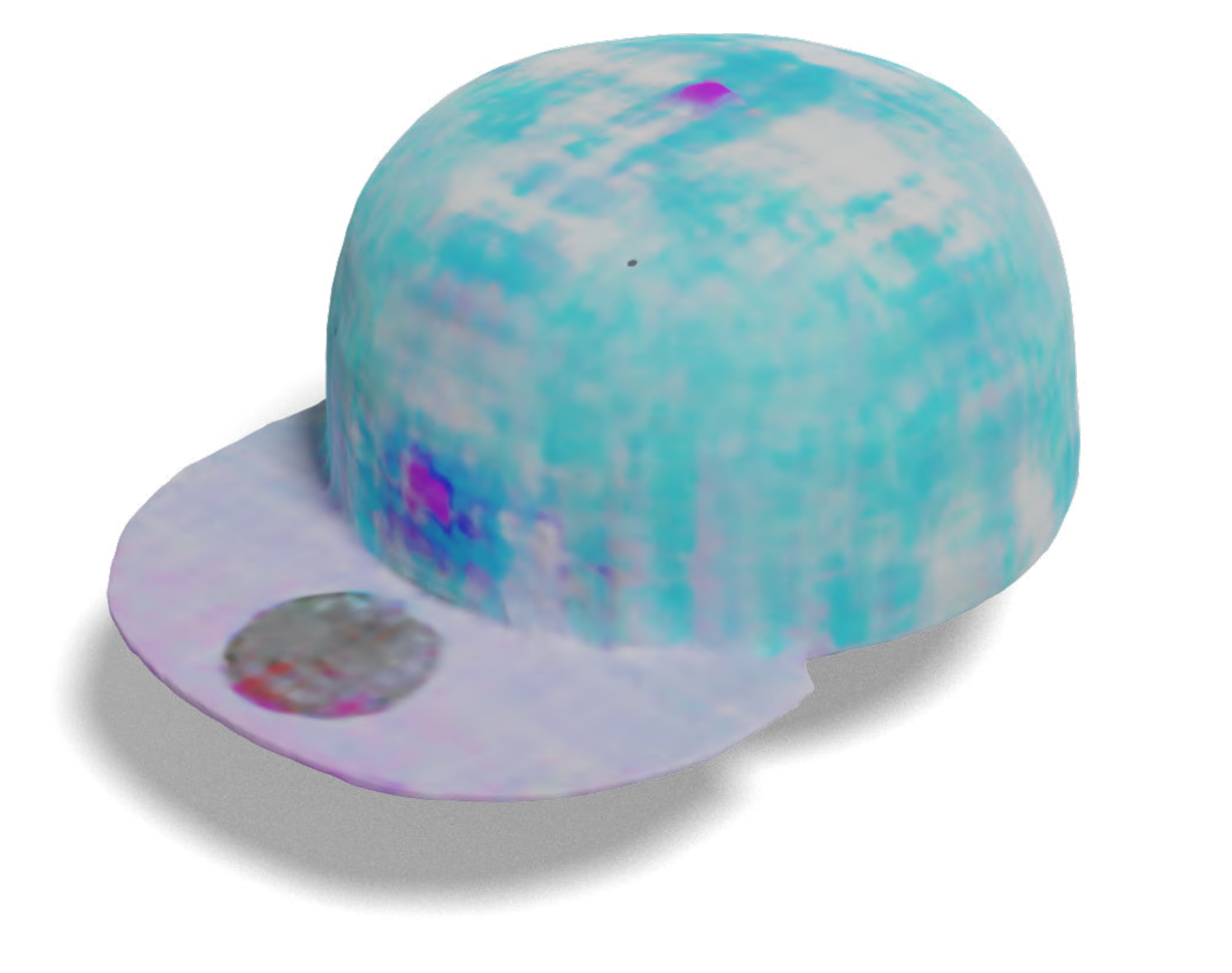}
    \includegraphics[width=0.118\linewidth,trim={400 300 400 800},clip]{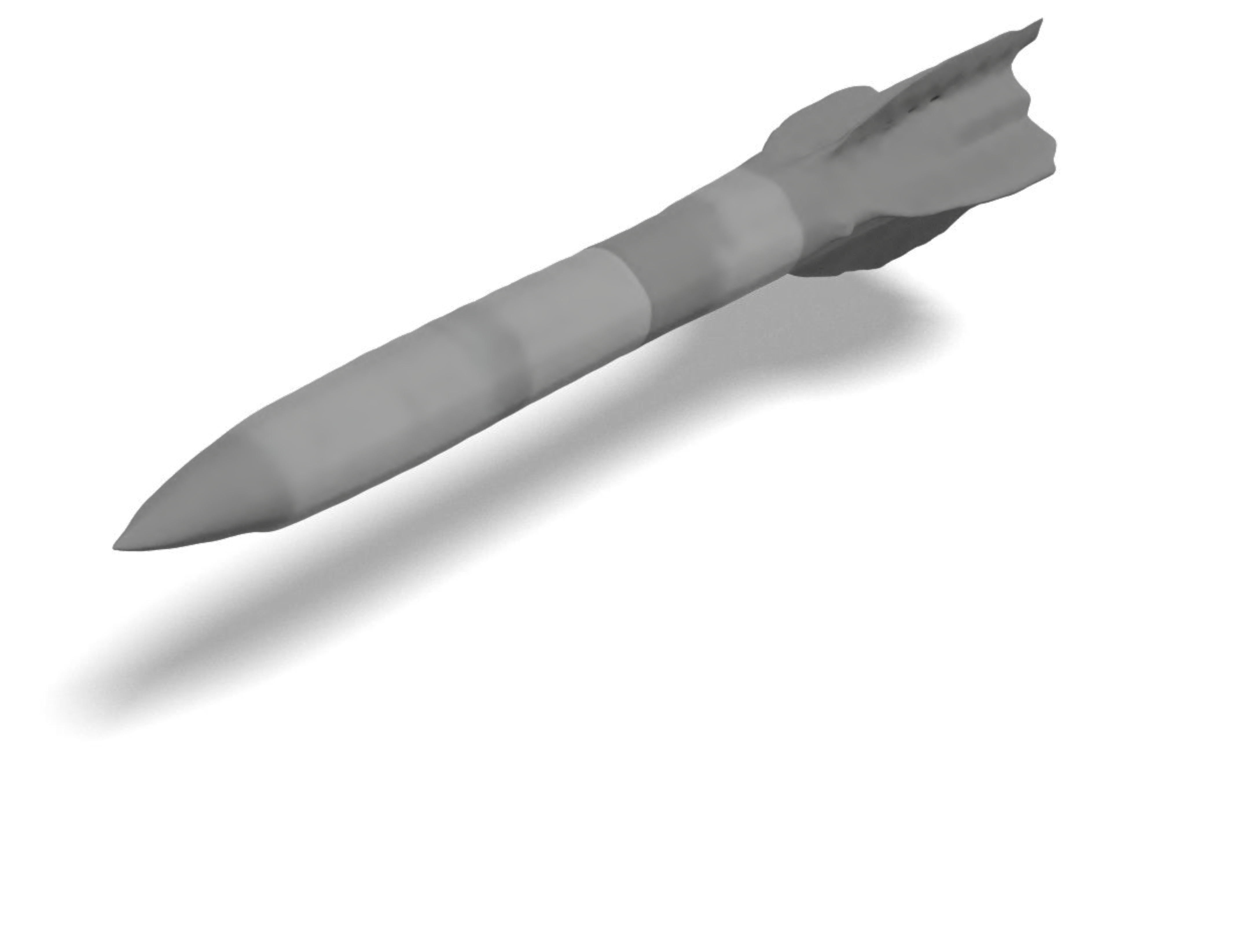}\\\vspace{-5pt}

    \caption{Generated textured meshes conditioned on respective input images}
    \label{fig:cond_tex}
\end{figure*}

\begin{figure}[t]
    \centering
    % \includegraphics[width=0.3\linewidth,trim={0 0 10 30},clip]{iccv2023AuthorKit/figures/cond_tex/g/11g_i.pdf}
    % \includegraphics[width=0.3\linewidth,trim={410 320 410 520},clip]{iccv2023AuthorKit/figures/cond_tex/g/11g_r.pdf}
    % \includegraphics[width=0.3\linewidth,trim={350 20 380 900},clip]{iccv2023AuthorKit/figures/cond_tex/pe/11g_pe.pdf}
    % \\
    % \vspace{-5pt}
    % \includegraphics[width=0.3\linewidth,trim={30 30 30 30},clip]{iccv2023AuthorKit/figures/cond_tex/g/16g_i.pdf}
    % \includegraphics[width=0.3\linewidth,trim={620 520 620 860},clip]{iccv2023AuthorKit/figures/cond_tex/g/16g_r.pdf}
    % \includegraphics[width=0.3\linewidth,trim={550 190 550 950},clip]{iccv2023AuthorKit/figures/cond_tex/pe/16g_pe.pdf}
    \includegraphics[width=0.9\linewidth]{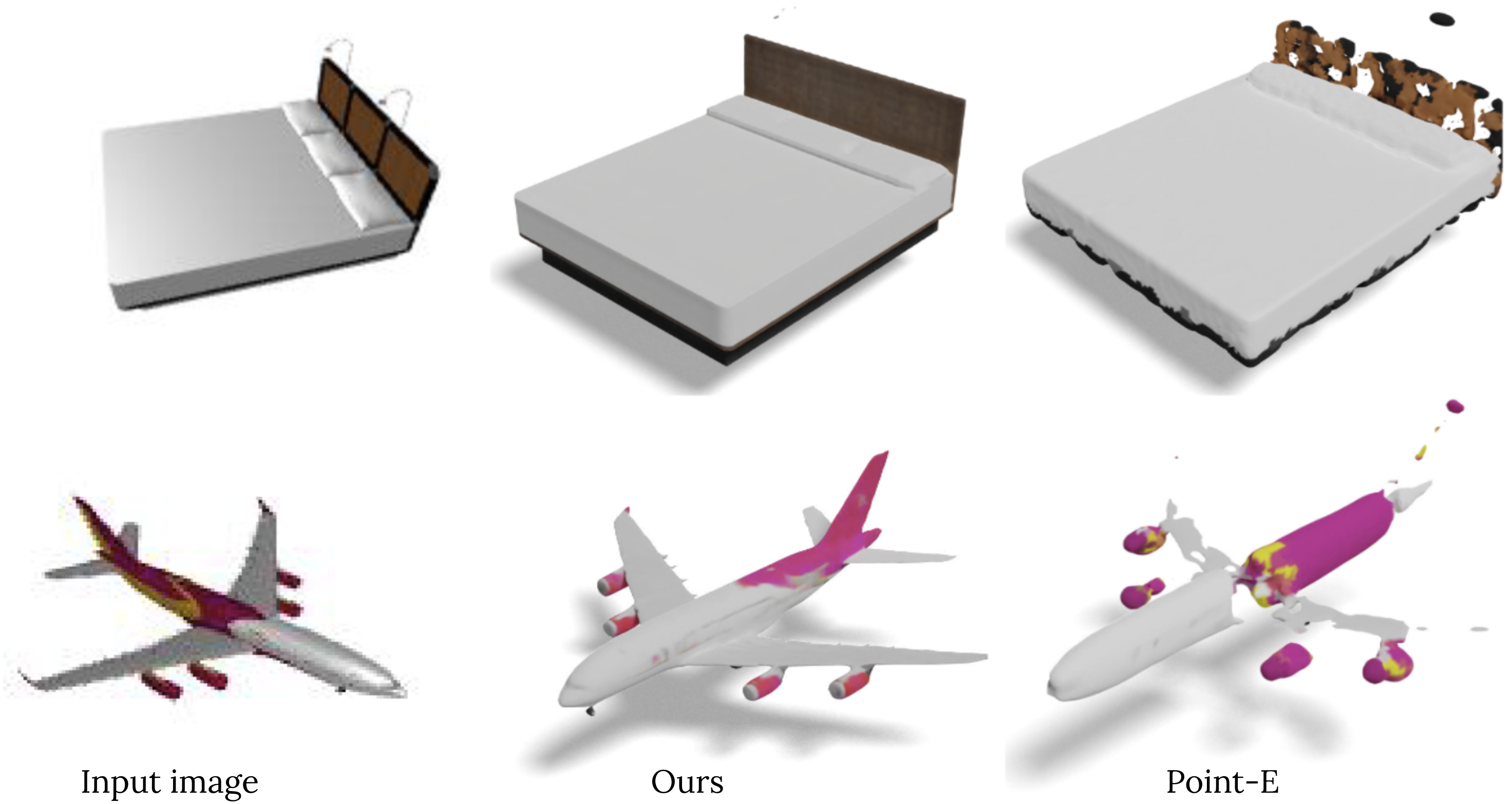}\vspace{-5pt}
    \caption{Our model {\footnotesize} vs. Point-E {\footnotesize} on image$\rightarrow$mesh}
    \label{fig:cond_tex_vs_pe}
\end{figure}

% 2g , 11g  , 16g
\paragraph{Image-Conditioned Generation} 
For conditioned generation, we train a VAE on all categories of the 3DWarehouse data and train the diffusion model similarly as the geometric-only model. Figure~\ref{fig:cond_tex} shows qualitative image$\rightarrow$mesh samples. Additionally we compare our model to Point-E, a recent transformer-based point cloud diffusion model. We used their 1B image-to-pointcloud model to generate the colored point cloud and then their pointcloud-to-mesh model to obtain a textured mesh, as instructed in their codebase\footnote{\url{https://github.com/openai/point-e}}. As shown in Figure~\ref{fig:cond_tex_vs_pe}, the meshes generated by our model have much higher quality than those by Point-E. We suspect that the limitation of Point-E is due to its point cloud nature and naively combining it with a point-to-mesh model fails to generate closed mesh surfaces and smooth textures. 

\begin{table}
\centering
    \begin{tabular}{l|c|c}
    \toprule
    Model & Chamfer-L1 ($\downarrow$) & Mask-IoU ($\uparrow$) \\
    \midrule
    3DILG & 0.069 & 0.938 \\
    3DILG + DMTet & 0.063 & 0.970 \\
    \textbf{3DGen} & \textbf{0.054} & \textbf{0.979} \\
    \bottomrule
    \end{tabular}%
\caption{VAE reconstruction quality metrics}%
\label{tab:vae}
\end{table}%
\section{Analysis}
\paragraph{VAE Variants}\label{sec:vae_variants} As discussed in Section~\ref{sec:vae}, we train the triplane VAE directly with render losses (using the differentiable mesh decoding algorithm DMTet) instead of training the decoder to regress the occupancy/SDF values. This allows us to skip the lossy watertight process that smooths out geometry details. As shown in Table~\ref{tab:vae}, we first use the 3DILG baseline to demonstrate the advantage of render losses and DMTet (silhouette and depth) over SDF losses in terms of reconstruction quality. Our triplane VAE replaces the discrete latents with triplane latents and can further reduce reconstruction error. Note that while the triplane latents are larger in dimension than 3DILG's latents, the continuous and 2D nature of them are more compatible with state-of-the-art image diffusion models.
\vspace{-0.4cm}
\paragraph{Effect of staged VAE training} \label{sec:staged_vae} Second stage decoder only finetuning improves fine details and mesh smoothness while avoiding the computational cost of directly training with a higher tetrahedral resolution. To quantify the benefits of this we take our unconditional diffusion model trained on planes(geometry only) and swap in different versions of the VAE decoder. As can be seen in Table \ref{tab:staged_vae}, increasing the tetrahedral grid resolution used in inference has a big impact on shading FiD. Finetuning the decoder with this higher resolution gives further improvements. We observe similar trends in the textured mesh models.
\paragraph{Effect of conditioning model}
The choice of the conditioning encoder used during diffusion is crucial in capturing nuances of the input. This is similar to the finding in text2image models where state of the art models like Imagen \cite{imagen} use very large text encoders(4.6B parameters). We experiment with 3 different sized bi-encoder models for encoding the image ranging from 300M to 1.8B parameters. Table \ref{tab:cond_model} shows the generated mesh alignment improvement in terms of chamfer distance on the validation set.
% \vspace{-0.5cm}
\begin{table}
\captionsetup{font=footnotesize,labelfont=footnotesize}
\footnotesize
\begin{minipage}{.5\linewidth}
\begin{tabular}{l|c|c}
\toprule
Encoder & \#params & CD($\downarrow$) \\
\midrule
ViT-L14 & 304M & 19.25 \\
ViT-H14 & 633M & 16.52\\
ViT-bigG14 & 1.8B & 16.08 \\
\bottomrule
\end{tabular}%
\caption{\footnotesize Image encoder ablation}
\label{tab:cond_model}
\end{minipage}%
\begin{minipage}{.5\linewidth}
\begin{center}
\begin{tabular}{l|c|c}
\toprule
$\psi_{dec}$ ft  & Grid res. & FiD($\downarrow$) \\
\midrule
$\times$ & 90 &  44.6 \\
$\times$  & 128 &  34.1 \\
$\checkmark$ & 128 & 24.9 \\
\bottomrule
\end{tabular}
\caption{\footnotesize VAE training ablation}
\label{tab:staged_vae}
\end{center}
\end{minipage}
\end{table}

\section{Related Work}
\paragraph{Latent Shape Representations:} To avoid directly dealing with high-dimension 3D data (e.g., meshes, voxels and point clouds), recent shape generation methods adopt a two-stage training process: an autoencoder is first trained to learn low-dimensional latent shape representations, and a generator is later trained to generate the latent representations instead of the original 3D data. CLIP-forge~\cite{clip_forge} learns a single-vector representations of shapes and trains normalizing flow network to generate the latent vector conditioned on CLIP embeddings. Methods like AutoSDF~\cite{autosdf}, 3DILG~\cite{3dilg} and Shapecrafter~\cite{shapecrafter} apply autoencoding to learn a sequence of discrete latent codes of 3D shapes and use transformer models to generate the codes autoregressively. Our work follow the same paradigm but choose to use continuous triplane latents which can more accurately reconstruct 3D shapes and are compatible with existing 2D diffusion generators under minor architecture changes.

% diffusion methods
\paragraph{Diffusion-based 3D Generation:} The success of diffusion models in image generation~\cite{ddpm,dalle2,stablediffusion,ho2022cascaded} has inspired research extending diffusion models to 3D. For instance, \cite{luo2021diffusion,zhou20213d} applied diffusion models to uncolored Point Clouds, with relatively light-weight architectures. Recently, Point-E~\cite{pointe} trains a transformer-based diffusion model which generates colored point cloud from CLIP embeddings. Thanks to data and model scaling, Point-E shows great generalization capability in generating diverse 3D Point Clouds from complex text prompts. However, like previous methods, Point-E requires an additional approximation step to recover the surface from Point Cloud and often fails to generate high-quality meshes usable in 3D applications. For mesh generation, methods including DiffusionSDF~\cite{diffusionsdf}, NFD~\cite{NFD}, Rodin~\cite{rodin} use diffusion models to generate latent mesh representations, which can be decoded into meshes by using marching cubes on signed distance field (SDF)~\cite{deepsdf} or occupancy fields~\cite{occnet}. Despite their capability in generating high-quality meshes, the effectiveness of these methods is mostly confined to narrow or single-class domains and/or un-textured meshes. Our approach, though using a similar paradigm, proposes new techniques to improve both the latent representations and the diffusion model, leading to a more generalized model across diverse categories with the ability to capture texture.
% \vspace{-0.2cm}
% dreamfusion, dreamfield
\paragraph{3D Generation Guided by Text-Image Models:} Another line of work leverages image-text models trained on massive datasets to provide supervision for 3D generation. For instance, methods like DreamField~\cite{dreamfield}, CLIP-Mesh~\cite{clip_mesh} and PureCLIPNeRF~\cite{pure_clip} learn to generate 3D models by optimizing the CLIP similarity scores between rendered images and the input text prompts. Instead of relying on CLIP which might be insufficient to capture high frequency details, more recent methods~\cite{dreamfusion,magic3d,dream3d} use text-to-image generation models to provide alignment supervision between text prompts and rendered images, and generate more photo-realistic 3D models. However these methods fit a separate network for every generation, resulting in impractical inference latency.

\section{Conclusion}
3D object generation has many exciting applications from game design to Augmented and Virtual reality(AR/VR). We have presented a strong textured mesh generation model which can produce high quality output in a few seconds on current GPU hardware.  We have also shown quality and diversity improvements with data scaling.  However, there is still a large gap between the generality of our model and state-of-the-art image generation models, which have been trained on billions of images.  Future work can focus on further closing this gap, using 2D image datasets as weak supervision, or utilising 2D generative models in various ways to aid 3D generation.

{\small
\bibliographystyle{ieee_fullname}
\bibliography{main}
}

\end{document}